\documentclass{article}

\usepackage{microtype}
\usepackage{graphicx}
\usepackage{subcaption}
\usepackage{booktabs} %
\usepackage{natbib} %
\usepackage{enumitem}
\usepackage{multirow}

\usepackage{hyperref}

\usepackage[accepted]{icml2026}

\usepackage{amsmath}
\usepackage{amssymb}
\usepackage{mathtools}
\usepackage{amsthm}
\usepackage{bbm}

\usepackage{tikz}
\usetikzlibrary{arrows}
\usetikzlibrary{shapes}
\usetikzlibrary{calc}
\usetikzlibrary{backgrounds}
\usepackage{tikzbricks}

\definecolor{brickred}{rgb}{0.8, 0.25, 0.33}

\RequirePackage{mathtools}  %
\mathtoolsset{centercolon}  %

\RequirePackage{bm}  %

\usepackage{mathrsfs}

\newcommand{\dif}{\mathop{}\!\mathrm{d}}

\makeatletter
\newcommand{\spx}[1]{%
  \if\relax\detokenize{#1}\relax
    \expandafter\@gobble
  \else
    \expandafter\@firstofone
  \fi
  {^{#1}}%
}
\makeatother

\newcommand\pd[3][]{\frac{\partial\spx{#1}#2}{\partial#3\spx{#1}}}

\newcommand{\od}[3][]{\frac{\dif\spx{#1}#2}{\dif#3\spx{#1}}}

\newcommand{\genericdel}[4]{%
  \ifcase#3\relax
  \ifx#1.\else#1\fi#4\ifx#2.\else#2\fi\or
  \bigl#1#4\bigr#2\or
  \Bigl#1#4\Bigr#2\or
  \biggl#1#4\biggr#2\or
  \Biggl#1#4\Biggr#2\else
  \left#1#4\right#2\fi
}

\let\norm\enVert
\newcommand{\comma}{\; \text{ ,}} %
\newcommand{\point}{\; \text{ .}} %

\DeclareMathOperator*{\argmin}{arg\,min}

\protected\def\subs#1{\ifmmode \textrm{\tiny #1} \else \textsubscript{\tiny #1}\fi}  %

\newcommand{\pr}[1][d]{\operatorname{pr}^{(#1)}}

\newcommand{\IC}[1]{\textit{\MakeTitlecase[words=all]{#1}}}

\usepackage{makecell}
\usepackage{multicol}

\usepackage[nameinlink,capitalize,noabbrev]{cleveref}
\crefformat{equation}{{\color{mydarkblue}(#2\textcolor{mydarkblue}{#1}#3)}} %
\theoremstyle{plain}
\newtheorem{theorem}{Theorem}[section]

\theoremstyle{definition}

\theoremstyle{remark}

\makeatletter
\def\@endtheorem{\hfill{\lower0.3ex\hbox{\ensuremath{\triangle}}}\endtrivlist\@endpefalse } %
\makeatother
\newtheoremstyle{ownexample}{\parskip}{\topsep}{}{}{}{:}{1em}{\textbf{\thmname{#1}\thmnumber{ #2}}\thmnote{ [#3]}}
\theoremstyle{ownexample}
\newtheorem{example}[theorem]{Example}

\usepackage{soul}

\newcommand{\new}[2][]{#2}

\usepackage{xspace}
\newcommand{\bvtheta}{\boldsymbol{\vartheta}}
\newcommand{\btheta}{\boldsymbol{\theta}}
\newcommand{\ba}{\boldsymbol{a}}
\newcommand{\LieSolver}{\textsc{LieSolver}\xspace}

\usepackage{pifont}
\newcommand{\yessym}{\ding{52}}
\newcommand{\nosym}{\ding{56}}
\newcommand{\circsym}{\ding{108}}
\newcommand{\opencircsym}{\ding{109}}

\usepackage[textsize=tiny]{todonotes}

\begin{document}

\twocolumn[
    \icmltitle{\textsc{LieSolver}: PDE-Constrained Learning for IBVPs via Lie Symmetries}

    \icmlsetsymbol{equal}{*}

    \begin{icmlauthorlist}
        \icmlauthor{René P. Klausen}{equal,hhi}
        \icmlauthor{Ivan Timofeev}{equal,hhi}
        \icmlauthor{Jonas Naujoks}{hhi}
        \icmlauthor{Johannes Frank}{}
        
        \icmlauthor{Thomas Wiegand}{hhi,tu,bifold}
        \icmlauthor{Sebastian Lapuschkin}{hhi,dublin}
        \icmlauthor{Wojciech Samek}{hhi,tu,bifold}
    \end{icmlauthorlist}

    \icmlaffiliation{hhi}{Department of Artificial Intelligence, Fraunhofer Heinrich Hertz Institute, Berlin, Germany}
    \icmlaffiliation{tu}{Department of Electrical Engineering and Computer Science, Technische Universität Berlin, Germany}
    \icmlaffiliation{bifold}{BIFOLD Berlin Institute for the Foundations of Learning and Data, Berlin, Germany}
    \icmlaffiliation{dublin}{Centre of eXplainable Artificial Intelligence, Technological University Dublin, Ireland}

    \icmlcorrespondingauthor{René P. Klausen}{rene.pascal.klausen@hhi.fraunhofer.de}

    \icmlkeywords{Machine Learning, ICML, Physics-Informed Machine Learning, Physics-Informed Neural Networks, Scientific Machine Learning, Lie Symmetries, Partial Differential Equations, Initial-Boundary Value Problems}

    \vskip 0.3in
]

\printAffiliationsAndNotice{\icmlEqualContribution}  %

\setcounter{footnote}{4} %

\begin{abstract}
    Initial-boundary value problems (IBVPs) provide the essential framework for modelling a wide range of phenomena in physics and engineering. We introduce a novel method for efficiently solving IBVPs using Lie symmetries to enforce the associated partial differential equation (PDE) exactly by construction. By leveraging symmetry transformations, our model embeds the underlying physical laws and learns the solution solely from initial and boundary data. Consequently, the boundary loss directly quantifies domain-wide error, enabling rigorous error estimation for well-posed IBVPs. We implement \LieSolver and demonstrate its application to linear homogeneous PDEs, showing that it outperforms physics-informed neural networks (PINNs) in both speed and accuracy while yielding compact models. Overall, our approach significantly enhances the efficiency and reliability of predictions for PDE-constrained problems.

\end{abstract}

\section{Introduction}

The growing availability of data and computational resources has established machine learning (ML) as a powerful tool across scientific disciplines. Despite this great success, many models operate as black boxes, limiting their transparency and reliability. This has driven demand for models that are not only accurate but also transparent, data-efficient, and consistent \cite{rudin_stop_2019,samek_explaining_2021}. Incorporating domain knowledge addresses several of these challenges \cite{von_Rueden_2021}. In scientific contexts, partial differential equations (PDEs) --- which govern a wide range of phenomena from physics to finance --- provide a natural framework for such integration. \emph{Physics-informed neural networks} (PINNs) \cite{raissi2019physics} represent the most prominent architecture in this field and have achieved strong results in diverse domains of application \cite{Karniadakis2021PhysicsInformed}. %

Concretely, models like PINNs aim to find specific solutions of PDEs that satisfy additional initial and/or boundary conditions; this setting is commonly referred to as an \emph{initial-boundary value problem} (IBVP). A typical example is predicting the evolution of a dynamical system from its initial state. These IBVPs appear in nearly every scientific discipline, making the development of reliable and efficient IBVP solvers a central challenge in computational science. While classical numerical methods, such as finite difference and finite element techniques \cite{leveque2007finite}, can effectively address these problems, their computational cost scales poorly with problem dimensionality. This curse of dimensionality makes high-dimensional PDEs particularly challenging and has motivated the development of meshless machine learning approaches.%

There are generally three strategies for embedding physical laws in ML models \cite{Karniadakis2021PhysicsInformed}: (i) incorporating them in the optimization, (ii) integrating them into the model architecture, or (iii) augmenting data using physics-based principles. While (iii) requires large amounts of costly-to-generate data, methods like PINNs (i) introduce atypical loss terms that may lead to serious failures \cite{krishnapriyan2021characterizingpossiblefailuremodes}. We therefore adopt the second approach, leveraging the inherent symmetries of PDEs --- namely their Lie symmetries --- by integrating them directly into the model structure. This is applicable in principle to any PDE with known Lie symmetries, as these symmetries inherently generate parametrized and expressive solutions. Moreover, using Lie symmetries instead of directly enforcing PDEs results in more stable and reliable model behaviour.

Our method integrates predetermined parametrized Lie symmetries as learnable building blocks, enabling the model to compose symmetry transformations while provably remaining in the solution space of the underlying PDE. Consequently, for any choice of model parameters, the output exactly satisfies the PDE, reducing the optimization problem to fitting the initial-boundary conditions only. This makes optimization significantly more tractable and eliminates the difficult-to-handle PDE loss terms widely considered the principal challenge in PINNs.

As a result, a decrease in the loss function directly corresponds to improved prediction accuracy, making the objective function, unlike in PINNs, a reliable metric for evaluating training progress. Additionally, for well-posed IBVPs, rigorous error bounds can be derived. This approach also enables significantly smaller models (in terms of parameters) while preserving expressiveness, allowing the use of robust second-order optimization schemes that yield more accurate approximations in significantly less time. Currently, \LieSolver is restricted to linear homogeneous PDEs. Although this class already contains many important problems --- such as the wave, heat, advection, convection-diffusion, Klein-Gordon, and Maxwell equations --- we are confident that this restriction can be relaxed, since the underlying Lie symmetry principle applies to any PDE.

\begin{figure}[t]
    \centering
    \begin{tikzpicture}[
        trafo/.style={ellipse, draw=black, inner sep=1pt, fill=teal!30}, 
        arr/.style={-latex, thick}, 
        brick/.style={rectangle, draw=brickred, fill=brickred, opacity=0.2, rounded corners=2pt, anchor=south west, minimum width=2cm, minimum height=0.6cm},
        infbox/.style={rectangle, draw=black, fill=teal!30, rounded corners=1pt},
        scbrick/.style={scale=0.5, opacity=0.4},
        bbrick/.style={color=red},
        plots/.style={color=blue!70!black},
        scale=.6, transform shape
        ]
        \def\brangA{85}
        \def\brangB{15}
        
        \node[inner sep=0pt] (input) at (-1,4.3) {\includegraphics[width=.17\textwidth]{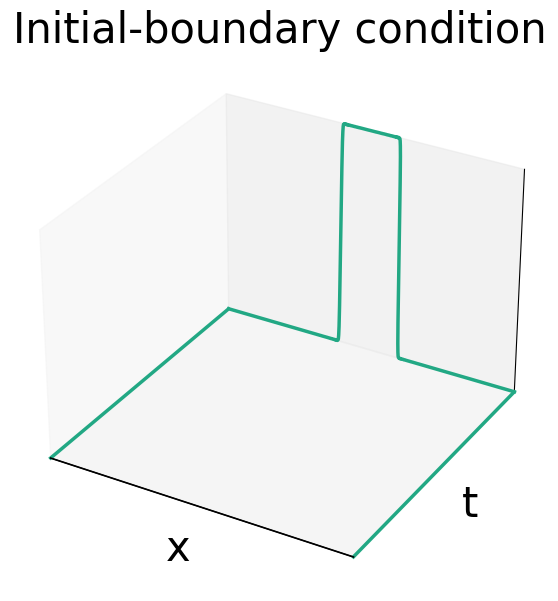}};
        \node[inner sep=0pt] (output) at (10,4.3) {\includegraphics[width=.17\textwidth]{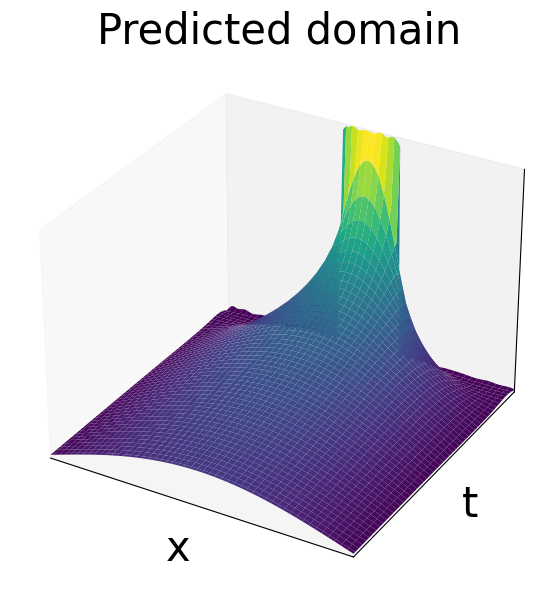}};
        \draw[arr, very thick] (0.5,4.1) -- ++(1,0); \draw[arr, very thick] (7.5,4.1) -- node[above,xshift=1pt] {\scriptsize $\displaystyle\argmin_{\ba,\bvtheta}$} ++(1,0);
        
        \draw[black, very thick, rounded corners=2pt, fill=lightgray!30] (0,0) rectangle (9,3);
        \draw[black, very thick, rounded corners=2pt, fill=lightgray!30] (1.5,3.2) rectangle (7.5,5);
        
        \node[above] at (4.5,5.1) {\textsc{LieSolver}};
        \node[below right] at (1.5,5) {\small Optimize Brick Combination};
        \node[below right] at (0,3) {\small Construct Brick Families $\mathcal S$};

        \node (form) at (4.5,3.75) {$\displaystyle f_\text{LS} (x;\ba,\btheta) = \sum_i a_i\ \ f_i^{j_i} \big(x;\bvtheta_i^{j_i}\big)$};
        \begin{scope}[scbrick, shift={($(form)+(1.3,-0.27)$)}]
            \tdplotsetmaincoords{\brangA}{\brangB}
            \brick[bbrick]{4}{1}
        \end{scope}
        \node at (form) {$\displaystyle f_\text{LS} (x;\ba,\btheta) = \sum_i a_i\ \ f_i^{j_i} \big(x;\bvtheta_i^{j_i}\big)$}; %

        \def\lh{1}
        \def\llh{2}

        \node (fstart) at (-2,\lh) {$f_\text{start}$};
        \begin{scope}[scale=0.5, shift={(fstart.south west)}, opacity=0.4]
            \tdplotsetmaincoords{\brangA}{\brangB}
            \brick[color=red]{4}{1}
        \end{scope}
        \node at (fstart) {$f_\text{start}$}; %
        
        \begin{scope}[shift={(-1.,\lh)}]
            \draw[scale=0.5,domain=-1:.9,smooth,variable=\x, plots] plot ({\x}, {0.15});
        \end{scope}
        \node[trafo] (T1) at (1.5,\llh) {$T^3$};
        \node[trafo] (T2) at (3,\llh) {$T^1$};
        \node[trafo] (T3) at (4.5,\llh) {$T^4$};
        \node (dots1) at (5.5,\llh) {$\ldots$};

        \node[trafo] (T4) at (1.5,\lh) {$T^2$};
        \node[trafo] (T5) at (3,\lh) {$T^3$};
        \node[trafo] (T6) at (4.5,\lh) {$T^2$};
        \node (dots2) at (5.5,\lh) {$\ldots$};

        \node (B1) at (7,\llh) {$f^1$};
        \begin{scope}[scbrick, shift={(B1.south west)}]
            \tdplotsetmaincoords{\brangA}{\brangB}
        \brick[bbrick]{4}{1}
        \end{scope}
        \node at (B1) {$f^1$};
        \begin{scope}[shift={(8,\llh)}]
            \draw[scale=0.5, domain=-1:1, samples=100, variable=\x, plots] plot ({\x}, {0.2 * sin ( 10 * \x r)});
        \end{scope}

        \draw[arr] ($(B1)+(0.5,0.5)$) to[bend angle=15, bend right] ($(form)+(2,-.2)$);

        \node (B2) at (7,\lh) {$f^2$};
        \begin{scope}[scbrick, shift={(B2.south west)}]
            \tdplotsetmaincoords{\brangA}{\brangB}
            \brick[bbrick]{4}{1}
        \end{scope}
        \node at (B2) {$f^2$};
        \begin{scope}[shift={(8,\lh)}]
            \draw[scale=0.5,domain=-1:1,smooth,variable=\x, plots] plot ({\x}, {0.7 * exp(- 5 * \x * \x) - 0.28});
        \end{scope}

        \node (vdots1) at ($(T4) - (0,0.5)$) {$\vdots$};
        \node (vdots2) at ($(B2) - (0,0.5)$) {$\vdots$};

        \draw[thick] (-.4,\lh) -- (0.25,\lh); \draw[arr] (0.25,\lh) to[bend angle=35, bend left] (T1.west); \draw[arr] (0.25,\lh) -- (T4.west);
        \draw[arr] (T1) -- (T2); \draw[arr] (T2) -- (T3); \draw[arr] (T3) -- ($(dots1.west)+(0.1,0)$); \draw[arr] (dots1.east) -- (B1);
        \draw[arr] (T4) -- (T5); \draw[arr] (T5) -- (T6); \draw[arr] (T6) -- ($(dots2.west)+(0.1,0)$); \draw[arr] (dots2.east) -- (B2);

        \node[infbox] (LS) at (4.5,-1) {Lie symmetries $T^j$};
        \node[align=center, infbox] (physics) at (-1.4,-1) {physical laws\\ $f_t=f_{xx}$};
        \draw[arr] (physics) -- (LS); \draw[arr] (LS) -- (4.5,0);
        
    \end{tikzpicture}
    \caption{The structure of \textsc{LieSolver}, a physics-constrained solver for IBVPs (e.g., time evolution problems). The method consists of two parts. Based on Lie symmetries, one constructs appropriate \emph{bricks} that provably satisfy the physical laws (lower part). This step needs to be performed only once for each physical law (e.g., the heat equation). These bricks are then combined by solving the associated optimization problem to match initial-boundary conditions (upper part). The model provably satisfies the physical laws exactly and approximates the initial-boundary conditions. A detailed description of the pipeline can be found in \cref{ssec:liesolver_pipeline}.}
    \label{fig:overview_LieSolver}
\end{figure}

To summarize the contributions of our work, we provide:
\begin{itemize}[itemsep=1pt,topsep=0pt]
    \item A novel machine learning framework, \textsc{LieSolver}\footnote{In German, ``Lies'' is the imperative of ``lesen'' (``read''); ``LieSolver'' playfully suggests ``Lies Olver'' --- to read Olver. We use Olver's book \cite{olver1993applications} as our primary source on Lie symmetries.} based on Lie symmetries for solving IBVPs. This method (i) enforces PDEs exactly through a PDE-informed architecture, (ii) aligns loss with prediction accuracy, (iii) requires no external training data, as arbitrary data points can be sampled as needed, (iv) allows for rigorous and sharp error bounds, and (v) provides improved interpretability through access to a symbolic decomposition of the solution.
    \item An implementation\footnote{The implementation is available \href{https://github.com/oduwancheekee/liesolver}{here}.} of \textsc{LieSolver}, with benchmark experiments demonstrating higher solution accuracy and outperforming vanilla PINNs by several orders of magnitude in computation time.
\end{itemize}

\section{Related Work}
\paragraph{ML Approaches for IBVPs}
Various methods have been developed over time that employ machine learning techniques to tackle IBVPs. Early approaches to solving differential equations with neural networks (NNs) date back to the pioneering work of \cite{LagarisArtificialNeuralNetworks1998}; later work such as \cite{TsoulosSolvingDifferentialEquations2006} explored the recovery of analytical closed-form solutions through genetic programming.

The introduction of PINNs \cite{raissi2019physics} revitalized interest in the field of physics-informed machine learning and the application of ML-based techniques for solving PDEs. 
Subsequent work has focused on improving training stability and scalability using a plethora of approaches, including domain decomposition \cite{jagtap2020extended}, adaptive loss balancing \cite{wang2021understanding}, improved optimizers \cite{wang2025gradientalignment,Rathore2024}, resampling \cite{WuComprehensiveStudyNonadaptive2023,LauPINNACLEPINNAdaptive2023,NaujoksLeveragingInfluenceFunctions2025} and multifidelity extensions \cite{meng2020ppinns}. Although promising, PINNs often suffer from unstable convergence \cite{WangTengPerdikaris2021, krishnapriyan2021characterizingpossiblefailuremodes} and practical error estimation and certification remain an ongoing field of research \cite{eiras2024efficienterrorcertificationphysicsinformed,hillebrecht2025predictionerrorcertificationpinns}.

\paragraph{Symmetry-Based Approaches}

In the realm of physics, symmetries play a dominant role \cite{wigner1967symmetries}, inspiring diverse machine learning approaches which are ``symmetry-informed''. More generally, this includes, for example, convolutional networks that hard-code translation symmetry \cite{cohen2016group}, equivariant networks for three-dimensional particle systems \cite{thomas2018tensor}, and structure-preserving models such as Hamiltonian \cite{greydanus2019hamiltonian} or Lagrangian NNs \cite{cranmer2020lagrangian}. These architectures ensure physical consistency by design, but are usually specialized to certain physical aspects. 

For PDE-related symmetry approaches more concretely, current research includes symmetry-informed data augmentation for neural PDE solvers \cite{BrandstetterLiePointSymmetry2022}, self-supervised pretraining based on Lie symmetry transformations \cite{MialonSelfSupervisedLearningLie2023}, and symmetry-regularized PINNs \cite{Akhound-SadeghLiePointSymmetry2023}, which all exploit the invariance structure of the underlying equations to improve generalization and sample efficiency. Similarly, \cite{ZhangSymmetryGroupBased2023} leverages Lie symmetry groups to generate data for supervised learning, providing an alternative to physics-informed training. Other methods embed invariant surface conditions into the loss \cite{zhang2022enforcing} to enhance the PINN method. Yet another recent line of work \cite{YangLatentSpaceSymmetry2024,YangSymmetryInformedGoverningEquation2024} utilizes Lie symmetries to discover them from data and integrate them into learning architectures, improving long-term predictive accuracy and facilitating practical applications such as equation discovery. 

\new{
\paragraph{Trefftz-type methods}
A natural strategy for solving linear homogeneous IBVPs is to approximate the solution by a linear combination of \emph{trial functions} that satisfy the governing PDE exactly by construction, reducing the problem to fitting boundary data alone. This idea underlies the Trefftz method~\cite{trefftz1926} and its variants~\cite{kolodziej2018many,hiptmair2016survey}, including the method of fundamental solutions~\cite{fairweather1998method}. The central limitation of these methods is that rich families of such trial functions are known only for a very limited number of PDEs, typically those admitting closed-form solution families such as harmonic polynomials or plane waves. \LieSolver shares this PDE-exact principle, but overcomes this limitation by generating trial functions --- called \emph{bricks} --- systematically from the Lie symmetry group of the PDE, making the construction applicable to any PDE with known symmetries. Moreover, the symmetry parameters $\bvtheta$ enter each brick nonlinearly, yielding a more expressive approximation than a fixed linear basis --- at the cost of a structured nonlinear optimization problem, which \LieSolver addresses via greedy composition and variable projection.

}

\section{Background}
\label{sec:theoretical_background}

\subsection{Initial-Boundary Value Problems (IBVPs)}
\label{ssec:IBVPs}

IBVPs model a wide range of phenomena in science and engineering and are the primary object of study in this work: They couple partial differential equations (PDEs) with initial-boundary conditions (IBCs). Concretely, let $\Omega\subset\mathbb R^n$ be an open, bounded, and connected domain and $f:\Omega\to\mathbb R^m$ be a sufficiently smooth function. We focus here on \emph{linear homogeneous PDEs}, i.e.
\begin{align}
    \mathcal D[f](x) = \sum_{|\alpha|\leq d} c_\alpha(x) \partial^\alpha f(x) = 0 \quad\forall x\in\Omega\ , \label{eq:LinearHomogeneousPDE}
\end{align}
where the coefficients $c_\alpha(x)$ are given functions and $\alpha\in\mathbb N^n_0$ is a multi-index, $|\alpha|:=\sum_i\alpha_i$, $\partial^\alpha:=\partial_1^{\alpha_1}\cdots\partial_n^{\alpha_n}$. This class covers many key PDEs, such as the wave, heat, advection, convection-diffusion, Klein-Gordon, and Maxwell equations. Any sufficiently smooth function $f:\Omega\to\mathbb R^m$ satisfying \cref{eq:LinearHomogeneousPDE} is called a \emph{solution of the PDE}. In particular, the constant function $f\equiv 0$ is always a solution for any linear homogeneous PDE.

Typically, there are many possible solutions to a PDE. Therefore, we restrict the possible solutions through IBCs. Let $\Gamma_1,\ldots,\Gamma_r\subseteq\partial\Omega$ be smooth connected components of the boundary of $\Omega$. We denote the IBCs as operators
\begin{align}
    \mathcal B_k[f](x) = 0  \quad \forall x\in\Gamma_k\;\; \text{and}\;\; k=1,\ldots,r\point \label{eq:IBC}
\end{align}
These operators $\mathcal B_k$ usually restrict the function $f$ itself (Dirichlet condition) or its normal derivative (Neumann condition) to certain predefined values, but they can model any kind of (mixed) initial or boundary conditions. Functions $f:\overline\Omega\to\mathbb R^m$ satisfying \cref{eq:LinearHomogeneousPDE} and \cref{eq:IBC} are called \emph{solutions of the IBVP} defined by $\mathcal D,\mathcal B_1,\ldots,\mathcal B_r$.

There are various strategies to solve IBVPs with numerical or symbolic methods. For example, PINNs \cite{raissi2019physics} formulate \cref{eq:LinearHomogeneousPDE} and \cref{eq:IBC} as an optimization problem, by training a NN $\phi_\theta(x)$ minimizing the loss terms $\norm{\mathcal D[\phi_\theta](x)}$ and $\norm{\mathcal B_k[\phi_\theta](x)}$.

However, not all IBVPs admit a solution. In the following, we therefore always focus on \emph{well-posed} IBVPs, meaning that (i) the IBVP has a solution, (ii) the solution is unique, and (iii) the solution depends continuously on the prescribed initial and boundary data \cite{John1978}. Further details and references on PDEs and IBVPs are provided in \cref{app-sec:PDEs_IBVPs}.

\subsection{Lie Symmetries}
\label{ssec:lie_symmetries}

PDEs are invariant under certain symmetries, and so are their solutions. These relations are called \emph{Lie symmetries}, which are continuous symmetries among the PDE solutions. Moreover, Lie symmetries are determined solely by the PDE and form parametrized groups \cite{olver1993applications}. Hence, if $T_\vartheta$ is such a Lie symmetry transformation of a given PDE depending on a real parameter $\vartheta\in\mathbb R$ and $f$ is an arbitrary solution of that PDE, then $T_\vartheta \circ f$ is a solution of the PDE as well. Due to the group property of Lie symmetries, the concatenation of Lie symmetries yields another valid Lie symmetry.

We close this short overview by discussing the Lie symmetries for the one-dimensional heat equation
\begin{align}
    \mathcal D[f](x,t) = \pd{f(x,t)}{t} - \pd[2]{f(x,t)}{x} = 0 \label{eq:heat_equation} \comma
\end{align}
which admits six one-parameter symmetry transformations \cite{olver1993applications}:
\begin{align}
    T^1_\vartheta &: f(x,t) \mapsto f(x - \vartheta, t)\ , \label{eq:heat1d_trafo1} \\
    T^2_\vartheta &: f(x,t) \mapsto f(x, t - \vartheta)\ , \label{eq:heat1d_trafo2} \\
    T^3_\vartheta &: f(x,t) \mapsto e^{\vartheta} f(x, t)\ , \label{eq:heat1d_trafo3} \\
    T^4_\vartheta &: f(x,t) \mapsto f(e^{-\vartheta} x, e^{-2\vartheta} t)\ , \label{eq:heat1d_trafo4} \\
    T^5_\vartheta &: f(x,t) \mapsto e^{-\vartheta x + \vartheta^2 t} f(x - 2\vartheta t, t)\ , \label{eq:heat1d_trafo5} \\
    T^6_\vartheta &: f(x,t) \mapsto e^{-\frac{\vartheta x^2}{1 + 4\vartheta t}} \frac{f\!\left( \frac{x}{1 + 4\vartheta t}, \frac{t}{1 + 4\vartheta t} \right)}{\sqrt{1 + 4\vartheta t}}\ . \label{eq:heat1d_trafo6}
\end{align}
Hence, for any solution $f(x,t)$ of the PDE \cref{eq:heat_equation}, $T^i_\vartheta\circ f(x,t)$ is another solution of the PDE \cref{eq:heat_equation}, for any $\vartheta\in\mathbb R$ such that $T^i_\vartheta\circ f(x,t)$ is defined. There exists an additional symmetry transformation that reflects the linearity of the PDE
\begin{align}
    T_{\vartheta}^\alpha : f(x,t) \mapsto f(x,t) + \vartheta\, \alpha(x,t) \label{eq:linearLieSym}
\end{align}
where $\alpha(x,t)$ is an arbitrary solution of \cref{eq:heat_equation}. Notably, these symmetries \cref{eq:heat1d_trafo1}--\cref{eq:linearLieSym} constitute the complete set of continuous point symmetries of the solution space.

Any PDE possesses its own set of Lie symmetries, which can be determined algorithmically \cite{Hereman1996}. However, linear homogeneous PDEs always admit symmetries of the form \cref{eq:heat1d_trafo3} and \cref{eq:linearLieSym}. We provide further details about Lie symmetries and their computation in \cref{app-sec:LieSymmetries}.

\section{Methodology of \LieSolver}
\label{sec:liesolver}

\subsection{The Structure of \textsc{LieSolver}}
\label{subsec:structure_of_liesolver}
The key idea of \LieSolver is to approximate the solution of an IBVP by a linear combination of functions generated by parametrized Lie symmetry transformations (see \cref{fig:overview_LieSolver}).
Since Lie symmetries form a group, they can be concatenated to generate new symmetries. Thus, given an arbitrary solution to start with, one can generate a family of new solutions by successive applications of Lie transformations. Finding such a starting solution is often straightforward. For instance, $f(x)\equiv 1$ solves any linear homogeneous PDE that lacks a zero-order term ($|\alpha|=0$), e.g., the heat, wave or Laplace equation. We refer to such solutions as \emph{start solutions}, denoted as $f_{\text{start}}$.

By applying a sequence of Lie transformations $T^{l_1},\ldots,T^{l_s}$ to a given $f_\text{start}$, we obtain an expressive $s$-parameter family
\begin{align}
    f(x;\vartheta_1,\ldots,\vartheta_s)= T^{l_s}_{\vartheta_s} \cdot \ldots\cdot T^{l_1}_{\vartheta_1} \circ f_\text{start}(x) \comma 
    \label{eq:brick}
\end{align}
that remains a solution of the underlying PDE for any value of $\boldsymbol\vartheta=(\vartheta_1,\ldots,\vartheta_s)\in\mathbb R^s$, whenever the composed function is defined. We call $f(x;\boldsymbol\vartheta)$ a \emph{brick solution}. In order to distinguish between symbolic $\bvtheta$ and numerical values, we introduce an index for any specific value of $\bvtheta$. Thus, $f(x;\bvtheta)$ represents a \emph{brick family} of solutions, whereas $f_i(x;\boldsymbol\vartheta_i)$ is a specific function we call \emph{brick}. Different brick families are distinguished by superscripts $f^j(x;\bvtheta^j)$.

We consider a set $\mathcal S = \{f^1(x; \boldsymbol\vartheta^1),\ldots,f^N(x;\boldsymbol\vartheta^N)\}$ of brick families, where each solution $f^j$ is constructed from a start solution by \cref{eq:brick}. The algorithm of \textsc{LieSolver} chooses and instantiates $M$ bricks from $\mathcal{S}$ (elements can repeat), and combines them linearly:
\begin{align}
    f_{\text{LS}}(x; \boldsymbol{a}, \boldsymbol\theta) = \sum_{i=1}^M a_i\, f_i^{j_i}(x;{\boldsymbol\vartheta}_i^{j_i}) \comma
    \label{eq:liesolver_structure}
\end{align}
where the parameters $\boldsymbol\theta=(\boldsymbol\vartheta_1^{j_1},\ldots,\boldsymbol\vartheta_M^{j_M})$ and $ \boldsymbol a=(a_1,\ldots,a_M)$ are specified to certain values. The resulting function $f_{\text{LS}}$ is very expressive and covers a major part of the solution space of the PDE \cref{eq:LinearHomogeneousPDE}. Note that each brick $f_i^{j_i}(x;{\boldsymbol\vartheta}_i^{j_i})$ may consist of several parameters as given in \cref{eq:brick}. We call the collection of bricks appearing in \cref{eq:liesolver_structure} the \emph{active set} $\mathcal A = \{ f_1^{j_1}(x; \boldsymbol\vartheta_1^{j_1}), \ldots, f_M^{j_M}(x; \boldsymbol\vartheta_M^{j_M})\}$. 

Since $f_\text{LS}$ is a solution of the PDE \emph{by construction} for any $(\ba,\btheta)$, the problem of finding a solution to the IBVP is reduced to matching given IBCs $\mathcal B_1,\ldots,\mathcal B_r$ \cref{eq:IBC} by constructing the active set $\mathcal{A}$ and finding optimal parameters $\boldsymbol a$ and $\btheta$. The set $\mathcal{S}$, built with start solutions and selected transformations applied to them, as well as the number of bricks $M$ are the design choices of the model.

\begin{table}[t]
    \centering
    \caption{Comparing purely data-driven NNs with PINNs, \textsc{LieSolver} and classical numerical methods FDM, FVM and FEM. See \cref{ssec:liesolver_scope} for discussion and details.}
    \scriptsize %
    \begin{tabular}{@{}p{.22\linewidth}p{.1\linewidth}p{.12\linewidth}p{.18\linewidth}p{.14\linewidth}@{}} 
        \toprule
        & \footnotesize NN & \footnotesize PINN & \footnotesize \textsc{LieSolver} & \footnotesize classical \\
        \toprule
        coverage of well-posed IBVPs & \multirow{2}{*}{--} & \multirow{2}{*}{non-stiff} & \parbox[t]{\linewidth}{linear\\ homogeneous} & \multirow{2}{*}{any}\\
        \midrule[.5\lightrulewidth]
        PDE constraint & -- & soft & hard & hard\\ %
        IBC constraint & soft & soft/hard & soft & hard\\
        \midrule[.5\lightrulewidth]
        error estimation & \nosym & \nosym & \yessym & \yessym\\
        loss alignment & \yessym & \nosym & \yessym & --\\
        respecting physics & \nosym & \yessym & \yessym & \yessym\\
        meshless/no CoD & \yessym & \yessym & \yessym & \nosym\\
        simple data fusion & \yessym & \yessym & \yessym & \nosym\\
        \# parameters & \circsym\circsym\circsym & \circsym\circsym\opencircsym & \circsym\opencircsym\opencircsym & --\\
        interpretability & \opencircsym\opencircsym\opencircsym & \circsym\opencircsym\opencircsym & \circsym\circsym\opencircsym & \circsym\circsym\circsym\\
        \bottomrule
    \end{tabular}
    \label{tab:method_comparison}
\end{table}

\subsection{The \textsc{LieSolver} Optimization}
\label{ssec:optimization_liesolver}

The construction of the solution $f_{\text{LS}}$ involves determining the optimal active set of bricks $\mathcal{A}$ and optimizing the nonlinear parameters $\boldsymbol\theta$ and the linear amplitudes $\boldsymbol{a}$. Unlike traditional deep learning approaches that minimize a single unified loss function via stochastic gradient descent, \textsc{LieSolver} employs a multi-stage optimization strategy. It utilizes different objectives for structural learning and parameter tuning. After discussing the data structure, we consider the key components of the \LieSolver: linear fitting, greedy composition of the active set $\mathcal{A}$ for structural learning, and nonlinear parameter tuning.

\paragraph{Data Sampling} 

Let $\mathcal X_k\subset\Gamma_k$ be finite sets of randomly sampled collocation points from the boundary parts $\Gamma_k$ according to \cref{ssec:IBVPs}. The full training set $\mathcal X = \bigcup_{k=1}^r \mathcal X_k\in\mathbb R^{L\times n}$ consists of $L$ points $x\in\mathbb{R}^n$.
Consequently, the IBCs $\mathcal B_1,\ldots,\mathcal B_r$ define a set of target values $\mathcal Y\in\mathbb{R}^{L\times m}$ at the datapoints $\mathcal X$. 

For the active set $\mathcal{A}$ of $M$ bricks parametrized by $\boldsymbol\theta$, we define the brick array $\mathbf{F}(\boldsymbol\theta) \in \mathbb{R}^{L \times m \times M}$ as the bricks evaluated at the datapoints $\mathbf{F}(\boldsymbol\theta) = \begin{bmatrix} f^{j_1}_1(\mathcal X;\boldsymbol \vartheta^{j_1}_1) & \dots & f^{j_M}_M(\mathcal X;\boldsymbol \vartheta^{j_M}_M) \end{bmatrix}$, where $f(\mathcal X;\cdot)$ is the element-wise evaluation of the brick on the dataset $\mathcal X$. The residual of the model is thus defined as $r (\ba,\btheta) = \mathcal Y - \mathbf{F}(\boldsymbol\theta)\boldsymbol a$.

\paragraph{Linear Fitting of Amplitudes} A key feature of the \textsc{LieSolver}'s structure \cref{eq:liesolver_structure} is the separability of parameters. For any fixed nonlinear configuration $\boldsymbol\theta$, the optimal amplitudes $\boldsymbol a^\star$ can be determined via linear regression. To ensure numerical stability, we employ Ridge regression with regularization parameter $\lambda>0$:
\begin{equation} \label{eq:amplitudes_by_ridge}
\begin{split}
    \boldsymbol a^\star(\boldsymbol\theta) &= \arg\min_{\boldsymbol a}\!\big(\norm{\mathcal Y - \mathbf{F}(\boldsymbol\theta)\boldsymbol a}_2^2 + \lambda \norm{\boldsymbol a}_2^2\big) \\
    &= \Big(\mathbf{F}(\btheta)\!^\top \mathbf{F}(\btheta)+\lambda\ I\Big)^{-1}\mathbf{F}(\btheta)\!^\top \mathcal Y \point
\end{split}
\end{equation}

\paragraph{Greedy Brick Initialization} 
Initializing all $M$ bricks simultaneously would lead to a stiff optimization landscape due to the heterogeneous nonlinearity of brick functions. Instead, we use a greedy selection strategy and build the active set $\mathcal{A}$ iteratively. At each step, we generate a candidate pool of random parameters $\{ \bvtheta_{p}^j \}_{j=1\dots N,\, p=1\dots P}$. We select the candidate that maximizes the cosine similarity score with the current residual $r(\ba,\btheta)$:
\begin{align}
    s_{j,p}(r) = \frac{ {\big|\langle r, f^j(\mathcal{X};\bvtheta_p^j})\rangle\big|}{\norm{r}_2\ \norm{f^j(\mathcal{X};\bvtheta_p^j)}_2} \comma \label{eq:score_cosine_similarity}
\end{align}
where $f^j(\mathcal X;\bvtheta_p^j)$ is the evaluation of the candidate brick. Note that the dependency on $\boldsymbol{a}$ factors out in \cref{eq:score_cosine_similarity}. The candidate with the highest score is added to $\mathcal{A}$ and $\boldsymbol{a}$ is updated via Ridge regression.

\paragraph{Nonlinear Refinement of Symmetry Parameters}
Greedy construction provides a robust initialization that eases finding a local minimum. To navigate the potentially stiff landscape, we employ nonlinear least squares (NLLS) optimization. The separability of the model structure \cref{eq:liesolver_structure} into linear and nonlinear parameters enables the use of the Variable Projection (VarPro) method \cite{oleary2013variable}. Instead of optimizing $(\boldsymbol{a}, \boldsymbol\theta)$ jointly, we treat the amplitudes as dependent variables $\boldsymbol{a}^\star(\boldsymbol\theta)$ that implicitly solve the linear subproblem \cref{eq:amplitudes_by_ridge} for any given $\boldsymbol\theta$. This reduces the optimization objective to solely the nonlinear parameters:
\begin{align}
    \boldsymbol\theta^{\star} = \arg\min_{\boldsymbol\theta}\! \norm{\mathcal Y - \mathbf{F}(\boldsymbol\theta)\,\boldsymbol a^\star(\boldsymbol\theta)}_2^2 \point
    \label{eq:theta_by_NLLS}
\end{align}
We solve \cref{eq:theta_by_NLLS} using the Trust Region Reflective algorithm \cite{BranchColemanThomasLi99}. It handles parameter bounds explicitly, ensuring the validity of applied symmetry transformations. As a second-order method, it effectively resolves stiffness, particularly when supplied with Jacobians derived from our symbolic brick definitions. To ensure flexibility during training, we apply VarPro optimization both partially, to refine the most recently added bricks, and globally, to refine all parameters of the model.

\paragraph{The \textsc{LieSolver} Algorithm}
We present the pseudocode for the basic implementation of \textsc{LieSolver} in \cref{alg:liesolver}. The procedure alternates between a \emph{structural learning phase}, consisting of $R$ greedy additions, and a \emph{global parameter refinement}, repeating this cycle until the convergence criteria are met.

\begin{algorithm}[tb]
    \caption{Basic LieSolver}
    \label{alg:liesolver}
    \begin{algorithmic}[1]
        \REQUIRE IBC data $(\mathcal X,\mathcal Y)$; brick families $\mathcal{S}$; hyperparameters \texttt{mse\_tol}, \texttt{max\_bricks}, $R$
        \ENSURE bricks $\mathcal A$, amplitudes $\boldsymbol a$
        \STATE $\mathcal{A}\gets\emptyset$, $\boldsymbol a\gets[\ ]$, $r\gets \mathcal Y$, $\mathrm{mse}\gets \|r\|_2^2/L$
        \WHILE{$\mathrm{mse}>\texttt{mse\_tol}$ \textbf{and} $|\mathcal{A}|<\texttt{max\_bricks}$}
            \FOR{1 \dots R}
                \STATE $(j^\star,p^\star) \gets \displaystyle \arg\max_{j, p} s_{j,p}(r)$ 
                        \hfill $\triangleright$ via \cref{eq:score_cosine_similarity}
                \STATE append $f^{j^\star}(x;\bvtheta_{p^\star}^{j^\star})$ to $\mathcal{A}$
                \STATE $\boldsymbol a \gets \text{Ridge}(\mathbf{F}(\boldsymbol{\theta}),\mathcal Y)$
                    \hfill $\triangleright$ via \cref{eq:amplitudes_by_ridge}
                \STATE $r\gets \mathcal Y - \mathbf{F}(\boldsymbol{\theta}) \boldsymbol a$; \quad $\mathrm{mse}\gets \|r\|_2^2/L$
                \IF{$\mathrm{mse}\le\texttt{mse\_tol}$ \textbf{or} $|\mathcal{A}|=\texttt{max\_bricks}$}
                    \STATE \textbf{break}
                \ENDIF
            \ENDFOR
            \STATE $(\boldsymbol{a}, \boldsymbol\theta) \gets \text{VarPro}(\mathbf{F}(\boldsymbol{\theta}), \mathcal{Y})$
                    \hfill $\triangleright$ via \cref{eq:theta_by_NLLS}, \cref{eq:amplitudes_by_ridge}
            \STATE $r\gets \mathcal Y - \mathbf{F}(\boldsymbol\theta) \boldsymbol a$; \quad $\mathrm{mse}\gets \|r\|_2^2/L$
        \ENDWHILE        
    \end{algorithmic}
\end{algorithm}

\subsection{The \textsc{LieSolver} Pipeline}
\label{ssec:liesolver_pipeline}

The workflow (see \cref{fig:overview_LieSolver}) of \LieSolver begins with the construction of the set $\mathcal{S}$ of brick families. One needs to compute Lie symmetries $T_\vartheta$ (see details in \cref{app-ssec:LieComputation}) and select start solutions $f_{\text{start}}$ to compose bricks according to \cref{eq:brick}. \new[An educated guess of transformations to compose a brick family is crucial, as the choice of brick families determines the expressivity of the model and should cover a wide range of solution patterns.]{The choice of brick families determines model expressivity. To guide this choice, we perform a \emph{brick preselection}: we compose random Lie transformation sequences with start solutions, run \LieSolver on generic IBVPs, and retain the families most frequently selected by the greedy algorithm. The preselection of expressive bricks is performed once per PDE and reused across all IBVPs.} Parameter bounds must also be defined to ensure the functions remain well-defined within the domain and avoid singularities. However, small symmetry parameters $|\vartheta|$ are always valid by \new[construction]{locality} of Lie groups.

We then sample points on the boundary parts $\Gamma_k$ to construct $\mathcal X_k$.
To ensure efficient coverage of the boundary components $\Gamma_k$, we employ Halton sampling \cite{halton1960efficiency} for training points. Standard uniform sampling is employed to generate IBC points for testing generalization and domain points for validating the solution against ground truth.

Finally, the \LieSolver configuration requires selecting hyperparameters: the candidate pool size $P$, the refinement interval $R$, and the maximum number of function evaluations \texttt{nfev} for the NLLS optimization \cref{eq:theta_by_NLLS}. These settings control the navigation in the loss landscape. While large $P$ and \texttt{nfev} ensure the current active set is at a local minimum, this rigidity can impede the effectiveness of subsequent greedy additions. Thus, balanced settings are preferred, as they maintain the flexibility required for the model to navigate the landscape efficiently and achieve lower final errors. Upon convergence, \LieSolver yields an interpretable symbolic expression for the solution.

\subsection{Scope of \textsc{LieSolver}}
\label{ssec:liesolver_scope}
The prevailing ML approach for IBVPs, the PINN framework, optimizes two often competing objectives: the PDE residual and the initial-boundary data fit. This means a lower total loss does not guarantee higher solution accuracy \cite{krishnapriyan2021characterizingpossiblefailuremodes}. Furthermore, balancing these terms often requires complex re-weighting schemes or hard-constraint architectures \cite{Lu2021}. 

\textsc{LieSolver} circumvents this by enforcing the PDE as a hard constraint within the model structure, reducing the optimization to supervised regression on the initial-boundary data. For well-posed IBVPs, the ratio of domain error to boundary error is upper bounded. This guarantees \emph{loss alignment}: minimizing the boundary residuals provably minimizes the global error. Consequently, the boundary loss serves as an intrinsic reliability metric that enables rigorous error estimation (see details in \cref{app-ssec:IBVPs}).

\Cref{tab:method_comparison} contrasts \textsc{LieSolver} with NNs, PINNs, and classical numerical methods, such as finite difference, finite volume, and finite element methods (FDM, FVM, FEM). Unlike classical methods, \textsc{LieSolver} is mesh-free, mitigates the curse of dimensionality, and allows for seamless fusion of sparse observational data. Compared to PINNs, our method yields a model with fewer parameters, expressed as weighted sums of analytical brick functions, offering symbolic interpretability.

The primary limitation of \textsc{LieSolver} is its restriction to linear homogeneous PDEs. While this class covers many fundamental equations in physics and engineering, extending the method to nonlinear systems remains future work. Additionally, the current pipeline requires the heuristic selection of brick families. While the computation of Lie symmetries can be automated \cite{Hereman1996}, automating the optimal brick composition is an open challenge.

\section{Experiments}
\label{sec:experiment}

\subsection{Experimental setup}
\label{ssec:exp_setup}

\paragraph{Studied IBVPs}

To evaluate the performance and demonstrate the capabilities of \textsc{LieSolver}, we study various IBVPs based on the heat equation $f_t-f_{xx}=0$ and the wave equation $f_{tt} - f_{xx} = 0$. To assess robustness, we select five initial condition (IC) profiles for each PDE, varying in regularity from smooth periodic functions to sharp, discontinuous steps (see \cref{app-ssec:ibc_data} for details). We evaluate the models on the domain $(x,t) \in (0,1)\times(0,0.1)$ for the heat equation and $(0,1)\times(0,1)$ for the wave equation. 
We set the boundary conditions (BCs) to constant values consistent with the ICs. For the wave equation, we additionally enforce zero initial velocity. We discretize these constraints using $1000$--$1500$ collocation points per boundary. %

\paragraph{Brick Solutions}
\textsc{LieSolver} requires \new[an educated choice]{a preselected set} of parametrized brick solutions $f(x;\boldsymbol\vartheta)$ as discussed in \cref{ssec:liesolver_pipeline}. For the heat equation, we use the start solutions $f_{\text{start},1}(x,t)=1$ and $f_{\text{start},2}(x,t)=e^{-t}\sin x$, and construct three complementary brick families generated by the Lie symmetries \cref{eq:heat1d_trafo1}--\cref{eq:heat1d_trafo6}:
\begin{itemize}[itemsep=1pt,topsep=0pt]
  \item $f^{1}(x,t;\boldsymbol\vartheta^1)=T^4_{\vartheta_2}\!\cdot T^1_{\vartheta_1} \circ \big(e^{-t}\sin x\big)$: \new[Fourier]{sine} mode with variable phase and frequency.
  \item $f^{2}(x,t;\boldsymbol\vartheta^2)=T^1_{\vartheta_2}\!\cdot T^6_{\vartheta_1} \circ (1)$: diffusion-generated Gaussian pattern with variable centre and width.
  \item $f^{3}(x,t;\boldsymbol\vartheta^3)=T^1_{\vartheta_4}\!\cdot T^6_{\vartheta_3}\!\cdot T^4_{\vartheta_2}\!\cdot T^1_{\vartheta_1} \circ \big(e^{-t}\sin x\big)$: Gaussian pattern modulated by a sine wave with four variable parameters.
\end{itemize}

For the wave equation, even a subgroup of the full Lie symmetry group comprising spatial translation $T^1_\vartheta: f(x,t) \mapsto f(x - \vartheta, t)$ and spatiotemporal scaling $T^2_\vartheta: f(x,t) \mapsto f(e^{\vartheta} x, e^{\vartheta} t)$ yields highly expressive solutions. Furthermore, we enforce the zero initial velocity condition by requiring $\partial_t f|_{t=0}=0$ in the start solutions, ensuring that every brick satisfies this constraint by construction:
\begin{itemize}[itemsep=1pt,topsep=0pt]
  \item $f^{1}(x,t;\boldsymbol\vartheta^1)=T^2_{\vartheta_2}\!\cdot T^1_{\vartheta_1} \circ \big(\sin x \cos t \big)$: standing wave with variable phase and frequency. 
  \item $f^{2}(x,t;\boldsymbol\vartheta^2)=T^1_{\vartheta_2}\!\cdot T^2_{\vartheta_1} \circ (e^{-(x-t)^2} + e^{-(x+t)^2})$: Gaussian pattern with variable centre and width. The condition of zero initial velocity leads to a counter-propagating pair of travelling waves rather than a single pulse.
\end{itemize}

\paragraph{Training configuration}
We evaluate model performance using the mean squared error (MSE) on the test IBC data and the Relative $L^2$ error (L2RE) on the domain $\Omega$. The L2RE is computed against analytical reference solutions $f_\text{ref}$ derived via Fourier series expansion (see \cref{app-ssec:reference_solution}). Since all target values are of order unity, the MSEs are directly comparable across instances. For the heat equation, we use an empirically determined configuration: $P=1000$, $R=5$, \texttt{nfev} $=4$, $\lambda=10^{-1}$, and $L = 3000$, with $\texttt{mse\_tol} = 10^{-6}$. For the wave equation, $L = 5000$ and we apply partial parameter refinement every $5$ brick additions and global every $20$. We benchmark against a standard PINN trained with a hybrid Adam/L-BFGS optimizer to minimize a composite loss of PDE residuals and weighted IBC errors (see details in \cref{app-ssec:pinn_setup}).

\subsection{Experimental Results}
\label{ssec:results}

\Cref{tab:liesolver_vs_pinn} summarizes the performance of \textsc{LieSolver} and PINNs for the considered IBC cases. Reported medians are computed over $5$ independent seeds. For all ICs except the \IC{step} IC, \textsc{LieSolver} reliably attains IBC MSE values of order $10^{-6}$ using between $5$ and $40$ bricks. The \IC{step} IC requires more bricks due to its sharp jump. PINNs match \textsc{LieSolver} in terms of attained MSE only on the easiest cases, but fall short on the more challenging ICs.

\textsc{LieSolver} is about one to three orders of magnitude faster than PINNs across the considered cases. However, its runtime\footnote{All experiments were run on an Intel Core Ultra 7 165H CPU with 11 cores and 16\,GiB RAM.\label{fn:compute}} increases sharply as the number of bricks grows. This growth is dominated by the global parameter refinement stage. More targeted refinement strategies could mitigate this bottleneck. 

The \new[choice of brick solutions]{preselected brick families} strongly influence performance. Employing sine and Gaussian bricks, \textsc{LieSolver} naturally fits IC profiles with similar sine and Gaussian patterns and reaches the target MSE in less than a minute. The \IC{step} IC remains a challenging stress test and highlights model capability. In the following, we discuss two representative cases that illustrate \textsc{LieSolver} behaviour. The full results can be found in \cref{app-sec:results}.

\def\myrulewidth{.1pt}
\newcommand{\myrule}{\specialrule{.1pt}{1pt}{1pt}}

\begin{table}
\centering
\caption{Comparison of \LieSolver and PINN performance. `Params' denotes the network architecture for PINNs and the number of nonlinear parameters for \textsc{LieSolver}. Reported values correspond to the median across $5$ independent seeds. Standard deviations are omitted for clarity, as the performance gap is stable across all trials.}
\label{tab:liesolver_vs_pinn}
\scriptsize
\setlength{\tabcolsep}{3pt} %
\renewcommand{\arraystretch}{1.15} %
\begin{tabular}{@{} p{.18\linewidth} p{.13\linewidth} p{.15\linewidth} >{\centering}p{.20\linewidth} >{\centering}p{.12\linewidth} >{\centering\arraybackslash}p{.1\linewidth}@{}}
\toprule
\textbf{Case} & \textbf{Method} & \textbf{MSE (IBC)} & \textbf{L2RE (Domain)} & \textbf{Time\textsuperscript{\ref{fn:compute}} [s]} & \textbf{Params} \\
\toprule
\multirow{2}{*}{Heat \IC{poly}}
 & \LieSolver & $\mathbf{8.2\cdot10^{-7}}$ & $\mathbf{4.4\cdot10^{-4}}$ & \bfseries 34 & 86 \\
 & PINN & $9.0\cdot10^{-7}$ & $4.8\cdot10^{-4}$ & 137 & $4\times50$ \\
\myrule
\multirow{2}{*}{Heat \IC{Gauss}}
 & \LieSolver & $\mathbf{2.4\cdot10^{-8}}$ & $\mathbf{2.0\cdot10^{-4}}$ & \bfseries11 & 40 \\
 & PINN & $3.2\cdot10^{-7}$ &  $4.9\cdot10^{-4}$ & 123 & $4\times50$ \\
\myrule
\multirow{2}{*}{Heat \IC{Sine}}
 & \LieSolver & $\mathbf{2.4\cdot10^{-8}}$ & $\mathbf{4.6\cdot10^{-4}}$ & \bfseries 4 & 12 \\
 & PINN & $7.8\cdot10^{-7}$ & $5.0\cdot10^{-3}$ & 183 & $4\times50$ \\
\myrule
\multirow{2}{*}{Heat \IC{Sine Mix}}
 & \LieSolver & $\mathbf{2.0\cdot10^{-8}}$ & $\mathbf{3.5\cdot10^{-4}}$ & \bfseries 10 & 38 \\
 & PINN & $4.6\cdot10^{-6}$ & $1.9\cdot10^{-2}$ & 1716 & $6\times100$ \\
\myrule
\multirow{2}{*}{Heat \IC{Step}}
 & \LieSolver & $\mathbf{9.1\cdot10^{-7}}$ & $\mathbf{1.3\cdot10^{-3}}$ & \bfseries 310 & 230 \\
 & PINN & $1.8\cdot10^{-3}$ & $2.2\cdot10^{-1}$ & 440 & $6\times100$ \\
\toprule
\multirow{2}{*}{Wave \IC{Gauss}}
 & \LieSolver & $\mathbf{4.9\cdot10^{-8}}$ & $\mathbf{1.1\cdot10^{-3}}$ & \bfseries 8 & 30 \\
 & PINN & $1.5\cdot10^{-7}$ & $1.6\cdot10^{-3}$ & 1119 & $4\times50$ \\
\myrule
\multirow{2}{*}{Wave \IC{Gauss Mix}}
 & \LieSolver & $\mathbf{7.8\cdot10^{-7}}$ & $\mathbf{5.0\cdot10^{-3}}$ & \bfseries 41 & 82 \\
 & PINN & $2.7\cdot10^{-6}$ & $8.2\cdot10^{-3}$ & 2456 & $4\times50$ \\
\myrule
\multirow{2}{*}{Wave \IC{Sine}}
 & \LieSolver & $\mathbf{1.1\cdot10^{-7}}$ & $\mathbf{8.1\cdot10^{-4}}$ & \bfseries 5 & 18 \\
 & PINN & $2.0\cdot10^{-7}$ & $8.5\cdot10^{-4}$ & 1172 & $4\times50$ \\
\myrule
\multirow{2}{*}{Wave \IC{Sine Mix}}
 & \LieSolver & $\mathbf{4.6\cdot10^{-7}}$ & $\mathbf{1.7\cdot10^{-3}}$ & \bfseries 4 & 14 \\
 & PINN & $8.2\cdot10^{-5}$ & $1.8\cdot10^{-2}$ & 11008 & $6\times100$ \\
\myrule
\multirow{2}{*}{Wave \IC{Step}}
 & \LieSolver & $\mathbf{9.4\cdot10^{-7}}$ & $\mathbf{4.0\cdot10^{-3}}$ & \bfseries 217 & 182 \\
 & PINN & $7.2\cdot10^{-5}$ & $4.5\cdot10^{-2}$ & 909 & $4\times50$ \\
\bottomrule
\end{tabular}
\end{table}

\begin{figure}
    \centering
    \begin{subfigure}{.515\linewidth}
        \centering
        \includegraphics[width=\linewidth]{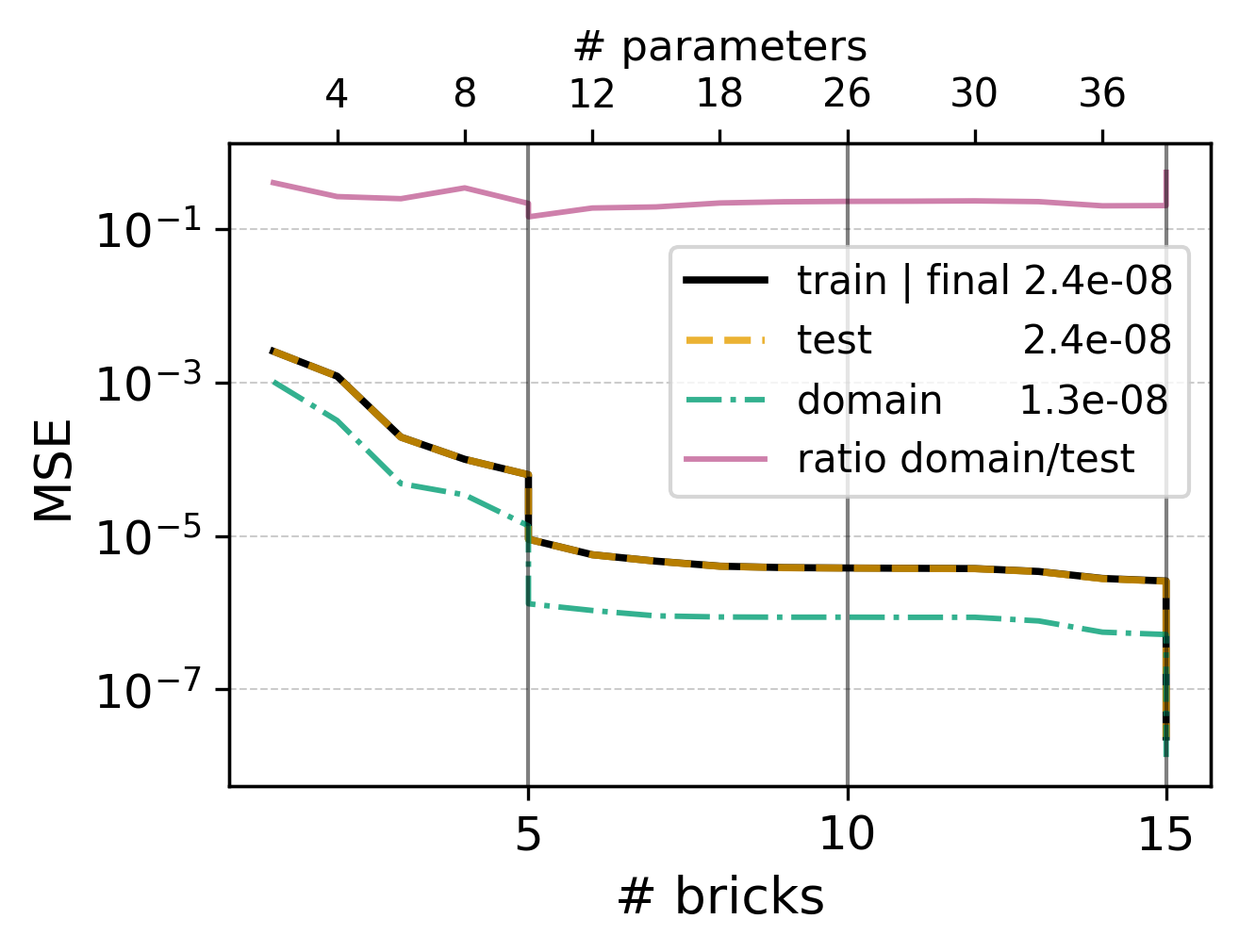}
        \caption{Heat PDE with \IC{gauss} IC}
        \label{fig:liesolver_fit_heat_gauss}
    \end{subfigure} %
    \begin{subfigure}{.475\linewidth}
        \centering
        \includegraphics[width=\linewidth]{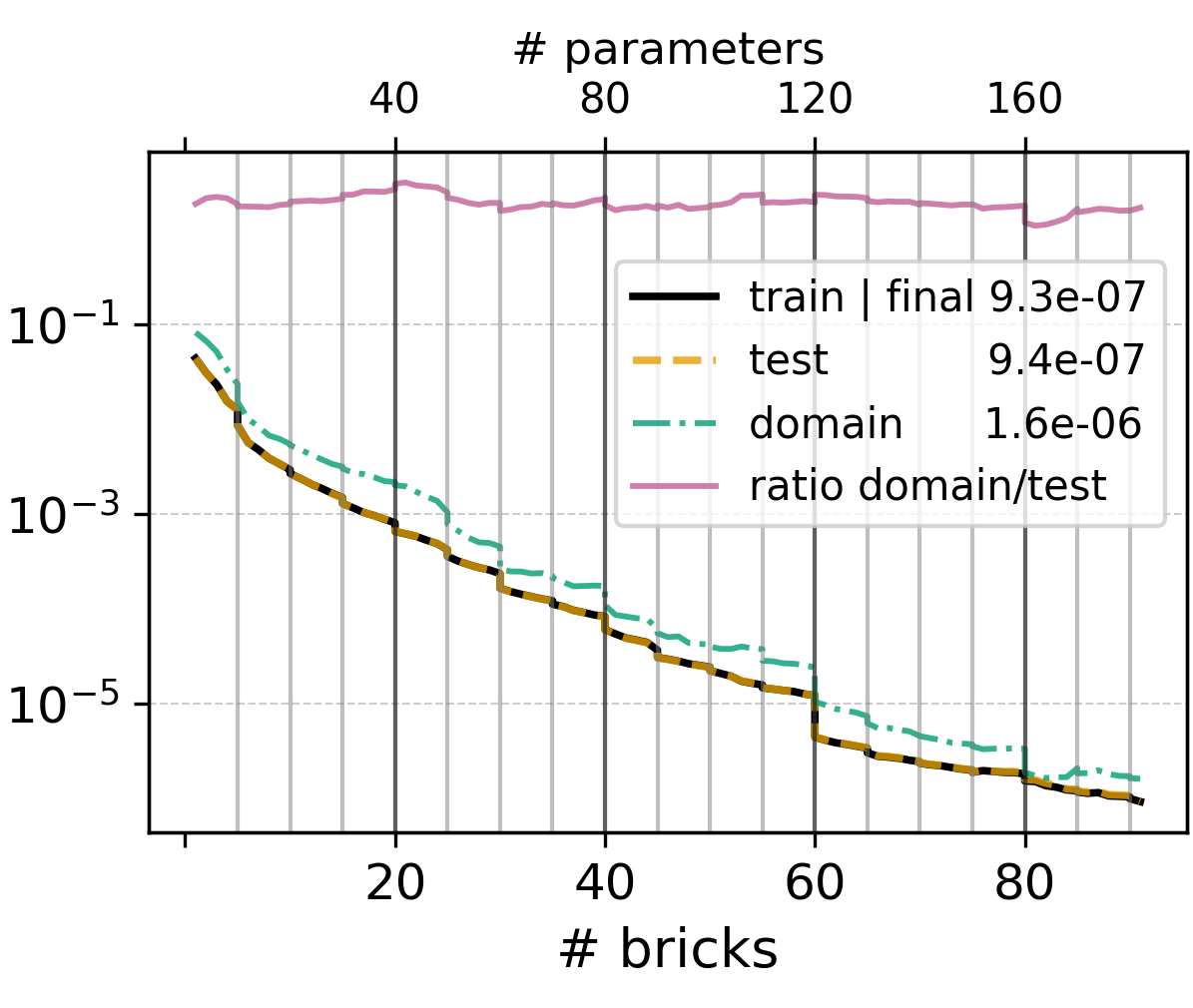}
        \caption{Wave PDE with \IC{step} IC}
        \label{fig:liesolver_fit_step_wave}
    \end{subfigure}
        \caption{Fit progress of \textsc{LieSolver}: metrics are shown as functions of the number of added bricks. The black vertical lines correspond to the global refinement steps, grey lines correspond to partial refinements of the last 5 added bricks.}
    \label{fig:liesolver_fit}
\end{figure}

\paragraph{Heat Equation with \IC{gauss} IC}
\Cref{fig:liesolver_fit_heat_gauss} shows the evolution of metrics as a function of the number of added bricks. The greedy addition begins with a Gaussian brick $f^2$ that already captures most of the IC (MSE $\sim10^{-3}$), followed by several \new[Fourier-like]{oscillatory} additions ($f^1$ and $f^3$) that enforce the Dirichlet BCs. The first global refinement stage after $5$ bricks yields a one-order-of-magnitude drop in MSE; the target tolerance ($<10^{-6}$) is met after $15$ bricks. Details on performance metrics and runtime are provided in \cref{tab:liesolver_vs_pinn}. Moreover, we observe that the ratio between the domain MSE and the IBC test MSE remains bounded throughout training, consistent with the theoretical expectations in \cref{app-ssec:IBVPs}. Thus, the IBC test MSE serves as a reliable measure for the domain MSE.

\begin{figure}
    \centering
    \includegraphics[width=\linewidth]{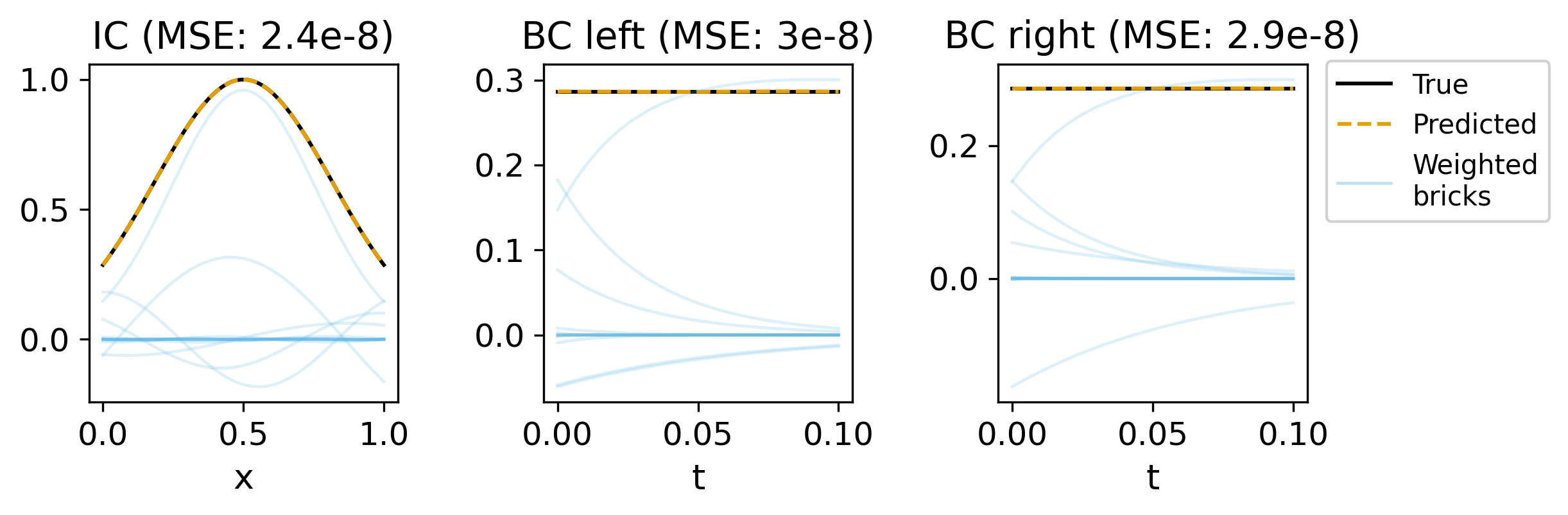}
    \caption{IBC fits for the heat equation with \IC{gauss} IC, including the decomposition into weighted brick contributions. The transparent blue curves depict the individual contributions.}
    \label{fig:liesolver_icbc_decomp_gauss}
\end{figure}

\begin{figure}
    \centering
    \includegraphics[width=0.8\linewidth]{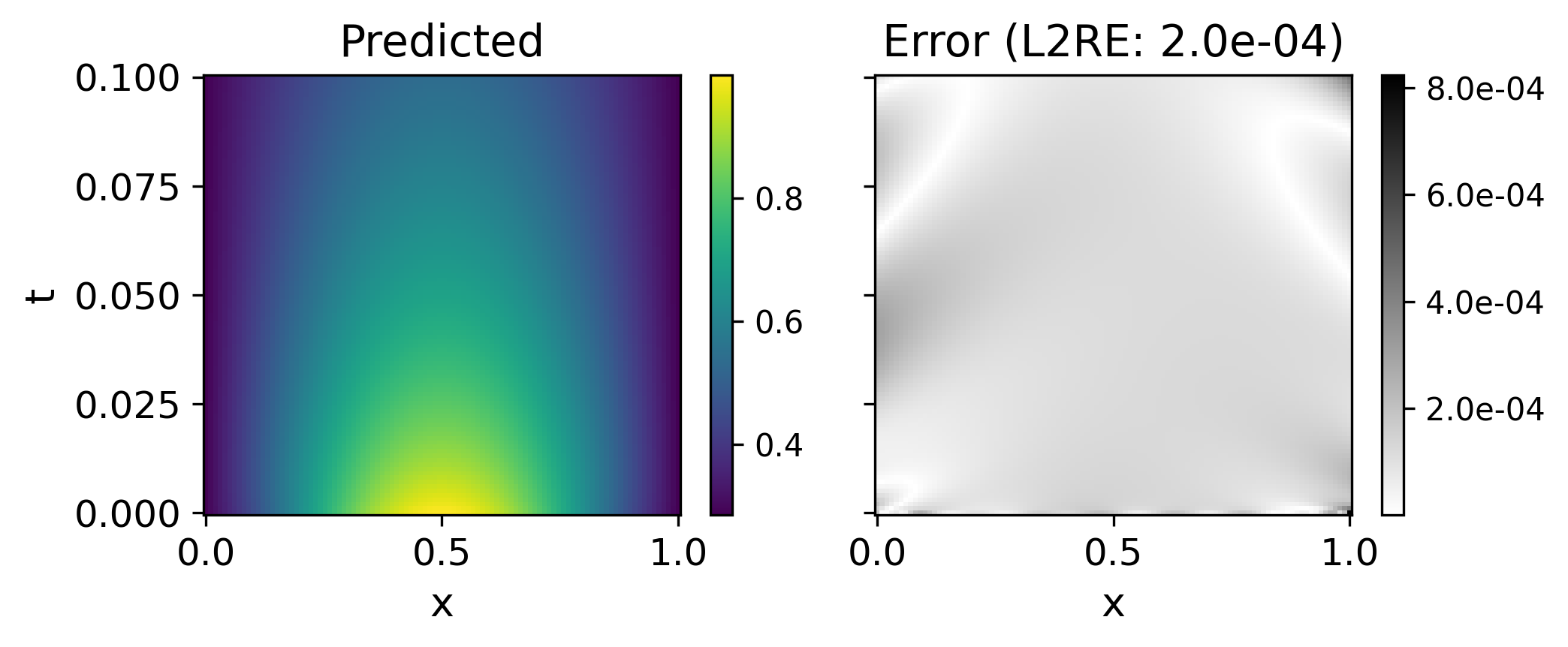}
    \caption{2D domain fields (prediction and error) for the heat equation with \IC{gauss} IC.}
    \label{fig:liesolver_domain_gauss}
\end{figure}

\Cref{fig:liesolver_domain_gauss,fig:liesolver_icbc_decomp_gauss} compare the model prediction with the ground truth on the IBCs and in the domain, respectively. The IBC plots include a decomposition into the weighted bricks, revealing dominant Gaussian-like contributors for the IC profile and additional bricks that fit the BCs.
For reference, a vanilla PINN also fits this smooth case to an IBC MSE  $<10^{-6}$, albeit at a longer runtime, as indicated in \cref{tab:liesolver_vs_pinn}.

\paragraph{Wave Equation with \IC{step} IC}
The \IC{step} IC presents the most challenging case due to the jump in the profile. \Cref{fig:liesolver_fit_step_wave} highlights two optimization regimes: (i) a gradual decrease in MSE where greedy initialization with refinement brings limited improvement, and (ii) rapid progress once sufficient expressivity is reached. Notably, a global refinement at $60$ bricks yields a pronounced drop in error. Furthermore, we again observe that the domain error tracks the IBC error throughout the training.

\begin{figure}
    \centering
    \includegraphics[width=\linewidth]{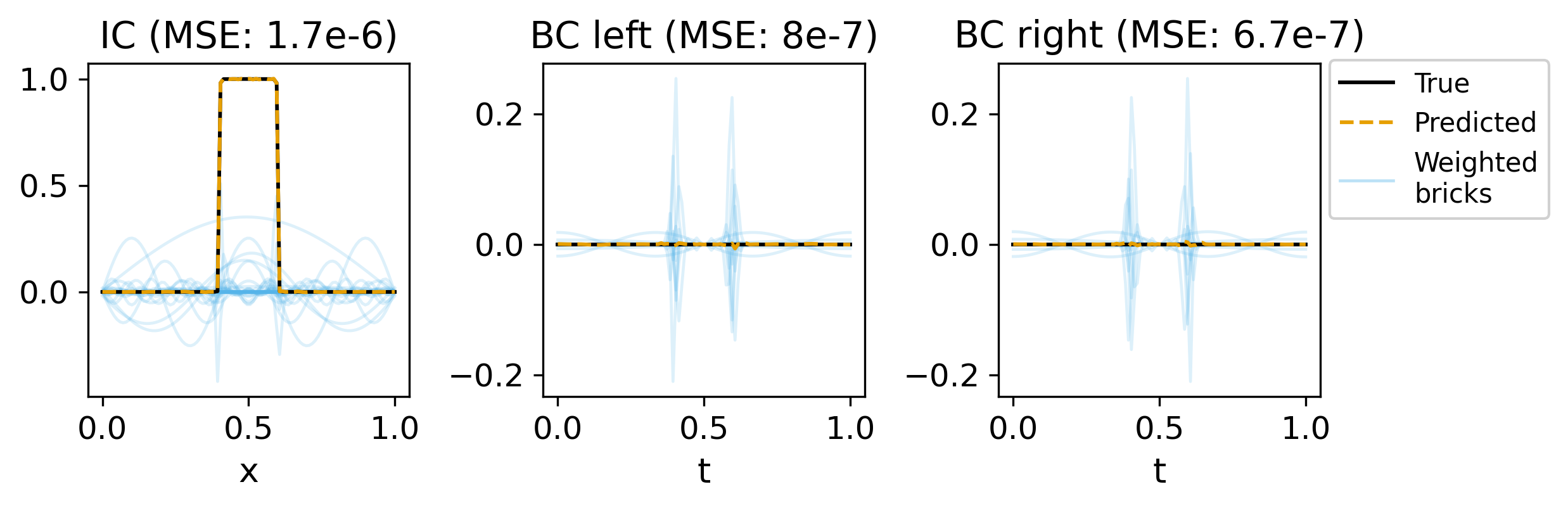}
    \caption{IBC fits for the wave equation with \IC{step} IC, including the decomposition into weighted brick contributions. The transparent blue curves depict the individual contributions.}
    \label{fig:liesolver_icbc_decomp_step_wave}
\end{figure}

\begin{figure}
    \centering
    \includegraphics[width=0.8\linewidth]{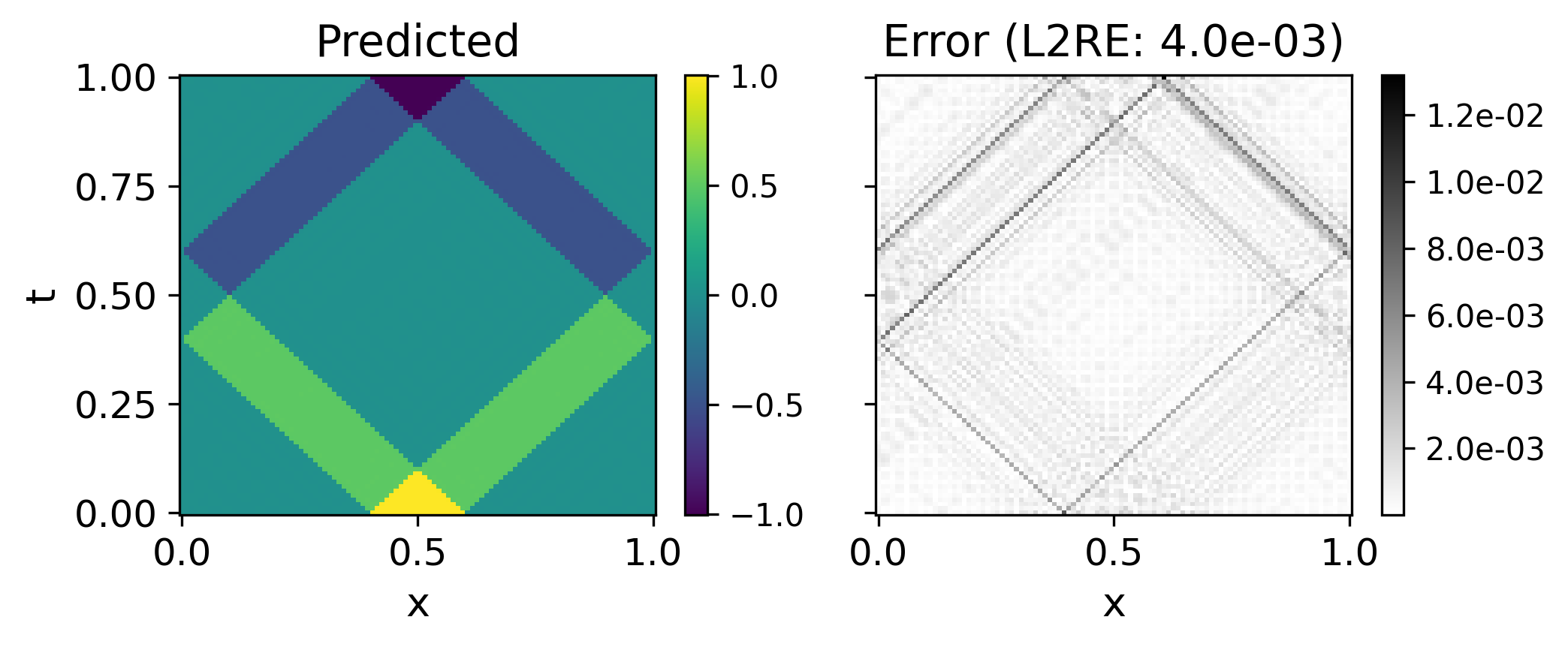}
    \caption{2D domain fields (prediction and error) for the wave equation with \IC{step} IC.}
    \label{fig:liesolver_domain_step_wave}
\end{figure}

\Cref{fig:liesolver_icbc_decomp_step_wave} shows that the solution is primarily composed of \new[Fourier sine]{standing-wave} modes with localized Gaussian corrections \new{around the discontinuity}. During optimization, the greedy stage first selects sine bricks to fit the dominant high-amplitude oscillatory component. As the remaining residual becomes localized and irregular, the algorithm increasingly adds Gaussian bricks to correct for edges and boundaries. This localization is evident in \Cref{fig:liesolver_domain_step_wave}, where the error density is concentrated along the propagating wavefronts ($x \pm t$). PINNs do not resolve the sharp feature sufficiently well and achieve only an L2RE of $4.5\cdot10^{-2}$ in the domain (see \cref{app-ssec:pinn_results}). Overall, the results demonstrate that \textsc{LieSolver} consistently achieves higher accuracy and significantly faster convergence than PINNs, particularly in cases involving complex or sharp initial conditions.

\section{Conclusion}

Using Lie symmetries, we generate parametrized functions that exactly solve the corresponding PDE. By incorporating the underlying symmetry structures directly into the model, solving IBVPs reduces to optimizing these parameters to satisfy the initial-boundary conditions (IBCs). Focusing on the common case of linear homogeneous PDEs, we introduce \textsc{LieSolver}, which represents the solution as a linear combination of Lie-symmetry-generated brick solutions whose linear amplitudes and nonlinear symmetry parameters are learned from the IBCs. \textsc{LieSolver} constructs this combination incrementally, alternating between greedily initializing bricks best aligned with the residual and refining parameters. 
By incorporating the PDE into the model's structure, \LieSolver offers several advantages: the loss function effectively serves as a faithful indicator of predictive accuracy, so that --- unlike in PINNs --- a decrease in the objective on the boundary directly corresponds to an improvement in accuracy on the domain. In addition, this property enables rigorous error bounds for well-posed IBVPs, in contrast to PINNs, where such guarantees are generally missing and are an active field of research. Moreover, these tailor-made models use comparatively few parameters, which results in significant computational speedups compared to PINNs without sacrificing expressivity. Furthermore, \LieSolver expresses its prediction as a sparse, analytical expression, thereby enhancing interpretability.

Across all benchmarks on various IBVPs, \textsc{LieSolver} consistently outperformed PINNs in both accuracy and computational efficiency, running one to three orders of magnitude faster. For smooth ICs, \textsc{LieSolver} achieved an IBC MSE below $10^{-6}$ with fewer than $100$ nonlinear parameters. For the harder step-like ICs, more parameters were required, yet \textsc{LieSolver} still obtained a domain L2RE on the order of $10^{-3}$, whereas PINNs plateaued between $10^{-2}$ and $10^{-1}$. 
In summary, \textsc{LieSolver} delivers substantially higher accuracy at a fraction of the computational cost, establishing it as a highly effective alternative to PINNs for solving IBVPs.

\paragraph{Limitations}
Despite the notable advantages of using Lie symmetries, it is important to acknowledge the limitations of our approach. Currently, \textsc{LieSolver} applies only to linear homogeneous PDEs. Although this encompasses a large class of important problems, we aim to generalize our framework in future work. \new[Another drawback is that for every PDE, appropriate bricks --- composed from Lie symmetries --- and at least one start solution must be selected. Still, Lie symmetries can be determined algorithmically; for the most standard problems, they can also be found in the literature. 
A further weakness is that the performance of \textsc{LieSolver} heavily depends on the chosen set of brick solutions, an aspect that requires further investigation to improve automation.]{Another requirement is that for every PDE, appropriate bricks --- composed from Lie symmetries --- and at least one start solution must be selected. We address this through a systematic preselection procedure (\cref{ssec:liesolver_pipeline}), which identifies expressive bricks. Lie symmetries themselves can be determined algorithmically \cite{Hereman1996} or found in the literature for standard problems \cite{olver1993applications}.}
However, as is the case in conventional NNs, performance highly depends on various design choices. Nonetheless, in our approach, the loss provides a rigorous criterion for assessing whether a particular choice of brick solution is effective or needs to be extended, providing useful criteria for these design choices.%
It is also important to note that we have tested \textsc{LieSolver} only on a limited number of problems. Therefore, more extensive studies will be necessary in future work. 

These findings suggest several promising directions: \new[a more detailed analysis of \textsc{LieSolver} on various problems, an automated selection of brick solutions, as well as an extension to more general classes of PDEs. For the latter task, genetic algorithms appear to be a promising direction.]{broader empirical evaluation, complete automation from symmetry derivation to brick preselection, and extension to general PDEs. The extension to inhomogeneous linear PDEs is straightforward: the difference of any two solutions of the inhomogeneous PDE is a solution of the homogeneous PDE. Hence, by adding an arbitrary particular solution to \cref{eq:liesolver_structure}, one can represent the full inhomogeneous solution space. For nonlinear PDEs, Lie symmetry theory still applies, but superposition no longer holds in general. Here, genetic algorithms and evolutionary approaches appear to be promising directions, since the lack of superposition removes the linear factorization our approach currently exploits.} By pursuing these steps, we aim to enhance the capabilities and applicability of \textsc{LieSolver} and provide an interpretable, reliable, and efficient IBVP solver for general PDEs.

\section*{Code availability}
The Python implementation is available at \url{https://github.com/oduwancheekee/liesolver}.

\section*{Impact Statement}

This work aims to advance the field of scientific machine learning by proposing an alternative method for solving IBVPs. We do not foresee specific societal consequences beyond those already associated with existing PDE solvers.

\newpage
\appendix
\crefalias{section}{appendix}
\crefalias{subsection}{appendix}
\onecolumn

\section{Extended Discussion of Initial-Boundary Value Problems}
\label{app-sec:PDEs_IBVPs}

Initial-boundary value problems (IBVPs) are essential for modelling various phenomena in physics and many other quantitative sciences. Since they are the main object of study in this article, we provide some useful background information. In contrast to the summary in \cref{ssec:IBVPs}, we will introduce partial differential equations (\cref{app-ssec:PDEs}) and IBVPs (\cref{app-ssec:IBVPs}) in a way, which is most convenient to discuss their symmetries in \cref{app-sec:LieSymmetries}. We split this overview in the two components of IBVPs: partial differential equations and initial-boundary conditions (IBCs).

\subsection{Partial Differential Equation (PDEs)}
\label{app-ssec:PDEs}

In a nutshell, PDEs are just equations containing derivatives. Thus, we can consider them for example as differential operators acting on functions, as done in \cref{ssec:IBVPs}. However, when studying symmetries of PDEs, it is often convenient to think of their solutions as parametrized surfaces, instead of functions, which enables a geometric treatment. In the following, we develop this idea rigorously and introduce the key concepts. This summary is oriented towards the description in \cite{olver1993applications}. For a more general perspective on PDEs we recommend \cite{Rauch1991, John1978, evans2010partial, Olver2013}.
 
Consider $n$ \emph{independent variables} $x=(x_1,\ldots,x_n)\in X=\mathbb{R}^n$ and $m$ \emph{dependent variables} $u=(u^1,\ldots,u^m)\in \mathcal{U} = \mathbb{R}^m$. This terminology reflects that we will later treat solutions as maps from $X$ to $\mathcal U$. In addition to the dependent and independent variables, the PDE also contains derivatives. Therefore, we denote by $d\in\mathbb N$ the \emph{order} of the considered PDE, i.e., the order of its highest derivative. For $n$ independent variables, the number of distinct partial derivatives up to order $d$ is given\footnote{We always assume sufficiently smooth functions, such that derivatives will commute. The number of partial derivatives follows from the combinatorial identity $\sum_{i=0}^d \binom{n+i-1}{i} = \binom{n+d}{d}$ for $d \geq 0$ and $n > 0$, which can be easily shown by induction.} by $\binom{n+d}{d}$. We \emph{prolong} the space $\mathcal U$ to include all partial derivatives up to order $d$, which we denote by $\mathcal{U}^{(d)} := \mathbb{R}^{m \binom{n+d}{d}}$. By these definitions, a \emph{system of $p$ PDEs of order $d$} is given as a set of equations
\begin{align}
    \mathcal D_j\big(x,u^{(d)}\big) = 0, \qquad j=1,\ldots ,p \label{eq:system_of_pdes}
\end{align}
where we assume $\mathcal D : X\times\mathcal U^{(d)}\to\mathbb R^p$ to be smooth. Accordingly, we describe PDEs as functions of the independent variables and the prolonged dependent variables. 

In order to connect the placeholder variable $u^{(d)}$ to the actual derivatives, we introduce the \emph{prolongation} of functions. For a sufficiently smooth function $f:\Omega\to\mathcal U$ with $\Omega\subseteq X$ open, its $d$-th prolongation $\pr f:\Omega\to \mathcal U^{(d)}$ is defined as the tuple
\begin{align}
    \pr\! f (x) := \Big(\partial_J f(x)\Big)_{J\in[n]^d} \in \mathcal U^{(d)} \comma
\end{align}
where $[n]^d := \big\{\sigma\subseteq\{1,\ldots,n\} : |\sigma|\leq d\big\}$ contains all subsets of $[n]$ with at most $d$ elements (including the empty set), and we write for a set $J=\{j_1,\ldots,j_k\}$
\begin{align}
    \partial_j := \frac{\partial}{\partial x_j}, \qquad \partial_{\{j_1,\ldots,j_k\}} := \partial_{j_1} \cdots\partial_{j_k} \quad\text{and}\quad \partial_\emptyset f := f
\end{align}
for brevity. Thus, the prolongation explicitly lists all derivatives up to order $d$. For instance, for a function $f:\mathbb{R}^2\to\mathbb{R}$ we get $\pr[2]f(x,y) = \bigl(f, f_x, f_y, $ $f_{xx}, f_{xy}, f_{yy}\bigr) \in \mathcal U^{(2)} \cong \mathbb{R}^6$.

A sufficiently smooth function $f:\Omega\to \mathcal U$, with $\Omega\subseteq X$ open, is called a \emph{solution of the PDE system} $\mathcal D$ if
\begin{align}
    \mathcal D(x,\pr\! f(x)) = 0 \qquad \forall x\in \Omega \point
\end{align}

It is important to note, that there is no general theory to find solutions of PDEs. However, for the further treatment, it is useful to classify PDEs with respect to their properties. We call a PDE (system) $\mathcal D$ \emph{linear} if we can write it in the form
\begin{align}
    \mathcal D(x,\pr f(x)) = h(x) + \sum_{J\in [n]^d} a_J(x) \partial_J f(x)
    \label{eq:linearPDE}
\end{align}
for some functions $h, a_J:X\to\mathbb{R}^p$. If in addition $h(x) \equiv 0$, we call the PDE (system) \emph{linear homogeneous}. For any linear homogeneous PDEs, we have that if $f_1$ and $f_2$ are two solutions of the PDE, then so is $\alpha f_1 + \beta f_2$ for any $\alpha,\beta\in\mathbb R$. In particular, it holds that $f(x)\equiv 0$ is always a solution for any linear homogeneous PDE. If a linear homogeneous PDE does not contain a term depending on $u$, it additionally is true that every constant function solves this PDE.

\subsection{Initial-Boundary Conditions and Well-Posed IBVPs}
\label{app-ssec:IBVPs}

Since PDEs usually admit infinitely many solutions, we impose additional constraints in the form of \emph{initial-boundary conditions} (IBC) to ensure uniqueness. To simplify the notation, we will not explicitly distinguish between time and spatial coordinates. Hence, initial and boundary conditions get the same form. In accordance with \cref{ssec:IBVPs}, let $\Omega\subset X$ be a bounded domain, where its boundary $\partial\Omega$ contains smooth, connected components $\Gamma_1,\ldots,\Gamma_r\subseteq\partial\Omega$. Analogously to PDEs, we can represent IBCs either by operators or by equations. For a consistent notation with \cref{app-ssec:PDEs}, which is also more oriented towards the way a computer handles these conditions, we describe them also by equations. Thus, for each $\Gamma_k$, we introduce a condition of the form
\begin{align}
    \mathcal B_k\!\left(x,u,u_{\boldsymbol n}\right) = 0 
    \quad \text{for all}\ x\in\Gamma_k
\end{align}
with $\mathcal B_k:\Gamma_k\times\mathcal U\times\mathcal U\to \mathbb R$. The symbol $u_{\boldsymbol n}$ is a placeholder, which will later be associated with the normal derivative. A PDE system $\mathcal D$ together with associated IBCs $\mathcal B_1,\ldots,\mathcal B_r$ defines an \emph{initial-boundary value problem} (IBVP). A sufficiently smooth function $f:\overline\Omega\to\mathcal U$ is called a \emph{solution of the IBVP} if
\begin{align}
    &\mathcal D(x, \pr f(x)) = 0 \qquad\forall x\in\Omega \\
    &\mathcal B_k \Big(x, f(x) ,\od{f}{\boldsymbol n} (x) \Big) = 0 
    \qquad \forall x\in\Gamma_k,\ k=1,\ldots,r \comma
\end{align}
where $\od{f}{\boldsymbol n} (x) = \nabla f(x) \cdot \hat n(x)$ is the normal outward derivative on $\Gamma_k$.

In this work, we always focus on \emph{well-posed} IBVPs, meaning that 
\begin{enumerate}[label=(\roman*), itemsep=1pt, topsep=0pt]
    \item the IBVP has a solution, 
    \item the solution is unique, and 
    \item the solution depends continuously on the prescribed initial and boundary data. 
\end{enumerate}
Problems not satisfying (i) and (ii) are not expected to appear in the modelling of a concrete physical phenomenon. However, there are some nonlinear problems, which are not satisfying (iii), often referred to as chaotic systems. Due to their numeric instability, they are inaccessible by classical numerical methods. The precise definition of these three properties depends on the specific problem being considered, and there is no commonly agreed definition, as noted in \cite{John1978}. 

In order to clarify the precise meaning of (i) and (ii), we typically specify the space of possible solutions to smooth functions and may add further restrictions, e.g., certain assumptions on the growth at infinity, in order to specify uniqueness. When formulating in metric spaces, the third condition (iii) means that the map from IBCs to solutions is (locally) Lipschitz continuous. To illustrate the last condition in practice, we give the following example.

\begin{example}[Heat Equation and Reverse Heat Equation]
    Consider the so-called reverse heat equation in the domain $x\in [0,2\pi)$, $t>0$ with vanishing boundary condition and two different initial conditions
    \begin{align}
        u_t + u_{xx} &= 0,\; u(x,0)=0,\; u(0,t) = u(2\pi,t) = 0 \\
        v_t + v_{xx} &= 0,\; v(x,0)=\epsilon \sin(x),\; v(0,t) = v(2\pi,t) = 0 \comma
    \end{align}
    with $\epsilon>0$. It is not hard to verify the two solutions $u(x,t)\equiv 0$ and $v(x,t)=\epsilon \sin(x) e^t$. These two problems only differ in the initial condition, having the distance $\norm{u(0)-v(0)}_\infty = \epsilon$, with the norm $\norm{f(t)}_\infty := \max_{x\in [0,2\pi)} |f(x,t)|$. Hence, the initial conditions are arbitrarily close for small $\epsilon$. However, the distance between solutions is given by $\norm{u(t)-v(t)}_\infty = \epsilon\ e^t$, which is not bounded on the domain. Hence, the solution will not depend continuously on the initial data, and this problem is ill-posed.

    In comparison, consider a slightly modified version of the problem, which is the usual heat equation
    \begin{align}
        u_t - u_{xx} &= 0,\; u(x,0)=0,\; u(0,t) = u(2\pi,t) = 0 \\
        v_t - v_{xx} &= 0,\; v(x,0)=\epsilon \sin(x),\; v(0,t) = v(2\pi,t) = 0 \comma
    \end{align}
    which results in the solutions $u(x,t)\equiv 0$ and $v(x,t)= \epsilon \sin(x) e^{-t}$. In that case, the distance between the solutions is given by $\norm{u(t)-v(t)}_\infty = \epsilon\ e^{-t} \leq \epsilon$. Hence, the map from initial data to solutions is (locally) Lipschitz continuous
    \begin{align}
        \norm{u(t)-v(t)}_\infty \leq C \norm{u(0)-v(0)}_\infty \label{eq:exupperbound}
    \end{align}
    with Lipschitz constant $C=1$. Hence, we confirmed the well-posedness of the heat equation with Dirichlet conditions. In other words, the ratio $\norm{u(t)-v(t)}_\infty / \norm{u(0)-v(0)}_\infty \leq C$ is bounded. 
\end{example}
The precise form of \cref{eq:exupperbound} depends naturally on the formulation of the IBVP and gets more involved, when increasing the number of conditions and variables. However, we always have that the ratio between changes in domain values and changes in initial-boundary values is bounded from above. This particular property of well-posed IBVPs, leads to a rigorous error estimation of \textsc{LieSolver}: If we know the deviation on the boundary, we can directly infer the (maximal) error in the domain.

\section{Extended Discussion of Lie Symmetries}
\label{app-sec:LieSymmetries}

We expand our summary from \cref{ssec:lie_symmetries} by presenting additional foundational concepts, followed by a detailed example illustrating the systematic derivation of Lie symmetry transformations for a given PDE. For a comprehensive exploration of these subjects, we refer to \cite{olver1993applications} as an extensive standard source, \cite{Olver1992, Bluman1989, Bluman2010} for alternative presentations, \cite{Duistermaat1999} for a focus on Lie groups, and \cite{Oliveri2010} for an overview, including a historical perspective.

\subsection{Basics on Lie Symmetries and Lie Groups}
\label{app-ssec:LieBasics}
PDEs can be invariant under certain symmetries, and so their set of solutions has to respect the same symmetries. Let us illustrate this behaviour on the simple differential equation $u_{xx}=0$. Obviously, if $u=f(x)$ is a solution, then so is $u=f(x)+\vartheta$, as well as $u=e^\vartheta f(x)$ for any $\vartheta\in\mathbb R$. Hence, the underlying PDE comes with a translation and scaling invariance
\begin{align}
    T^1_\vartheta : f(x)\mapsto f(x)+\vartheta \quad\text{and}\quad T^2_\vartheta : f(x)\mapsto e^\vartheta f(x) \point
\end{align}
Moreover, in this example, these symmetry transformations will generate a group structure, i.e., we have a neutral element $T^i_0 = e$, inversions $\big(T^i_\vartheta\big)^{-1}=T^i_{-\vartheta}$ and $T^i_{\vartheta_1}\circ T^i_{\vartheta_2} = T^i_{\vartheta_1+\vartheta_2}$. Thus, we obtain a two-dimensional group, which is spanned by two one-parameter groups.

The same structure persists for general PDEs. The rigorous study of these symmetry groups naturally leads to the framework of \emph{Lie groups} \cite{Duistermaat1999}, which provide the mathematical structure for continuous transformation groups. A Lie group is a smooth, finite-dimensional manifold endowed with a group structure\footnote{A group $(G,\cdot)$ is a set $G$ equipped with a binary operation $\cdot: G \times G \to G$ that is associative, has a unique identity element $e\in G$, and assigns to each $g \in G$ a unique inverse $g^{-1}$.} such that both the multiplication and inversion maps are smooth. 

Symmetry transformations can be described as actions of the Lie groups on manifolds. For a Lie group $G$ and a smooth manifold $M$, this action $\circ:G\times M \to M$ is required to be compatible with the group axioms, i.e., we demand $g_1 \circ (g_2 \circ x) = (g_1\cdot g_2)\circ x$ and $e\circ x = x$ for any $g_1,g_2\in G$ and all $x\in M$. 

Depending on the group $G$ and the manifold $M$, not all of these actions may be defined, which is why we allow restricting these actions to a subset $\mathscr U\subseteq G\times M$ containing the identity $\{e\}\times M \subseteq \mathscr U$. We call this restriction a \emph{local group of transformations}. Note that this restriction to group elements around the identity does not pose any serious limitations, since every element of the full Lie group $G$ can be written as a finite product of elements which are close to the identity \citep[compare][thm. 1.22 and prop. 1.24]{olver1993applications}.

In order to connect a PDE solution to a manifold, let $f : \Omega \to \mathcal{U}$ be a sufficiently smooth function with $\Omega \subseteq X$ open. Its \emph{graph} is defined as 
\begin{align}
    \Gamma_f := \{ (x, f(x)) \mid x \in \Omega \} \subseteq \Omega \times \mathcal{U} \comma
\end{align}
a smooth submanifold of $X \times \mathcal{U}$. If $G$ is a (local) group of transformations acting on $X \times \mathcal{U}$, then for $g \in G$ we define the transformed graph as  
\begin{align}
    g \circ \Gamma_f := \{ (\tilde{x}, \tilde{u}) = g \circ (x, u) \mid (x, u) \in \Gamma_f \} \point
\end{align}
In the case where we can write the transformed graph $g\circ \Gamma_f=\Gamma_{\tilde f}$ as a graph of another function $\tilde f$, we simply write $\tilde f = g\circ f$. Note that for elements $g$ close to the identity, we always find a (single-valued) function $\tilde f$ with that property.

A \emph{symmetry group of a PDE $\mathcal D$} is a local group of transformations $G$ acting on $M\subseteq X\times \mathcal U^{(d)}$, such that whenever $u=f(x)$ is a solution of $\mathcal D$, and $g\circ f$ is defined for $g\in G$, then $u=g\circ f(x)$ will be a solution of $\mathcal D$ as well. There are several constructive methods to determine the full symmetry group of a PDE, where we will discuss one option in the following section.

\subsection{Computation of Lie Symmetries}
\label{app-ssec:LieComputation}

We will close this short summary on Lie symmetries, by providing a brief overview of how to compute the symmetry group of a system. To keep the description as short as possible, we will omit many details and focus on a practical guide for calculating the symmetry group. For further details, we recommend \citep[ch. 2]{olver1993applications}, which the following outline is oriented towards. Before stating the so-called infinitesimal criterion, which is the key concept for the determination of symmetry groups, we have to introduce some further definitions.

Every Lie group $G$ is related to a so-called \emph{Lie algebra} $\mathfrak g$, which is defined to be the tangent space of $G$ at its identity element \citep[for further details see e.g.][]{Duistermaat1999}. Due to their vector space structure, it is often much more convenient to work with the Lie algebra $\mathfrak g$, instead of the Lie group $G$. By Lie's third theorem \citep[thm. 1.54]{olver1993applications} there is a one to one correspondence between finite-dimensional Lie algebras $\mathfrak g$ and connected Lie groups $G$. Thus, we lose no generality, when considering Lie algebras. Elements of the Lie algebras are also called the \emph{infinitesimal generators} of a Lie group.

A system of PDEs $\mathcal D(x,u^{(d)})$ is called to be of \emph{maximal rank}, if its Jacobian $J_{\mathcal D}(x,u^{(d)})$ with respect to all variables has the maximal rank $p$, whenever $\mathcal D(x,u^{(d)})=0$. E.g., $\mathcal D(x,u^{(2)})= u_t-u_{xx}$ has Jacobian $J_{\mathcal D}(x,t;u, u_x, u_t, u_{xx},u_{xt},u_{tt})=(0,0;0,0,1,-1,0,0)$, which is of maximal rank, whereas $\mathcal D(x,u^{(2)})= (u_t-u_{xx})^2$ is not of maximal rank.

\begin{theorem}[Infinitesimal criterion {\citep[for a proof see][thm. 2.31]{olver1993applications}}] \label{thm:infinitesimal_criterion1}
    Let $\mathcal D$ be a system of PDEs with maximal rank defined over $M\subseteq X\times U$ and let $G$ be a local group of transformations acting on $M$. If 
    \begin{align}
        \pr \mathbf v [\mathcal D(x,u^{(d)})] = 0, \quad\text{whenever}\;\; \mathcal D(x,u^{(d)})=0 \label{eq:infinitcriterion}
    \end{align}
    holds for any infinitesimal generator $\mathbf v$ of $G$, then $G$ is a symmetry group of $\mathcal D$.
\end{theorem}

In fact, this theorem gives us a computable criterion to determine the infinitesimal generators $\mathbf v$ of the symmetry group. To evaluate \cref{eq:infinitcriterion}, we also need to prolong a vector fields, which can be computed for a general vector field
\begin{align}
    \mathbf v = \sum_{i=1}^n \xi^i(x,u) \pd{}{x_i} + \sum_{k=1}^m \phi_k(x,u) \pd{}{u^k} \label{eq:vector-field_ansatz}
\end{align}
by \citep[thm. 2.36 and prop. 2.35]{olver1993applications}
\begin{align}
    \pr \mathbf v = \mathbf v + \sum_{k=1}^m \sum_{J\in [n]^d\setminus \emptyset} \phi_k^J(x,u^{(d)})\pd{}{u_J^k} \label{eq:prolongation1}
\end{align}
with 
\begin{align}
    \phi_k^J(x,u^{(d)})=D_J \Big(\phi_k-\sum_{i=1}^n\xi^i u_i^k \Big) + \sum_{i=1}^n \xi^i u^k_{J\cup\{i\}} \label{eq:phiJ}
\end{align}
and the total derivatives
\begin{align}
    D_{\{j_1,\ldots,j_l\}} = D_{j_1}\cdots D_{j_l} \quad\text{with}\;\; D_j = \pd{}{x_j} + \sum_{k=1}^m \sum_{J\in [n]^d} u^k_{J\cup\{j\}} \pd{}{u^k_J} \point \label{eq:totalderivative}
\end{align}

Let us illustrate \cref{thm:infinitesimal_criterion1} and these relatively lengthy formulas through a practical example.

\begin{example}[$1$d-heat equation] \label[example]{ex:heat-equation-generators}
    Consider the heat equation
    \begin{align}
        \mathcal D(x,t,u^{(2)})=u_t - u_{xx} \label{eq:ex-heateq}
    \end{align}
    with $n=2$, $m=1$, $p=1$ and $d=2$. Applying \cref{eq:prolongation1} and \cref{eq:infinitcriterion} to a generic vector field
    \begin{align}
        \mathbf v = \xi(x,t,u) \pd{}{x} + \tau(x,t,u) \pd{}{t} + \phi(x,t,u)\pd{}{u}
    \end{align}
    we will obtain the infinitesimal criterion for symmetry groups of the heat equation as
    \begin{align}
        \phi^t - \phi^{xx} = 0 \quad\text{whenever}\;\; u_t = u_{xx} \point \label{eq:ex_determiningsystem}
    \end{align}
    To resolve the functions $\phi^t$ and $\phi^{xx}$, we get by application of \cref{eq:phiJ} and \cref{eq:totalderivative}
    \begin{align}
        \phi^t &= D_t(\phi-\xi u_x-\tau u_t)+\xi u_{xt}+\tau u_{tt} = D_t\phi - u_x D_t\xi - u_t D_t\tau \nonumber\\
        &= \phi_t - \xi_t u_x + (\phi_u-\tau_t) u_t - \xi_u u_x u_t - \tau_u u_t^2 \label{eq:ex_phit}
    \end{align}
    and
    \begin{align}
        \phi^{xx} =& \ D_x^2(\phi-\xi u_x-\tau u_t) + \xi u_{xxx} + \tau u_{xxt} \nonumber\\
        =& \ D_x^2 \phi - u_x D_x^2\xi - u_t D_x^2\tau - 2 u_{xx} D_x\xi - 2u_{xt}D_x\tau \nonumber\\
        =& \ \phi_{xx} + (2\phi_{xu}-\xi_{xx}) u_x - \tau_{xx}u_t + (\phi_{uu}-2\xi_{xu})u_x^2 - 2\tau_{xu}u_x u_t - \xi_{uu}u_x^3 \nonumber\\
        &- \tau_{uu}u_x^2 u_t + (\phi_u - 2\xi_x)u_{xx}-2\tau_x u_{xt} - 3 \xi_u u_x u_{xx} - \tau_u u_t u_{xx} - 2\tau_u u_x u_{xt} \point \label{eq:ex_phixx}
    \end{align}
    Since $\phi^t=\phi^{xx}$ has to hold independently of the value of derivatives of $u$, we can compare them term-wise. When using in addition that $u_t=u_{xx}$, we get the following system of equations, which are often called the \emph{determining equations}:
    \begin{multicols}{2}
        \begin{enumerate}[label=(\roman*)]
            \item $\phi_t=\phi_{xx}$
            \item $-\xi_t = 2\phi_{xu}-\xi_{xx}$
            \item $0=\phi_{uu}-2\xi_{xu}$
            \item $0=-\xi_{uu}$
            \item $\phi_u-\tau_t=-\tau_{xx}+\phi_u-2\xi_x$
            \item $-\xi_u=-2\tau_{xu}-3\xi_u$
            \item $0=-\tau_{uu}$
            \item $-\tau_u=-\tau_u$
            \item $0=-2\tau_x$
            \item $0=-2\tau_u$
        \end{enumerate}
    \end{multicols}
    In order to find the most general infinitesimal generator $\mathbf v$ from \cref{eq:vector-field_ansatz}, we have to solve this system, to determine the functions $\xi(x,t,u), \tau(x,t,u)$ and $\phi(x,t,u)$. From (ix) and (x), we get that $\tau=\tau(t)$ depends only on $t$. In combination with (vi), $\xi$ will not depend on $u$. Hence, (v) takes the form $\tau_t=2\xi_x$, which implies $\xi(x,t) = \frac{1}{2} \tau_t x + \gamma(t)$ for some function $\gamma$. Therefore, using (iii), $\phi$ will only linearly depend on $u$, i.e., $\phi(x,t,u)=\alpha(x,t) + \beta(x,t) u$ for some functions $\alpha,\beta$. Hence, we have used every equation except (i) and (ii), which takes the form
    \begin{multicols}{2}
        \begin{enumerate}[label=(\roman*)]
            \item $\alpha_t=\alpha_{xx}$, $\beta_t=\beta_{xx}$
            \item $-\frac{1}{2} \tau_{tt}x-\gamma_t = 2\beta_x$
        \end{enumerate}
    \end{multicols}
    By integrating (ii) w.r.t. $x$, we get $\beta(x,t)=-\frac{1}{8}\tau_{tt}x^2-\frac{1}{2}\gamma_tx+\rho(t)$. Comparing with (i), we get $\tau_{ttt}=0$, $\gamma_{tt}=0$ and $\rho_t=-\frac{1}{4} \tau_{tt}$. Hence, $\tau(t) = 4c_6 t^2 + 2 c_4 t + c_2$ is a quadratic function, $\gamma(t)=2c_5t+c_1$ and $\rho(t)=-2c_6 t +c_3$ are linear functions, with $c_1,\ldots,c_6\in\mathbb R$. Thus, the general solution of the determining equations is given by
    \begin{align}
        \xi &= c_1 + c_4x + 2c_5t + 4c_6xt \\
        \tau &= c_2 + 2c_4t+4c_6t^2 \\
        \phi &= (c_3-c_5x-2c_6t-c_6x^2)u + \alpha(x,t)
    \end{align}
    where $c_1,\ldots,c_6\in\mathbb R$ and $\alpha(x,t)$ is any solution of the heat equation $\alpha_t=\alpha_{xx}$. Using the usual basis in $\mathbb R^6$, the Lie algebra is spanned by $6$ vector fields
    \begin{align}
        &\mathbf v_1 = \partial_x \qquad \mathbf v_2 = \partial_t \qquad \mathbf v_3 = u \partial_u\qquad \mathbf v_4 = x\partial_x + 2t\partial_t \nonumber\\
        &\mathbf v_5 = 2t\partial_x - xu\partial_u \qquad \mathbf v_6=4xt\partial_x+4t^2\partial_t-(x^2+2t)u\partial_u \label{eq:heat-equation-v-generators}
    \end{align}
    and the subalgebra $\mathbf v_\alpha=\alpha(x,t)\partial_u$.
\end{example}

Hence, by \cref{thm:infinitesimal_criterion1} we can determine the infinitesimal generators of the symmetry group. In a next step, we have to relate these generators in the Lie algebra to the actual group elements. This connection is provided by the \emph{exponential map} $\exp : \mathfrak g\to G$, whose action on an element $(x,u)\in M\subseteq X\times \mathcal U$ is described by
\begin{align}
    \exp(\vartheta \mathbf v) \circ (x,u) = \sum_{k\geq 0} \frac{\vartheta^k}{k!} \mathbf v^k\circ (x,u) \point \label{eq:expmap}
\end{align}
Let us illustrate the application of the exponential map by continuing the previous example.

\begin{example}[Continuation of \cref{ex:heat-equation-generators}] \label[example]{ex:heat-equation-group}
    To compute the one-parameter group $G_1$ associated to the generator $\mathbf v_1$ from \cref{eq:heat-equation-v-generators} we have to apply $\mathbf v_1$ to a graph of a function, i.e., to an element $(x,t,u)\in M$. We get $\mathbf v_1 \circ (x,t,u) = \partial_x \circ (x,t,u) = (1,0,0)$ and $\mathbf v^k \circ (x,t,u)=(0,0,0)$ for any $k>1$. Thus, the group $G_1$ is acting by
    \begin{align}
        G_1: (x,t,u) \mapsto \exp(\vartheta \mathbf v_1)\circ (x,t,u) = (x,t,u) + (1,0,0) \vartheta = (x+\vartheta,t,u)
    \end{align}
    which is nothing else than a translation in the $x$ coordinate. Similarly, we get a translation in time
    \begin{align}
        G_2: (x,t,u) \mapsto \exp(\vartheta \mathbf v_2)\circ (x,t,u) = (x,t+\vartheta,u) \point
    \end{align}
    To compute the group $G_3$ note, that $\mathbf v_3^k\circ u = u$ for all $k\in\mathbb N$. Hence, the generator $\mathbf v_3$ leads to a scaling invariance
    \begin{align}
        G_3: (x,t,u) \mapsto \exp(\vartheta \mathbf v_3)\circ (x,t,u) = \Big(x,t,u \sum_{k\geq 0}\frac{\vartheta^k}{k!}\Big) = (x,t,e^\vartheta u) \point
    \end{align}
    Quite similar is the computation for $\mathbf v_4$, and with $\mathbf v_4^k\circ x = x$ and $\mathbf v_4^k\circ t = 2^k t$ we obtain
    \begin{align}
        G_4: (x,t,u) \mapsto \exp(\vartheta \mathbf v_4)\circ (x,t,u) = (e^\vartheta x,e^{2\vartheta} t,u) \point
    \end{align}
    To compute the group action of $G_5$, note that
    \begin{align}
        \mathbf v_5^k \circ u = u \sum_{0\leq m\leq k/2} \frac{k!}{m!(k-2m)!} (-x)^{k-2m} (-t)^m \comma
    \end{align}
    which can be shown by induction including a distinction between even and odd $k$. Including that relation in the series of \cref{eq:expmap} we obtain
    \begin{align}
        G_5: (x,t,u) \mapsto \exp(\vartheta \mathbf v_5)\circ (x,t,u) = \Big(x+2 \vartheta t,t, u e^{-\vartheta x-\vartheta^2 t}\Big) \point
    \end{align}
    To find the action of the group $G_6$, we will use another useful property of the exponential map, which is
    \begin{align}
        A \circ \exp(B) \circ A^{-1} = \exp(A\circ B \circ A^{-1})
    \end{align}
    for any functions $A, B$ with appropriate domains and codomains and where $A\circ A^{-1}=\mathbf 1$. This relation can directly be shown by the definition \cref{eq:expmap}. In particular, the so-called inversion map 
    \begin{align}
        I(x,t) = \Big(\frac{x}{t^2-x^2},\frac{t}{t^2-x^2}\Big)
    \end{align}
    has the property $I\circ I = \mathbf 1$. As we can write $\exp(\vartheta \mathbf v_6) = I \circ\exp(-\vartheta\partial_x)\circ I$ we will find
    \begin{align}
        G_6: (x,t,u) \mapsto \exp(\vartheta \mathbf v_6)\circ (x,t,u) = \Big(\frac{x}{1-4\vartheta t}, \frac{t}{1-4\vartheta t}, u \sqrt{1-4\vartheta t}\ e^{-\frac{\vartheta x^2}{1-4\vartheta t}} \Big) \point
    \end{align}
    The last generator $\mathbf v_\alpha$ will result in the action
    \begin{align}
        G_\alpha : (x,t,u) \mapsto (x,t, u +\vartheta\, \alpha(x,t)) \point
    \end{align}
\end{example}

In summary, we have now determined the group actions on the graph of a function $\Gamma_f \subset X\times \mathcal U$, i.e., the action on every point $(x,u)$ of that graph. In a final step, we would like to transfer this action onto the function itself. For simplicity, we will focus on the case where the action of the group element $g\in G$ takes the form $(\tilde x,\tilde u) = g \circ (x,u) = (\Xi_g(x),\Phi_g(x,u))$, i.e., that the action of $g$ on $x$ will not depend on $u$. In this particular case, we have $x=\Xi_g(\tilde x)^{-1} = \Xi_{g^{-1}}(\tilde x)$, and therefore
\begin{align}
    \tilde f (\tilde x) = \Phi_g(\Xi_{g^{-1}}(\tilde x), f(\Xi_{g^{-1}}(\tilde x))) \point\label{eq:group-functrafo-correspondence}
\end{align}
For the general case, we refer to \citep[sec. 2.2]{olver1993applications}. 

\begin{example}[Continuation of \cref{ex:heat-equation-group}] \label{ex:heat-equation-trafo}
    By using \cref{eq:group-functrafo-correspondence}, we can relate the group actions from \cref{ex:heat-equation-group} to transformations of functions. For example, for $G_1$ we have $\Xi_{g_{1,\vartheta}}(x) = x+\vartheta$ and $\Phi_{g_{1,\vartheta}}(x,u) = u$. With $\Xi_{g_{1,\vartheta}^{-1}}(x) = \Xi_{g_{1,-\vartheta}}(x)=x-\vartheta$, we get the transformation
    \begin{align}
        T^1_\vartheta : f(x,t) \mapsto f(x-\vartheta,t) \label{eq:ex_heat1d_trafo1}
    \end{align}
    as expected. By applying \cref{eq:group-functrafo-correspondence} to the other groups $G_2,\ldots,G_6,G_\alpha$, we obtain the results presented in \cref{ssec:lie_symmetries}
    \begin{align}
        T^2_\vartheta &: f(x,t) \mapsto f(x, t - \vartheta)\ , \label{eq:ex_heat1d_trafo2}\\
        T^3_\vartheta &: f(x,t) \mapsto e^{\vartheta} f(x, t)\ , \label{eq:ex_heat1d_trafo3}\\
        T^4_\vartheta &: f(x,t) \mapsto f(e^{-\vartheta} x, e^{-2\vartheta} t)\ , \label{eq:ex_heat1d_trafo4}\\
        T^5_\vartheta &: f(x,t) \mapsto e^{-\vartheta x + \vartheta^2 t} f(x - 2\vartheta t, t)\ , \label{eq:ex_heat1d_trafo5}\\
        T^6_\vartheta &: f(x,t) \mapsto \frac{1}{\sqrt{1 + 4\vartheta t}} \exp\!\left( \frac{-\vartheta x^2}{1 + 4\vartheta t} \right) f\!\left( \frac{x}{1 + 4\vartheta t}, \frac{t}{1 + 4\vartheta t} \right) \ , \label{eq:ex_heat1d_trafo6}\\
        T_{\vartheta}^\alpha & : f(x,t) \mapsto f(x,t) + \vartheta\, \alpha(x,t) \point \label{eq:ex_heat1d_linear}
    \end{align}    
    for any $\vartheta\in\mathbb R$, such that the functions are defined and an arbitrary solution $\alpha(x,t)$ of the heat equation. Hence, for any solution $u=f(x,t)$ of the heat equation \cref{eq:ex-heateq}, applying any of the aforementioned transformations produces another possible solution. As an example, for the constant solution $f(x,t)\equiv 1$ and after applying the transformation $T^6_\vartheta$, one obtains the function $u=\frac{1}{\sqrt{1+4\vartheta t}} \exp\!\left( \frac{-\vartheta x^2}{1 + 4\vartheta t} \right)$ that also solves the heat equation, whenever this function is defined, i.e.\ for $1+4\vartheta t\neq 0$.

    While the transformations $T^1,\ldots,T^4$ can be found by carefully observing the differential equation \cref{eq:ex-heateq}, $T^5$ and $T^6$ are far from being trivial to be obtained by reading the PDE.
\end{example}

Thus, we have a framework that, in principle, allows us to compute a symmetry group for any PDE system. For PDE systems with a certain additional property, we can give even a stronger statement. According to \citep[Def. 2.70]{olver1993applications}, a system of PDEs $\mathcal D$ is called \emph{local solvable} at a point $(x_0,u^{(d)}_0)\in X \times \mathcal U^{(d)}$, if there is a smooth solution $u=f(x)$ for $x$ in a neighbourhood of $x_0$, with $u_0^{(d)}=\pr f(x_0)$. The whole PDE system $\mathcal D$ is called locally solvable, if this property holds for any point $(x_0,u^{(d)}_0)$ satisfying $\mathcal D(x_0,u^{(d)}_0)=0$. For a locally solvable system, we can give a sufficient and necessary description of symmetry groups:

\begin{theorem}[Infinitesimal criterion {\citep[see][thm. 2.71]{olver1993applications}}] \label{thm:infinitesimal_criterion2}
    Let $\mathcal D$ be a locally solvable system of PDEs with maximal rank defined over $M\subseteq X\times U$ and let $G$ be a local group of transformations acting on $M$. $G$ is a symmetry group of $\mathcal D$, \emph{if and only if}
    \begin{align}
        \pr \mathbf v [\mathcal D(x,u^{(d)})] = 0, \qquad\text{whenever}\ \mathcal D(x,u^{(d)})=0
    \end{align}
    holds for any infinitesimal generator $\mathbf v$ of $G$.
\end{theorem}
Thus, for locally solvable systems, we can derive \emph{all} symmetry groups of the PDE system with the aforementioned framework. This ability to characterize the complete symmetry group of the solution manifold is precisely what renders Lie symmetries so powerful for our approach.

We would like to conclude this section by recapping all the steps to compute the symmetry group:
\begin{enumerate}[itemsep=1pt,topsep=1pt]
    \item[0.] Check maximal rank and local solvability of $\mathcal D$.
    \item Write an ansatz for a vector field $\mathbf v$ on $X\times \mathcal U$ according to \cref{eq:vector-field_ansatz} and compute its prolongation by \cref{eq:prolongation1}, \cref{eq:phiJ} and \cref{eq:totalderivative}.
    \item Derive the infinitesimal criterion \cref{thm:infinitesimal_criterion1,thm:infinitesimal_criterion2}, i.e., the prolongation of $\mathbf v$ applied on the PDE should vanish whenever the PDE is satisfied.
    \item By comparing the coefficients of the infinitesimal criterion, one obtains an overdetermined system of simple PDEs (``determining equations'').
    \item Solve the determining equations by elementary methods, to get the most general generators $\mathbf v$ and choose a basis of the subspace generated by $\mathbf v$.
    \item Derive one-parameter Lie groups from infinitesimal generators by the exponential map $g_\vartheta \circ (x,u) = \exp(\vartheta \mathbf v) \circ(x,u)$.
    \item Derive the transformation maps from these group actions by \cref{eq:group-functrafo-correspondence}.
\end{enumerate}

There are several implementations available, which will solve this task step by step. We refer to \cite{steinberg2014practicalguidesymboliccomputation, Hereman1996, Butcher2003, Champagne1991}, which give an overview about different available implementations.

\section{Experimental Details}
\label{app-sec:experiments}

\subsection{Reference Solution}
\label{app-ssec:reference_solution}
In order to compare the predictions of \textsc{LieSolver}, we compute reference ground-truth solutions by expanding the initial data in a sine basis according to \cite{Boyce2016}. For the heat equation with Dirichlet BC, we set $b_L=f(0,0)$ and $b_R=f(1,0)$ and define $w(x)=b_L+(b_R-b_L)\,x$ so that $v_0(x)=f(x,0)-w(x)$ satisfies homogeneous Dirichlet conditions. We sample $v_0$ at the interior points $x_m=\frac{m}{M+1}$ for $m=1,\dots,M$ and obtain the coefficients $A_m$ via orthonormal projection onto $\sin(m\pi x)$. The solution is
\begin{align}
    f_\text{ref}(x,t)=w(x)+\sum_{m=1}^{M} A_m\,\sin\!\big(m\pi x\big)\, e^{- m^2\pi^2\,t}.
\end{align}
For the wave equation with homogeneous Dirichlet boundaries, we sample $f(x,0)$ and $f_t(x,0)$ at the same interior points $x_m$ and compute $A_m$ and $B_m$ by orthonormal projection onto $\sin(m\pi x)$. The solution is then given by
\begin{align}
    f_\text{ref}(x,t)=\sum_{m=1}^{M}\Big[A_m\cos\!\big(m\pi t\big)+\frac{B_m}{\omega_m}\sin\!\big(m\pi t\big)\Big]\sin\!\big(m\pi x\big).
\end{align}

\subsection{PINN Baseline Setup}
\label{app-ssec:pinn_setup}
We implement our PINN baseline using the DeepXDE framework \cite{lu2021deepxde} with a PyTorch backend. The model $f_{\text{PINN}}:\mathbb{R}^2\to\mathbb{R}$ is a fully connected NN with $\tanh$ activation and Glorot normal initialization. The number of layers and width are specified per case in \cref{tab:PINN_results}, following recommendations from \cite{wang2023expertsguidetrainingphysicsinformed}.

Using the operators $\mathcal D$ and $\mathcal B_k$ from \cref{ssec:IBVPs}, the PINN loss is
\begin{align}
\mathcal{L}_{\mathrm{res}}(\boldsymbol w)
&=\frac{1}{N_r}\sum_{x_k\in \mathcal X_{r}}\big(\mathcal D[f_{\text{PINN}}](x_k;\boldsymbol w)\big)^2,\\
\mathcal{L}_{\mathrm{IBC}}(\boldsymbol w)
&=\frac{1}{L}\sum_{k=1}^{r}\sum_{x_j\in \mathcal X_k}\big(\mathcal B_k[f_{\text{PINN}}](x_j;\boldsymbol w)\big)^2,\\
\mathcal{L}(\boldsymbol w)
&=\mathcal{L}_{\mathrm{res}}(\boldsymbol w)+w_{\mathrm{IBC}}\,\mathcal{L}_{\mathrm{IBC}}(\boldsymbol w) \comma
\end{align}
where $\mathcal X_{r}\subset\Omega$ are domain collocation points and $\mathcal X_k\subset\Gamma_k$ are boundary points. For the wave equation, $\mathcal{L}_{\mathrm{IBC}}$ includes the zero initial velocity condition $\partial_t f|_{t=0}=0$ as an additional boundary term. The weight $w_{\mathrm{IBC}}$ is tuned per case to balance the loss components (see \cref{tab:PINN_results}).

We sample $4000$ domain points and $1000$--$1500$ boundary points using Hammersley sequences. Training proceeds with Adam for $1000$ iterations followed by L-BFGS until convergence or optimizer stall. All experiments were run on the same hardware\footnote{Intel Core Ultra 7 165H CPU with 11 cores and 16\,GiB RAM.} as \LieSolver. Reported runtimes measure wall-clock time from initialization to convergence, excluding data generation.

\subsection{IBC Data Specifications}
\label{app-ssec:ibc_data}

\begin{figure}[h]
  \centering
  \includegraphics[width=0.8\linewidth]{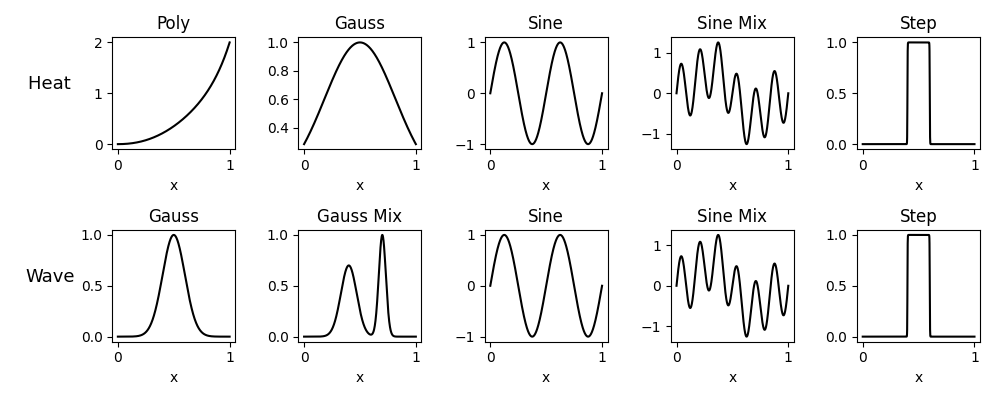}
  \caption{Initial conditions (solution shapes at $t=0$) for the test cases for the heat (top row) and the wave (bottom row) equations.}
  \label{fig:ic_shapes}
\end{figure}

\paragraph{Heat Equation} 
We consider the domain $(x,t)\in[0,1]\times[0,0.1]$. The boundary conditions are fixed to the constant values of the initial profile at the boundaries: $f(0,t)=f_0(0)$ and $f(1,t)=f_0(1)$. We employ five distinct initial profiles $f(x,0)=f_0(x)$, visualized in the top row of \cref{fig:ic_shapes}:
\begin{align}
    \text{\IC{poly}}: \quad & f_0(x) = x^2 + x^3 - x^5 + x^7 \\
    \text{\IC{gauss}}: \quad & f_0(x) = e^{-5(x-0.5)^2} \\
    \text{\IC{sine}}: \quad & f_0(x) = \sin(4\pi x) \\
    \text{\IC{sine mix}}: \quad & f_0(x) = 0.5\sin(2\pi x) - 0.2\sin(4\pi x) + 0.7\sin(12\pi x) \\
    \text{\IC{step}}: \quad & f_0(x) = 0.5 \Big[\tanh\big(500(x-0.4)\big) - \tanh\big(500(x-0.6)\big)\Big]
\end{align}
To generate the training dataset $\mathcal{X}$, we sample points using a low-discrepancy Halton sequence to ensure uniform coverage of the initial and boundary manifolds. Specifically, we sample $1000$ points at $t=0$, $1000$ points at each boundary $x=0$ and $x=1$, resulting in a total dataset size of $L=3000$. We generate a test set of $3000$ uniformly sampled points on the IBCs and a domain set of $10000$ uniformly sampled points.

\paragraph{Wave Equation}
We consider the domain $(x,t)\in[0,1]\times[0,1]$. We apply homogeneous Dirichlet boundary conditions $f(0,t)=f(1,t)=0$ and a zero initial velocity condition $f_t(x,0)=0$. The initial displacement profiles $f(x,0)=f_0(x)$ mirror the complexity of the heat equation cases (see bottom row of \cref{fig:ic_shapes}):
\begin{align}
    \text{\IC{gauss}}: \quad & f_0(x) = e^{-50(x-0.5)^2} \\
    \text{\IC{gauss mix}}: \quad & f_0(x) = 0.7 e^{-100(x-0.4)^2} + e^{-500(x-0.7)^2} \\
    \text{\IC{sine}}: \quad & f_0(x) = \sin(4\pi x) \\
    \text{\IC{sine mix}}: \quad & f_0(x) = 0.5\sin(2\pi x) - 0.2\sin(4\pi x) + 0.7\sin(12\pi x) \\
    \text{\IC{step}}: \quad & f_0(x) = 0.5 \Big[\tanh\big(500(x-0.4)\big) - \tanh\big(500(x-0.6)\big)\Big]
\end{align}
Training data is similarly sampled via the Halton sequence. To prevent overfitting we increase the sampling density relative to the heat equation. We use $1500$ points for the initial displacement, $1000$ points for the initial velocity, and $1250$ points for each boundary, resulting in a total dataset size of $L=5000$. We generate a test set of $5000$ uniformly sampled points on the IBCs and a domain set of $10000$ uniformly sampled points.

\begin{figure}
  \centering
  \includegraphics[width=1.\linewidth]{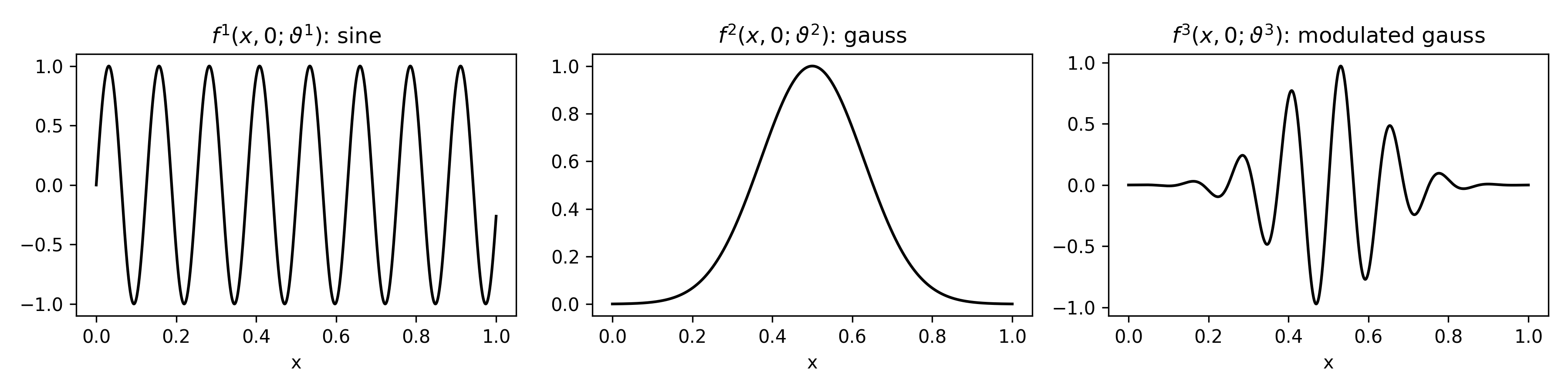}
  \caption{Representative brick families for one-dimensional heat equation at $t=0$: $f^1$ (sine), $f^2$ (Gaussian), and $f^3$ (Gaussian modulated by a sine) with illustrative parameter choices.}
  \label{fig:brick_solutions}
\end{figure}

\paragraph{Brick specification details}
\Cref{fig:brick_solutions} illustrates the variety of brick solutions at $t=0$ constructed to solve the one-dimensional heat equation. The choice of parameter bounds and sampling strategies for the Lie symmetries is crucial for model performance; ranges must be physics-informed to ensure that the generated features lie within the physical domain and cover the relevant scales. E.g. we employ log-uniform sampling for scaling parameter of transformation \cref{eq:heat1d_trafo6} to capture features across multiple orders of magnitude.

For the heat equation, we utilize the following specifications:
\begin{itemize}[itemsep=0pt, topsep=2pt]
    \item $f^1$ (sine mode): $T^4_{\vartheta_2} \cdot T^1_{\vartheta_1} \circ (e^{-t}\sin x)$, with phase shift $\vartheta_1 \sim \mathcal{U}(0, \pi)$ and frequency $\vartheta_2 \sim \mathcal{U}(0, 100)$.
    \item $f^2$ (Gaussian): $T^1_{\vartheta_2} \cdot T^6_{\vartheta_1} \circ (1)$, with position $\vartheta_2 \sim \mathcal{U}(0, 1)$ and shape parameter $\vartheta_1 \sim \text{Log}\mathcal{U}(10^{-1}, 10^6)$.
    \item $f^3$ (Modulated Gaussian): $T^1_{\vartheta_4} \cdot T^6_{\vartheta_3} \cdot T^4_{\vartheta_2} \cdot T^1_{\vartheta_1} \circ (e^{-t}\sin x)$, combining the bounds of the previous families to generate localized oscillatory features.
\end{itemize}

For the wave equation, we specify:
\begin{itemize}[itemsep=0pt, topsep=2pt]
    \item $f^1$ (Standing wave): $T^2_{\vartheta_2} \cdot T^1_{\vartheta_1} \circ (\sin x \cos t)$, with phase $\vartheta_1 \sim \mathcal{U}(0, \pi)$ and frequency scaling $\vartheta_2 \sim \mathcal{U}(0, 100)$.
    \item $f^2$ (Travelling wave): $T^1_{\vartheta_2} \cdot T^2_{\vartheta_1} \circ (e^{-(x+t)^2} + e^{-(x-t)^2})$, with translation $\vartheta_2 \sim \mathcal{U}(-1, 2)$ to allow waves to enter from boundaries, and width scaling $\vartheta_1 \sim \mathcal{U}(0, 1000)$.
\end{itemize}

\clearpage
\section{Experimental Results}
\label{app-sec:results}

We present the detailed quantitative results for all experimental cases in \Cref{tab:LieSolver_results} for \textsc{LieSolver} and \Cref{tab:PINN_results} for the baseline PINN. For reference, a standard finite difference method (FDM) \cite{leveque2007finite} solves these benchmarks in under a second with comparable accuracy. 

For completeness, we also evaluated additional neural baselines: a hard-constrained PINN \cite{Lu2021}, which modifies the network output to satisfy IBCs exactly by construction, and a physics-informed Kolmogorov-Arnold network \cite{liu2025kan}. Both methods performed comparably to or worse than the standard PINN baseline in accuracy and runtime on our considered benchmarks. We therefore focus our comparison on the standard PINN only.

In the following, we discuss the \LieSolver optimization behaviour and qualitative results for the specific ICs.

\begin{table}
\centering
\caption{Median \textsc{LieSolver} runs for the one-dimensional heat and wave equations, reporting metrics, runtime and model complexity.}
\label{tab:LieSolver_results}
\setlength{\tabcolsep}{4pt}
\renewcommand{\arraystretch}{1.1}
\scriptsize
\begin{tabular}{lcccccc}
\hline
PDE & \makecell{IBC\\Type} & \makecell{MSE\\IBC} & \makecell{L2RE\\domain} & Runtime\ [s]& \# Bricks & \# Parameters \\
\hline
     & \IC{poly}           & $8.2\cdot10^{-7}$ & $4.4\cdot10^{-4}$ & 34  & 31 & 86 \\
     & \IC{gauss}          & $2.4\cdot10^{-8}$ & $2.0\cdot10^{-4}$ & 11   & 15 & 40  \\
Heat & \IC{sine}           & $2.4\cdot10^{-8}$ & $4.6\cdot10^{-4}$ & 4   & 5  & 12  \\
     & \IC{sine mix}       & $2.0\cdot10^{-8}$ & $3.5\cdot10^{-4}$ & 10   & 15 & 38  \\
     & \IC{step}           & $9.1\cdot10^{-7}$ & $1.3\cdot10^{-3}$ & 310  & 78 & 230 \\
\hline
     & \IC{gauss}          & $4.9\cdot10^{-8}$ & $1.1\cdot10^{-3}$ & 8   & 15 & 30 \\
     & \IC{gauss mix}      & $7.8\cdot10^{-7}$ & $5.0\cdot10^{-3}$ & 41  & 41 & 82  \\
Wave & \IC{sine}           & $1.1\cdot10^{-7}$ & $8.1\cdot10^{-4}$ & 5   & 9  & 18  \\
     & \IC{sine mix}       & $4.6\cdot10^{-7}$ & $1.7\cdot10^{-3}$ & 4   & 7  & 14  \\
     & \IC{step}           & $9.4\cdot10^{-7}$ & $4.0\cdot10^{-3}$ & 217 & 91 & 182 \\
\hline
\end{tabular}
\end{table}

\begin{table}
\centering
\caption{Median PINN runs for the one-dimensional heat and wave equations, reporting metrics, runtime, and model configuration.}
\label{tab:PINN_results}
\setlength{\tabcolsep}{4pt}
\renewcommand{\arraystretch}{1.1}
\scriptsize
\begin{tabular}{lccccccc}
\hline
PDE & \makecell{IBC\\Type} & \makecell{MSE\\(IBC)} & \makecell{L2RE\\(domain)} & Runtime [s] & \makecell{Training\\iterations} & Network & \makecell{IBC\\weight} \\
\hline
     & \IC{poly}     & $9.0\cdot10^{-7}$ & $4.8\cdot10^{-4}$ & 137  & 4472 & $4\times50$  & $10^{3}$ \\
     &\IC{gauss}       & $3.2\cdot10^{-7}$ & $4.9\cdot10^{-4}$ & 123  & 4554 & $4\times50$  & $10^{3}$ \\
Heat & \IC{sine}           & $7.8\cdot10^{-7}$ & $5.0\cdot10^{-3}$ & 183  & 5238 & $4\times50$  & $10^{3}$ \\
     & \IC{sine mix}       & $4.6\cdot10^{-6}$ & $1.9\cdot10^{-2}$ & 1716 & 10674 & $6\times100$ & $10^{5}$ \\
     & \IC{step}           & $1.8\cdot10^{-3}$ & $2.2\cdot10^{-1}$ & 440 & 4005 & $6\times100$ & $10^{5}$ \\
\hline
     &\IC{gauss}          & $1.5\cdot10^{-7}$ & $1.6\cdot10^{-3}$ & 1119 & 8321 & $4\times50$  & $10^{3}$ \\
     & \IC{gauss mix}      & $2.7\cdot10^{-6}$ & $8.2\cdot10^{-3}$ & 2456 & 17864 & $4\times50$  & $10^{3}$ \\
Wave & \IC{sine}           & $2.0\cdot10^{-7}$ & $8.5\cdot10^{-4}$ & 1172 & 8744 & $4\times50$  & $10^{3}$ \\
     & \IC{sine mix}       & $8.2\cdot10^{-5}$ & $1.8\cdot10^{-2}$ & 11008 & 22000 & $6\times100$ & $10^{3}$ \\
     & \IC{step}           & $7.2\cdot10^{-5}$ & $4.5\cdot10^{-2}$ & 909 & 7786 & $4\times50$  & $10^{3}$ \\
\hline
\end{tabular}
\end{table}

\subsection{LieSolver Results}
\label{app-ssec:liesolver_results}
We analyze the performance of \textsc{LieSolver} across different ICs, highlighting optimization specifics.

\begin{figure}
    \centering
    \begin{subfigure}{.25\linewidth}
        \centering
        \includegraphics[width=\linewidth]{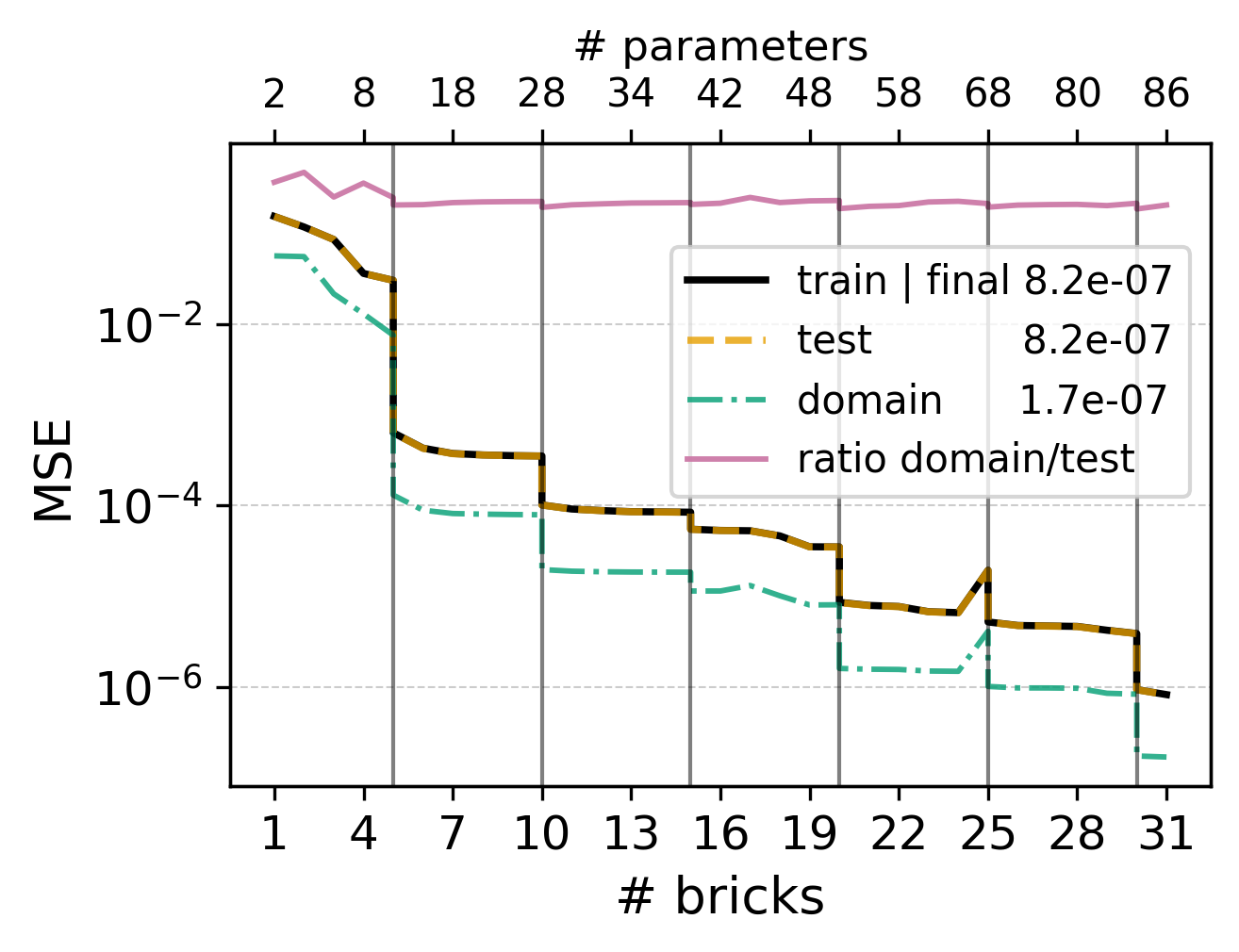}
        \caption{Fit progress.}
        \label{fig-app:liesolver_fit_heat_poly}
    \end{subfigure} %
    \begin{subfigure}{.33\linewidth}
        \centering
        \includegraphics[width=\linewidth]{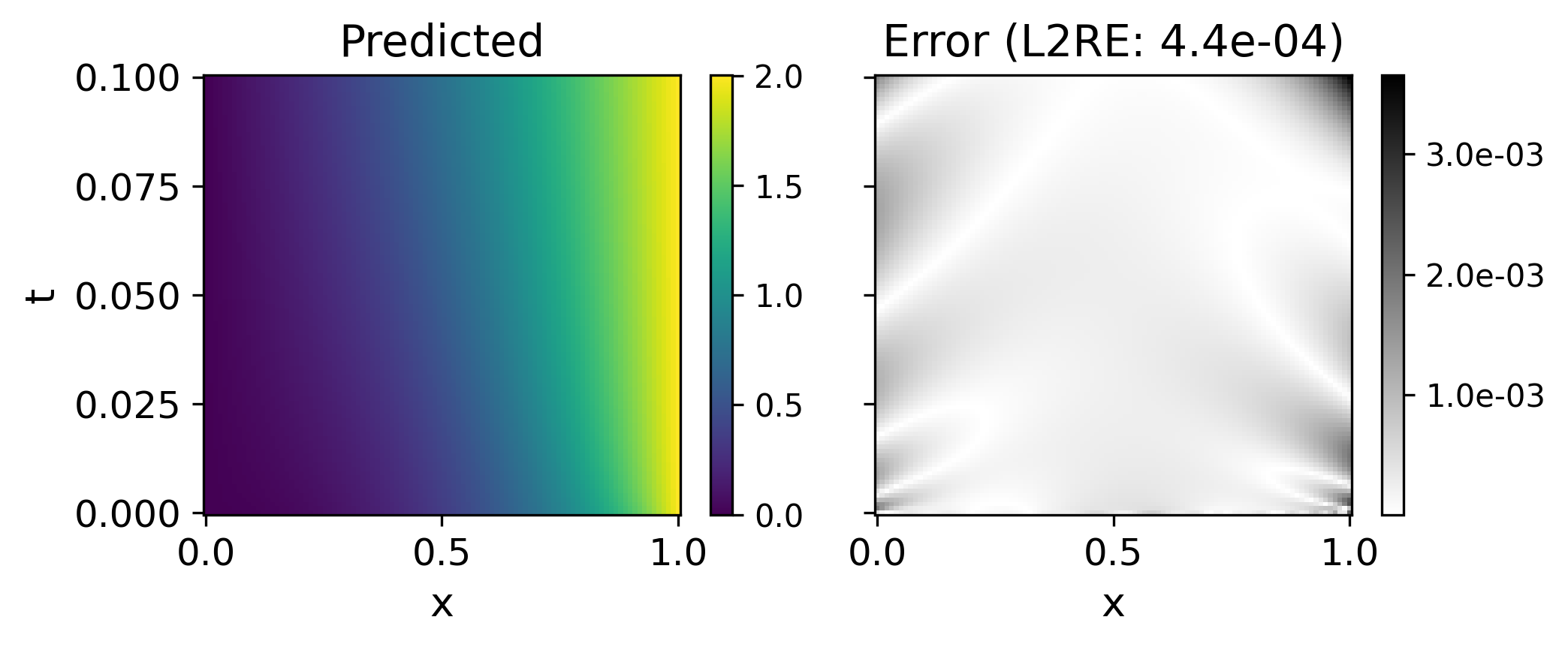}
        \caption{2D domain fit.}
        \label{fig-app:liesolver_domain_heat_poly}
    \end{subfigure}
    \begin{subfigure}{.41\linewidth}
        \centering
        \includegraphics[width=\linewidth]{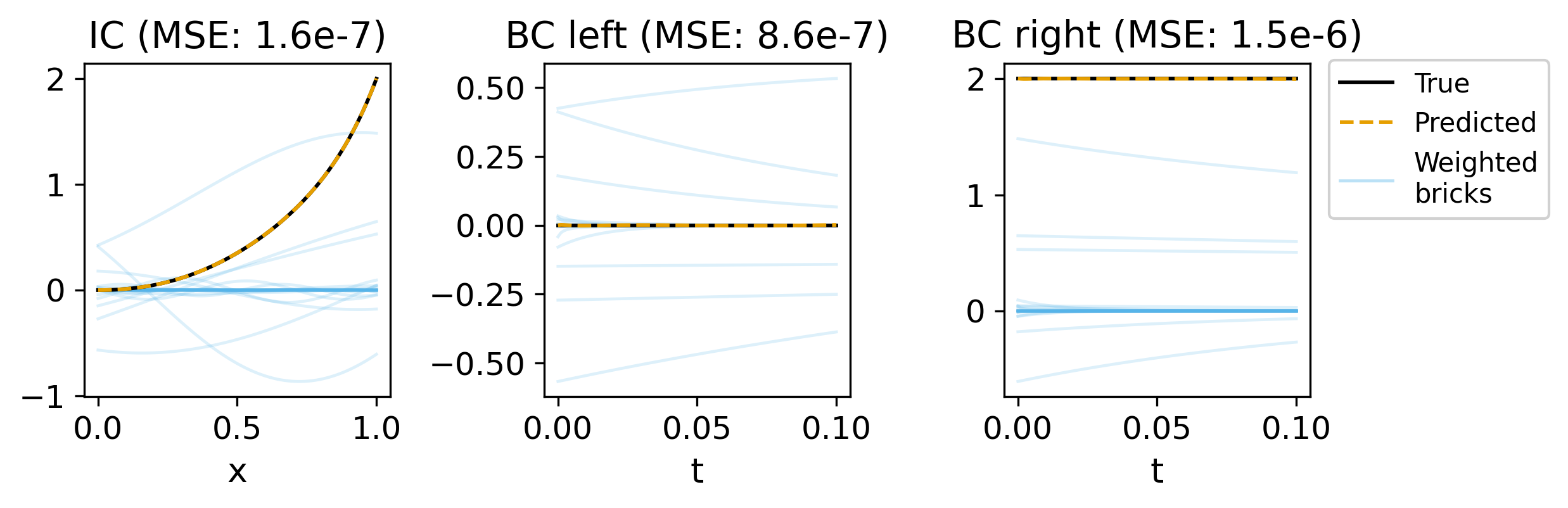}
        \caption{IBC fit.}
        \label{fig-app:liesolver_ibc_heat_poly}
    \end{subfigure}
        \caption{\LieSolver results for heat equation with \IC{Poly} IC.}
    \label{fig-app:liesolver_heat_poly}
\end{figure}

\textbf{Heat \IC{Poly}:} The fit progress shown in \Cref{fig-app:liesolver_fit_heat_poly} illustrates a fallback of the greedy structural learning stage. Around the 25th brick addition, we observe a temporary worsening of the MSE. This occurs because the greedy initialization selects a brick from a finite random pool based on cosine similarity with the current residual. While this candidate is locally optimal with respect to the pool, its initial addition to the linear combination may temporarily disrupt the global fit due to the irregular residual profile. However, the subsequent global parameter refinement stage successfully resolves this, realigning the parameters to minimize the global loss and allowing the model to monotonically converge to the target MSE with 31 bricks.

\begin{figure}
    \centering
    \begin{subfigure}{.25\linewidth}
        \centering
        \includegraphics[width=\linewidth]{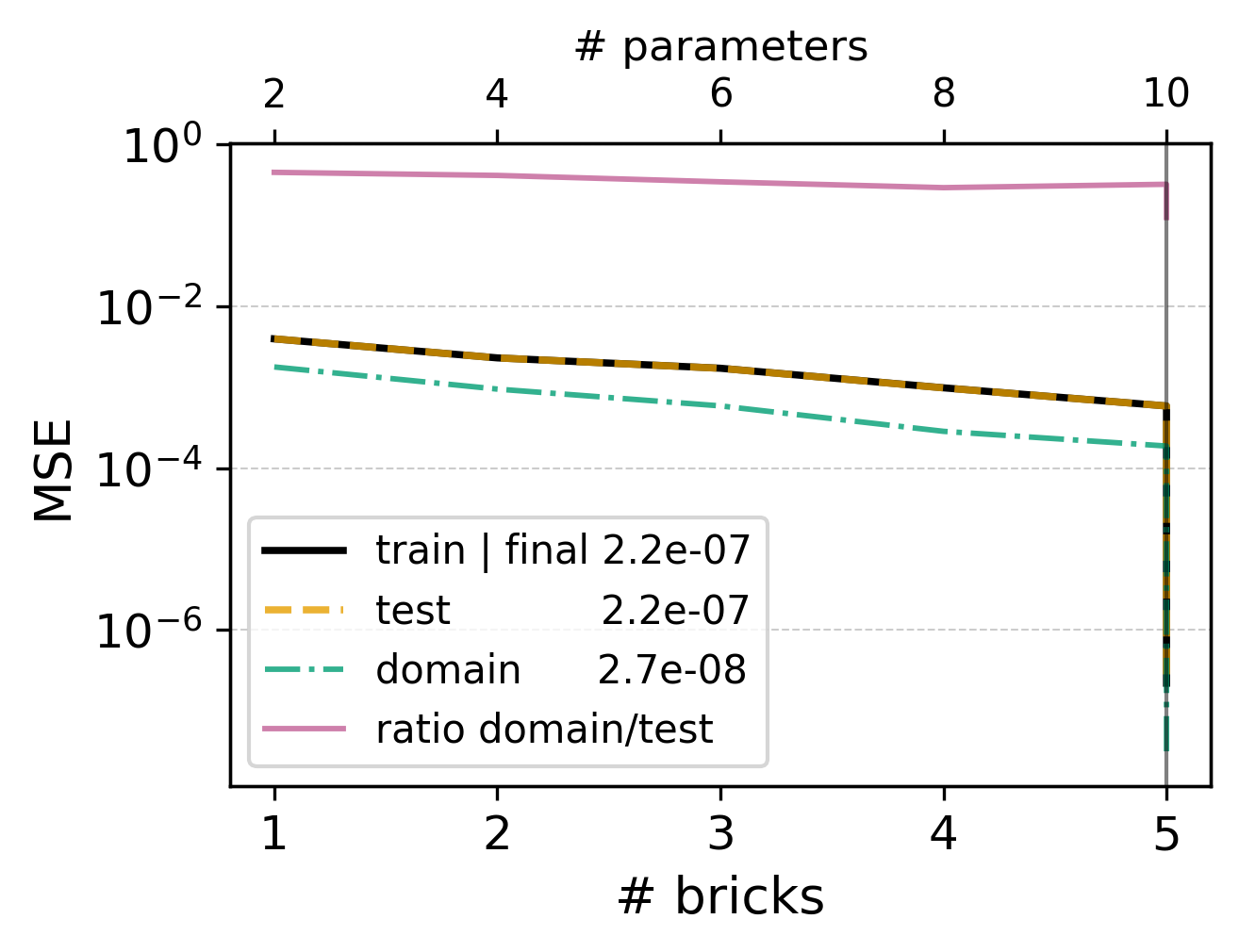}
        \caption{Fit progress.}
        \label{fig-app:liesolver_fit_heat_gauss}
    \end{subfigure} %
    \begin{subfigure}{.33\linewidth}
        \centering
        \includegraphics[width=\linewidth]{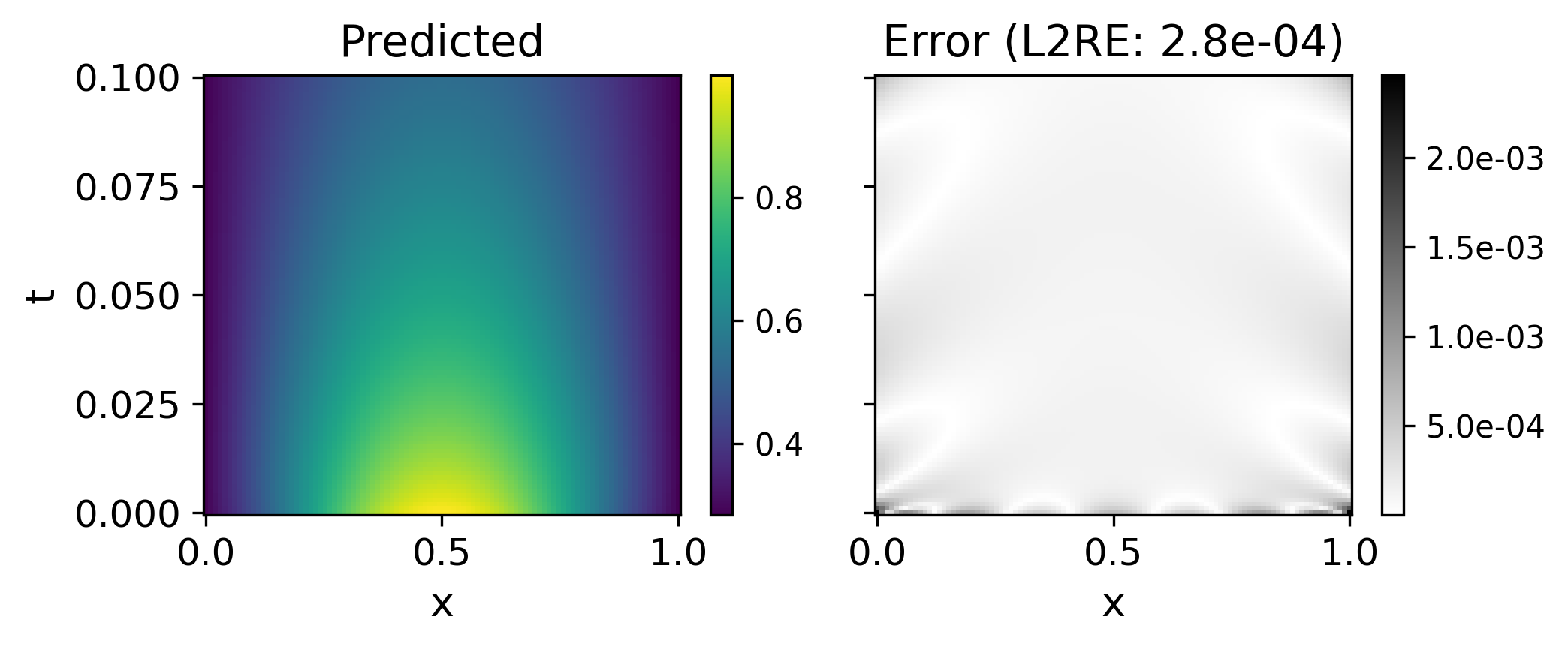}
        \caption{2D domain fit.}
        \label{fig-app:liesolver_domain_heat_gauss}
    \end{subfigure}
    \begin{subfigure}{.41\linewidth}
        \centering
        \includegraphics[width=\linewidth]{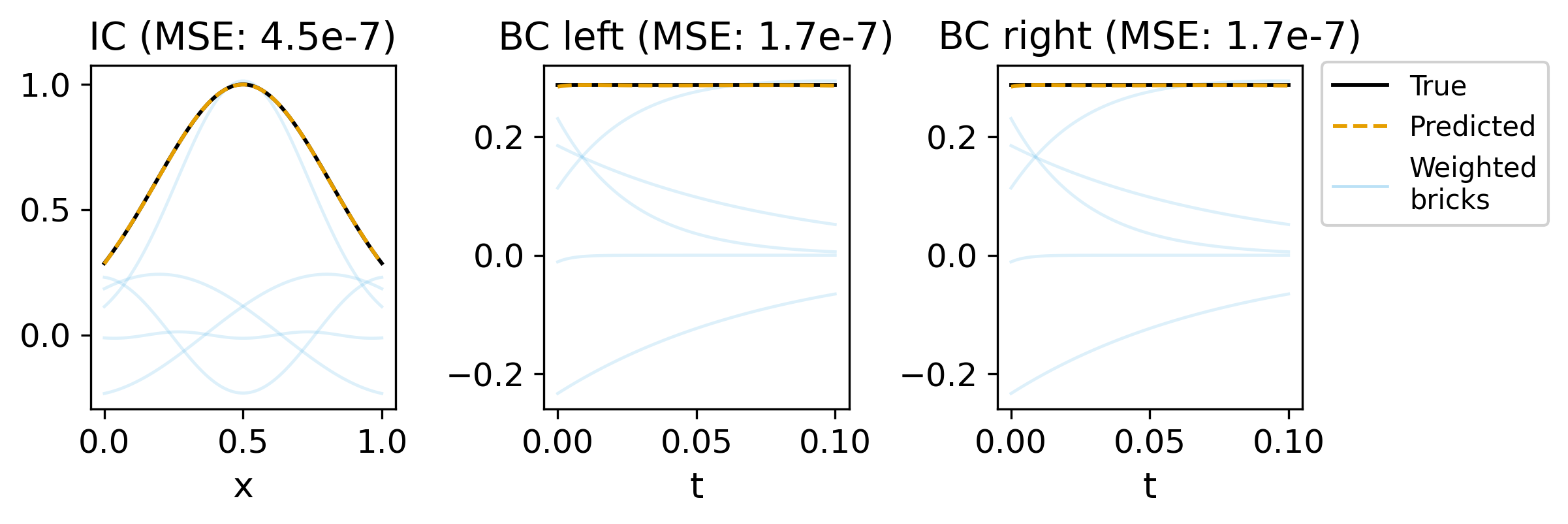}
        \caption{IBC fit.}
        \label{fig-app:liesolver_ibc_heat_gauss}
    \end{subfigure}
        \caption{\LieSolver results for heat equation with \IC{Gauss} IC. \Cref{tab:LieSolver_results} shows the result for run from the main text.}
    \label{fig-app:liesolver_heat_gauss}
\end{figure}

\textbf{Heat \IC{Gauss}:} While the aggregate results in \Cref{tab:LieSolver_results} utilized our standard hyperparameter setting (\texttt{nfev}=4), we illustrate a specific run in \Cref{fig-app:liesolver_heat_gauss} where the NLLS refinement budget was increased to \texttt{nfev}=40. This demonstrates the trade-off between structural search and parameter optimization: with deeper refinement, the optimizer locates a high-quality local minimum more effectively, allowing \textsc{LieSolver} to satisfy the tolerance with only 5 bricks (compared to 15 bricks in the median run in the main text). The decomposition in \Cref{fig-app:liesolver_ibc_heat_gauss} shows the solution is comprised of a primary Gaussian brick supplemented by two pairs of symmetric sine waves to correct boundary deviations.

\begin{figure}
    \centering
    \begin{subfigure}{.25\linewidth}
        \centering
        \includegraphics[width=\linewidth]{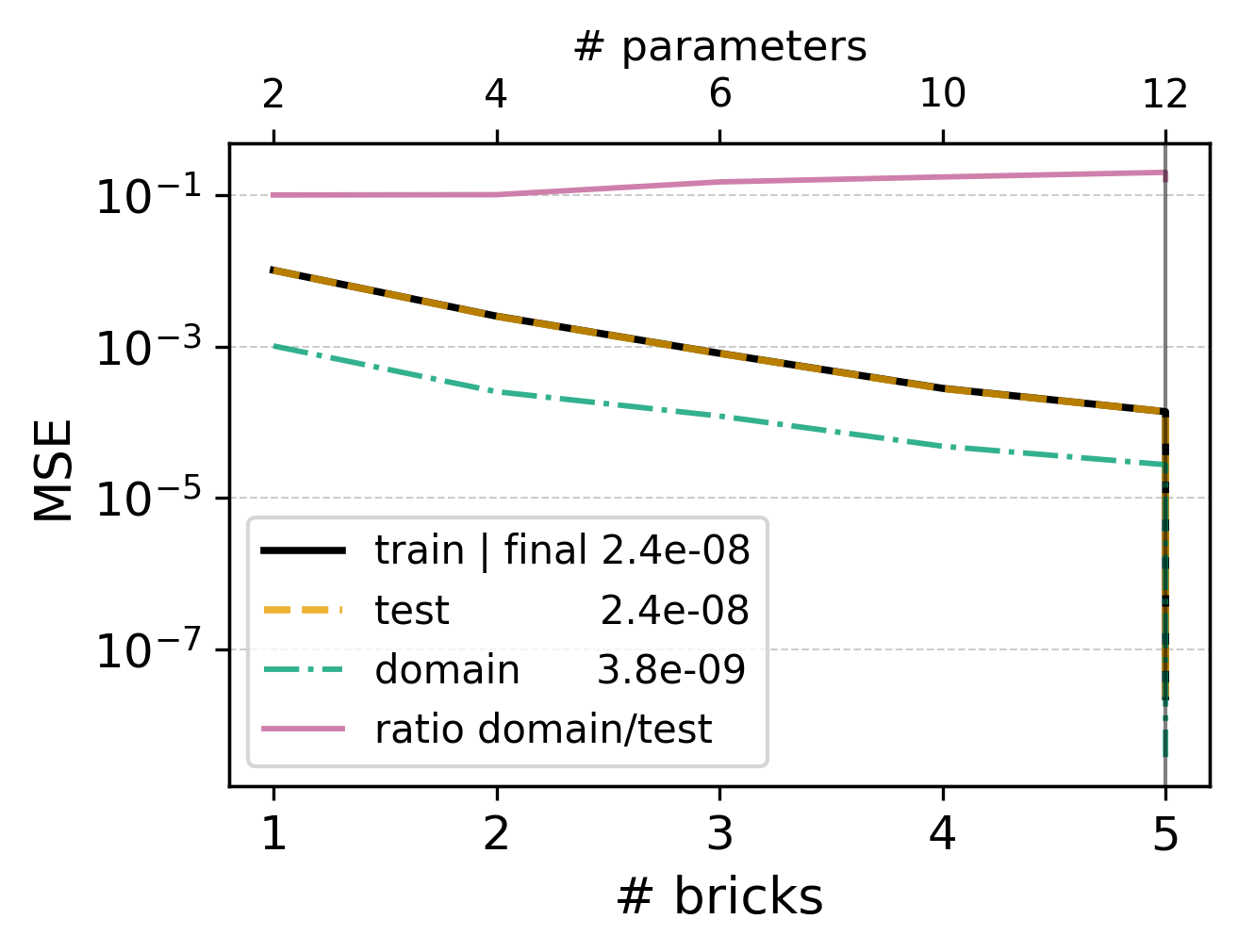}
        \caption{Fit progress.}
        \label{fig-app:liesolver_fit_heat_sine}
    \end{subfigure} %
    \begin{subfigure}{.33\linewidth}
        \centering
        \includegraphics[width=\linewidth]{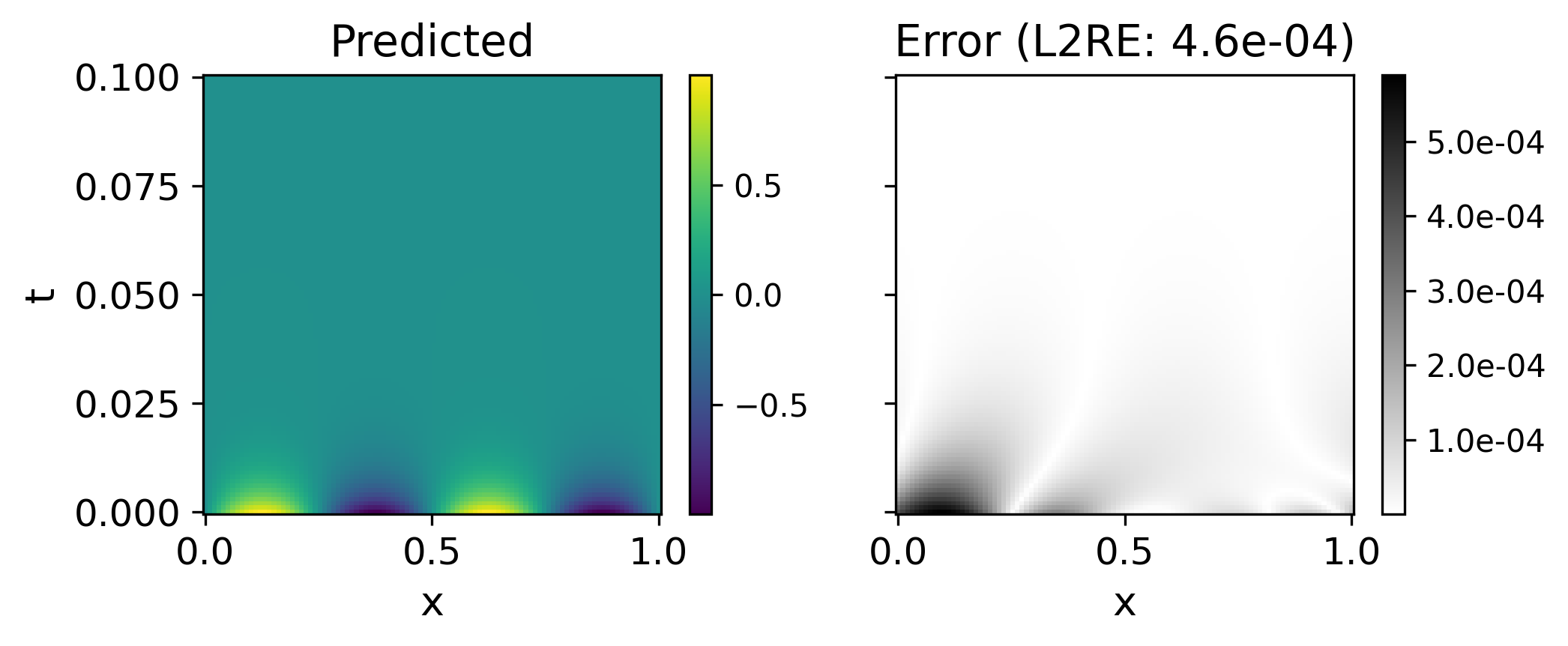}
        \caption{2D domain fit.}
        \label{fig-app:liesolver_domain_heat_sine}
    \end{subfigure}
    \begin{subfigure}{.41\linewidth}
        \centering
        \includegraphics[width=\linewidth]{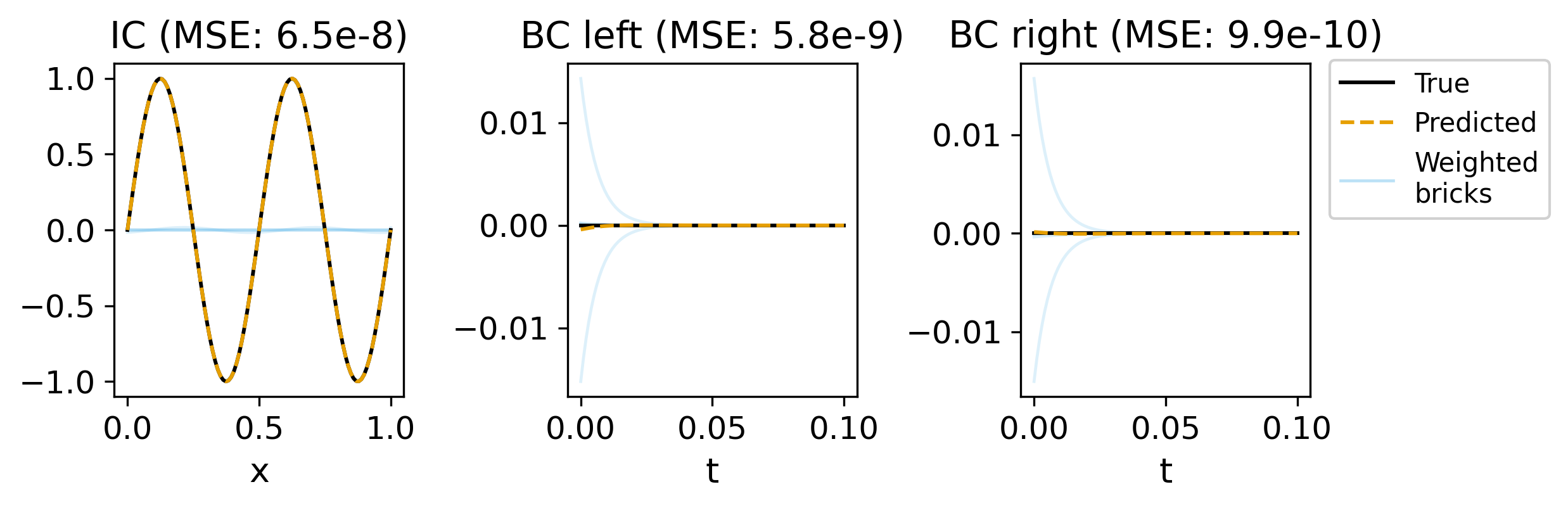}
        \caption{IBC fit.}
        \label{fig-app:liesolver_ibc_heat_sine}
    \end{subfigure}
        \caption{\LieSolver results for heat equation with \IC{Sine} IC.}
    \label{fig-app:liesolver_heat_sine}
\end{figure}

\begin{figure}
    \centering
    \begin{subfigure}{.25\linewidth}
        \centering
        \includegraphics[width=\linewidth]{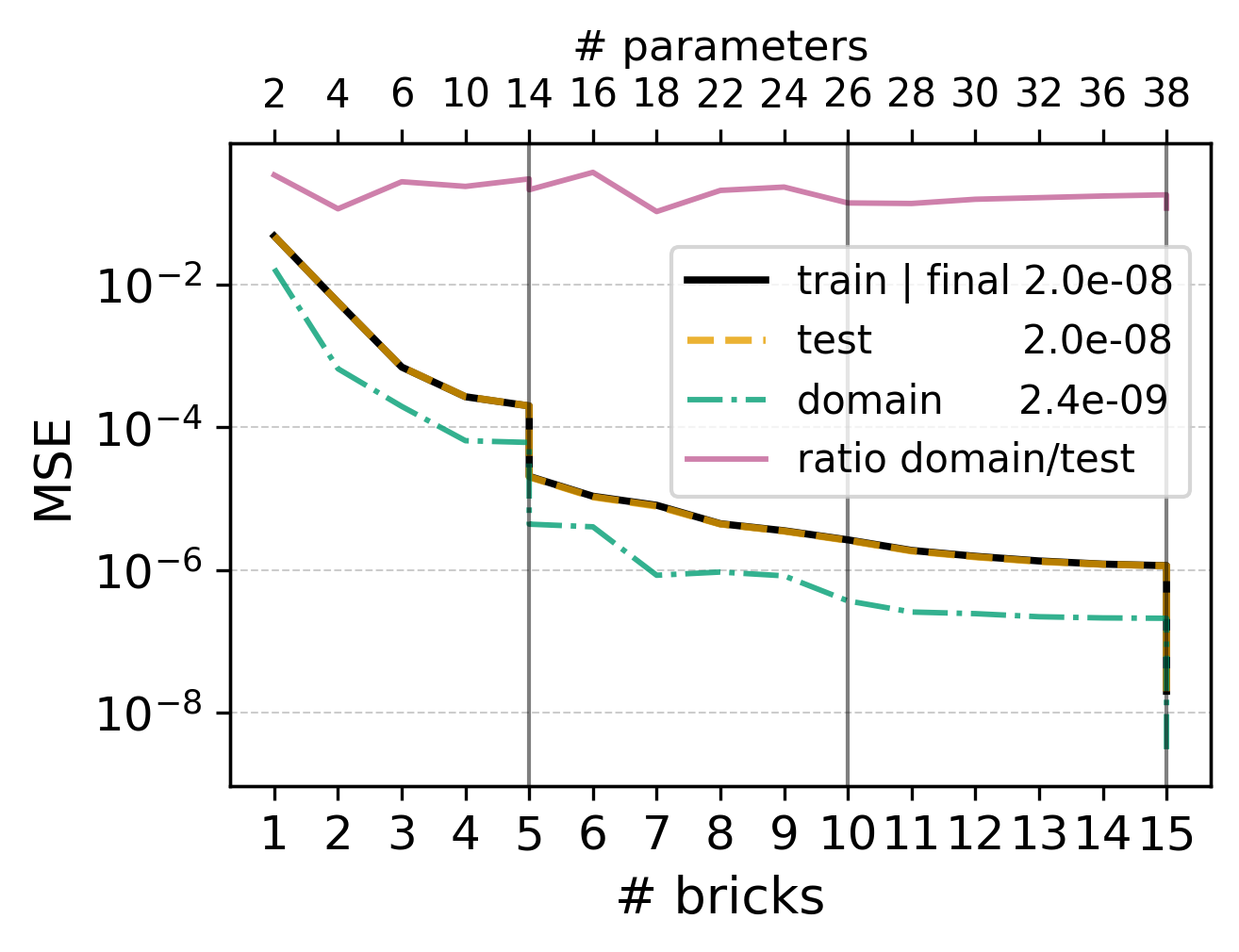}
        \caption{Fit progress.}
        \label{fig-app:liesolver_fit_heat_sinemix}
    \end{subfigure} %
    \begin{subfigure}{.33\linewidth}
        \centering
        \includegraphics[width=\linewidth]{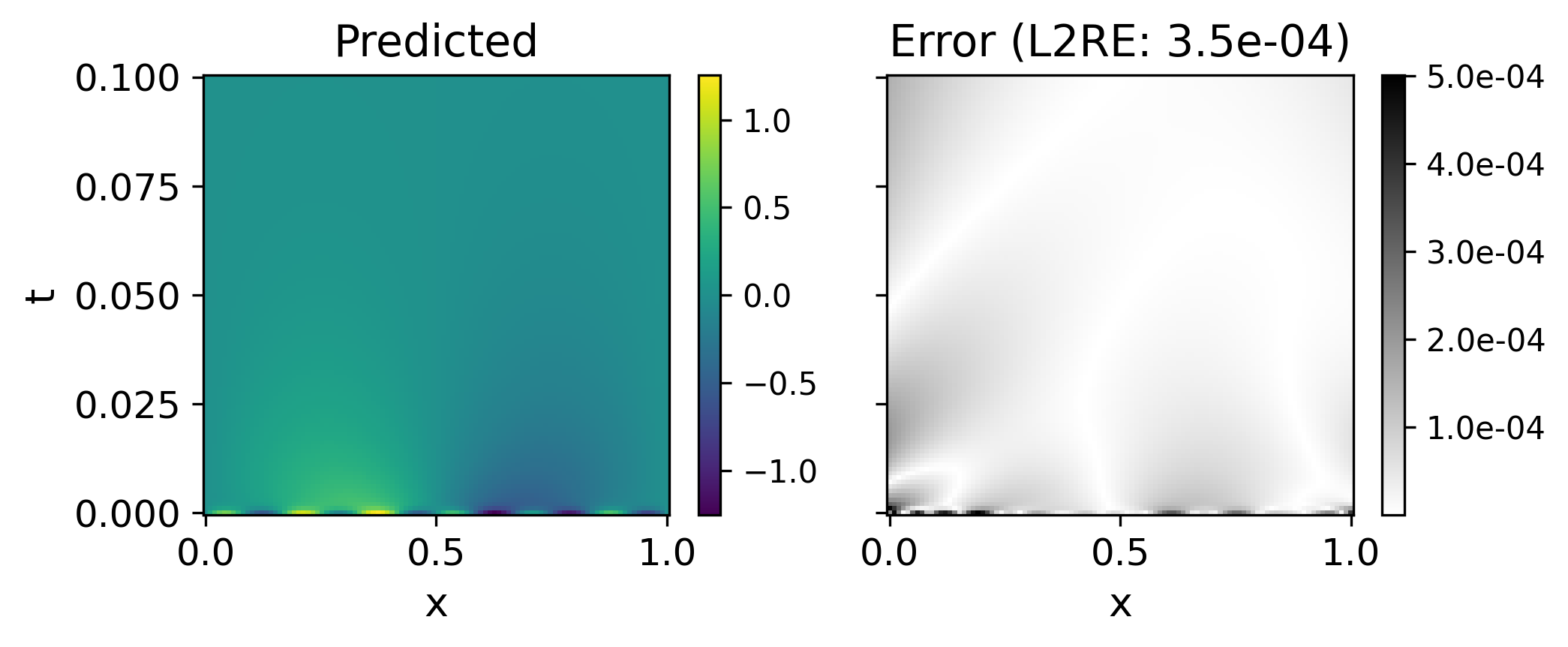}
        \caption{2D domain fit.}
        \label{fig-app:liesolver_domain_heat_sinemix}
    \end{subfigure}
    \begin{subfigure}{.41\linewidth}
        \centering
        \includegraphics[width=\linewidth]{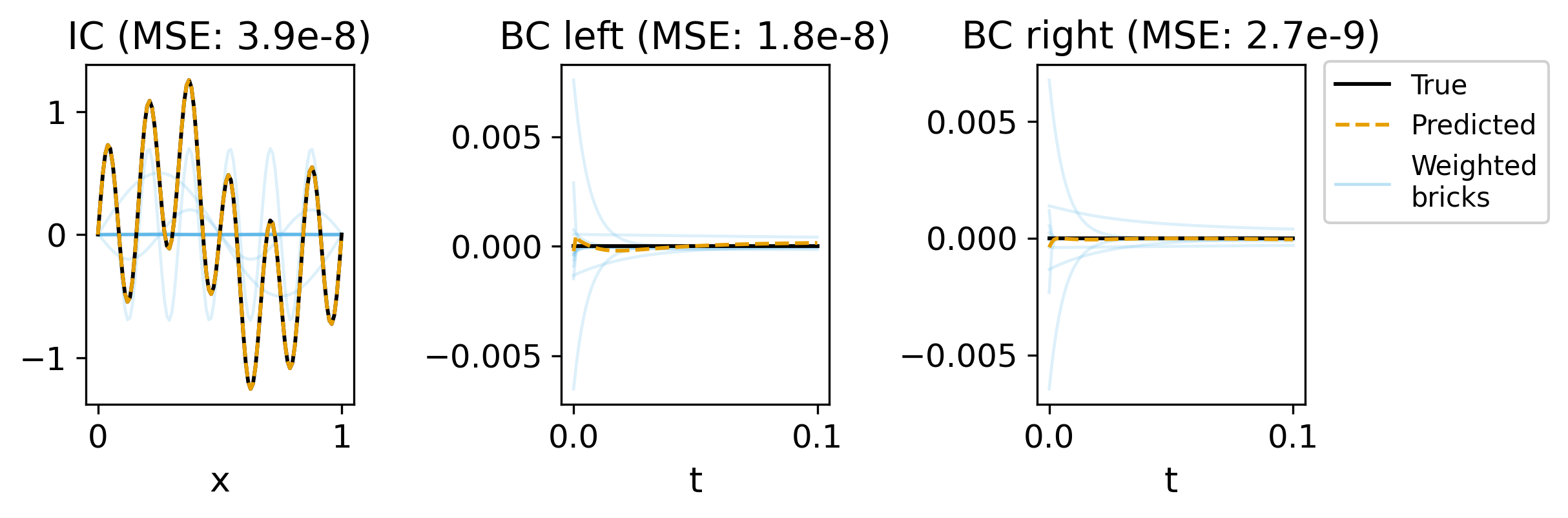}
        \caption{IBC fit.}
        \label{fig-app:liesolver_ibc_heat_sinemix}
    \end{subfigure}
        \caption{\LieSolver results for heat equation with \IC{Sine Mix} IC.}
    \label{fig-app:liesolver_heat_sinemix}
\end{figure}

\textbf{Heat \IC{Sine} and \IC{Sine Mix}:} As the set of brick families contains sine-generated bricks, \textsc{LieSolver} handles these cases (\Cref{fig-app:liesolver_heat_sine,fig-app:liesolver_heat_sinemix}) with high efficiency. Technically, a single sine brick suffices for the \IC{Sine} case due to the rapid dissipation of boundary values under homogeneous Dirichlet conditions. However the obtained results confirm that our generalized training pipeline robustly identifies these simple solutions without instability, fitting \IC{Sine} with 5 bricks and \IC{Sine Mix} with 15 bricks.

\begin{figure}[H]
    \centering
    \begin{subfigure}{.25\linewidth}
        \centering
        \includegraphics[width=\linewidth]{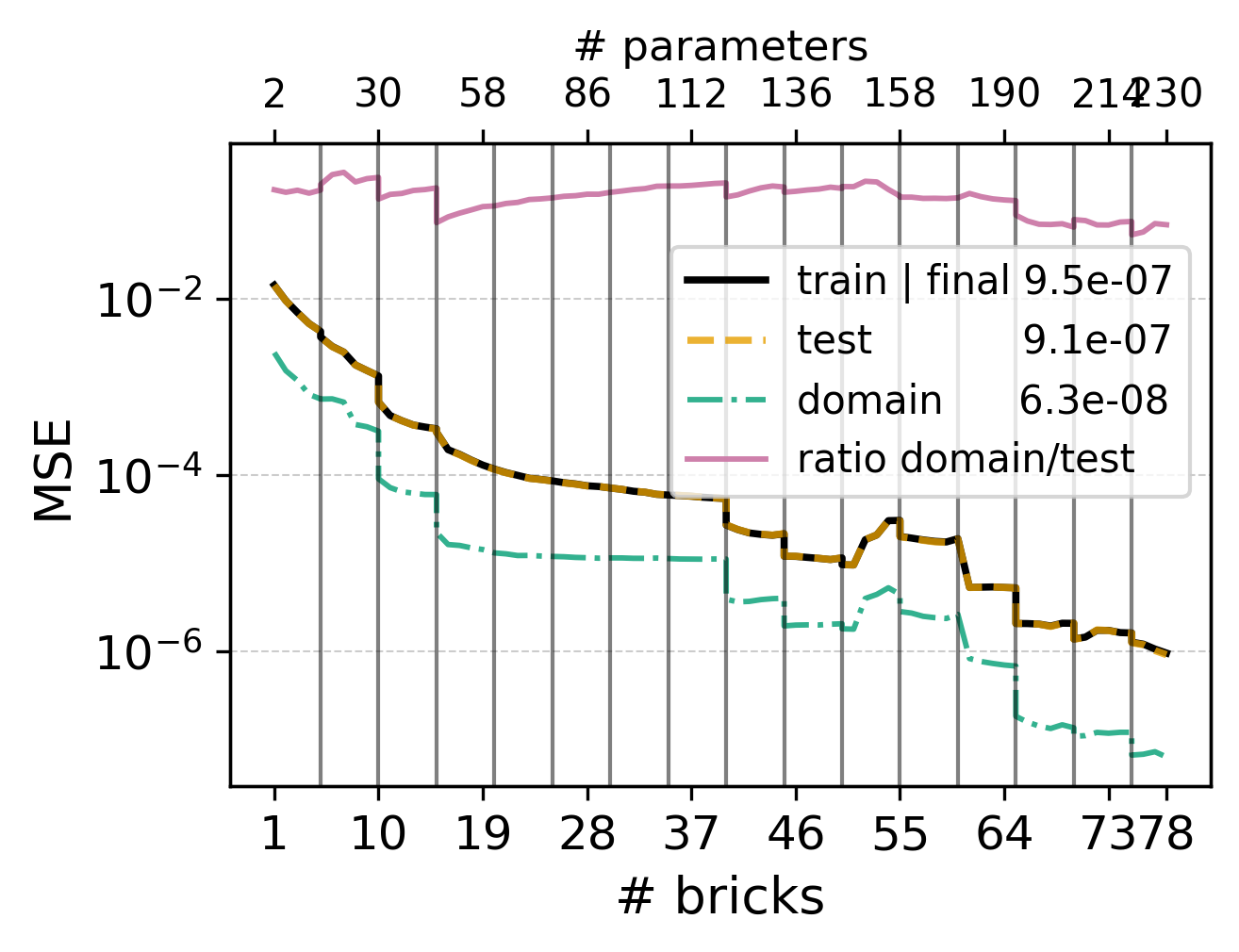}
        \caption{Fit progress.}
        \label{fig-app:liesolver_fit_heat_step}
    \end{subfigure} %
    \begin{subfigure}{.33\linewidth}
        \centering
        \includegraphics[width=\linewidth]{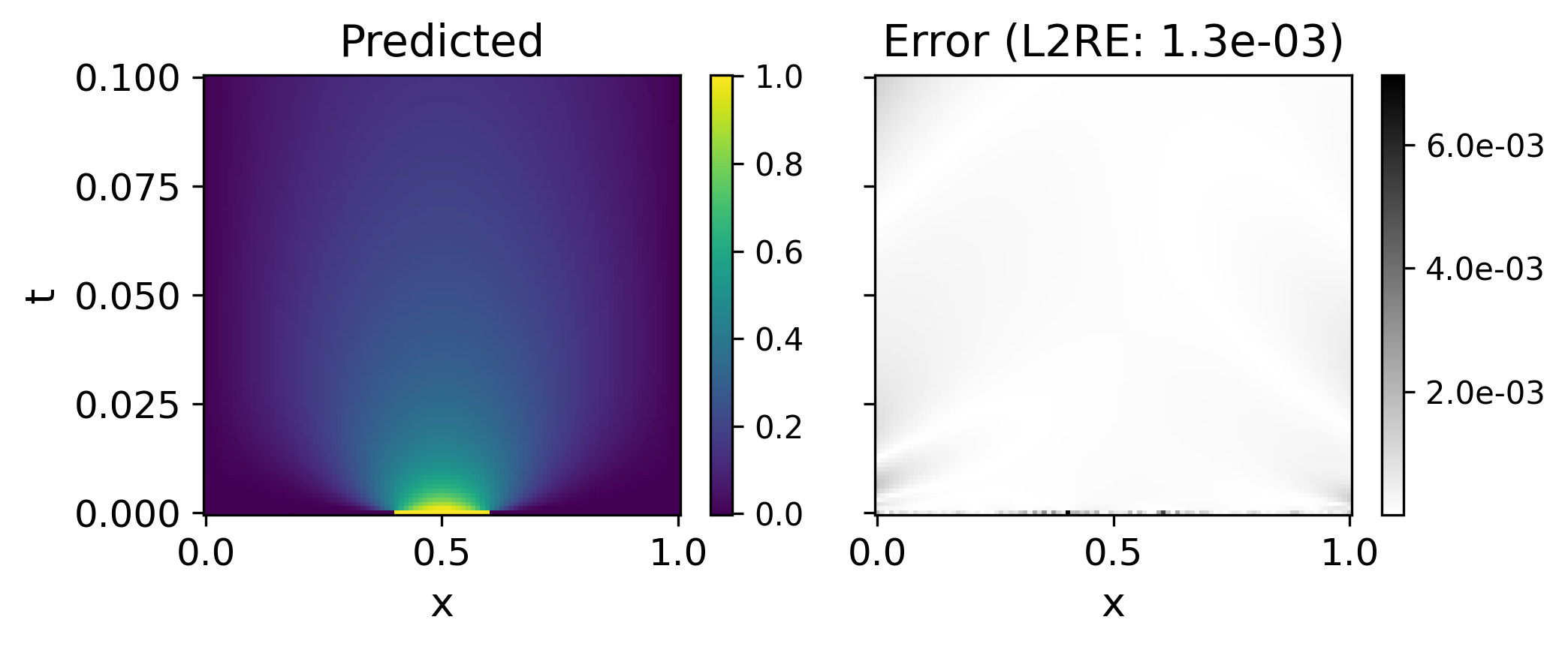}
        \caption{2D domain fit.}
        \label{fig-app:liesolver_domain_heat_step}
    \end{subfigure}
    \begin{subfigure}{.41\linewidth}
        \centering
        \includegraphics[width=\linewidth]{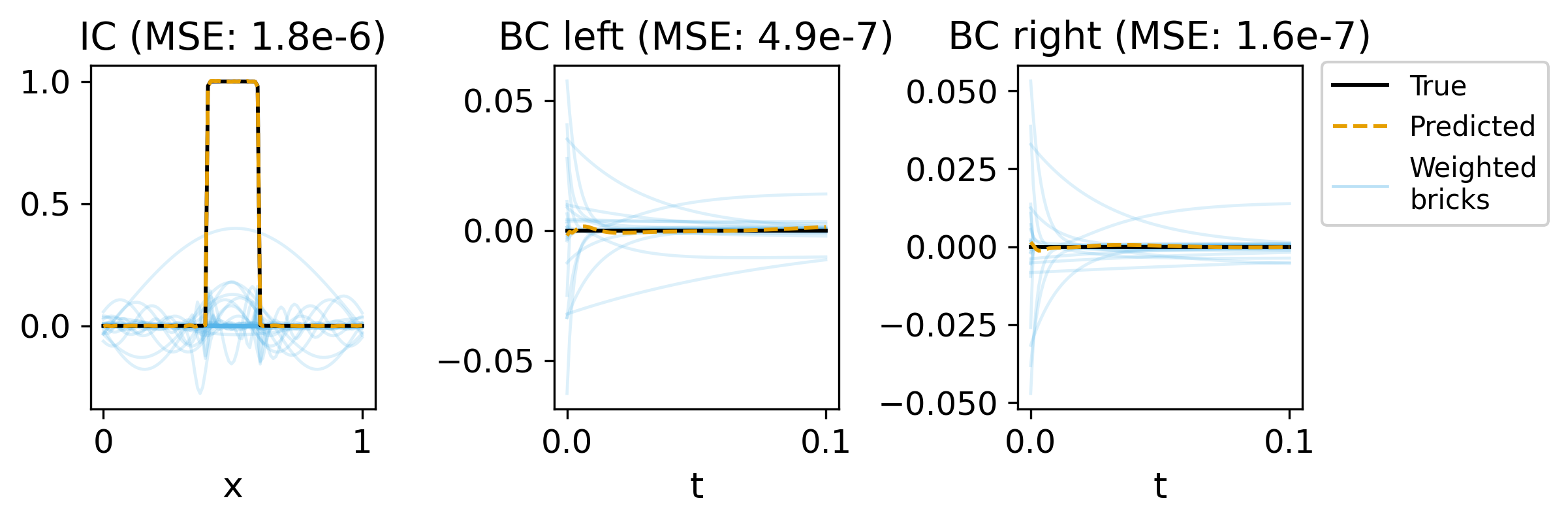}
        \caption{IBC fit.}
        \label{fig-app:liesolver_ibc_heat_step}
    \end{subfigure}
        \caption{\LieSolver results for heat equation with \IC{Step} IC.}
    \label{fig-app:liesolver_heat_step}
\end{figure}

\textbf{Heat \IC{Step}:} This profile (\Cref{fig-app:liesolver_heat_step}) represents the most significant challenge due to the sharp jump. While a coarse fit ($10^{-5}$ MSE) is achieved in approximately 100 seconds and 40 bricks, reaching the target $10^{-6}$ requires over 300 seconds and nearly 80 bricks. Similar to the \IC{Poly} case, the fit history (\Cref{fig-app:liesolver_fit_heat_step}) displays non-monotonic behaviour during greedy initialization, which is eventually corrected with global refinement upon increase of model expressivity.

While PINNs perform adequately on smooth profiles, they exhibit significant failure modes on the \IC{step} IC. As seen in \Cref{fig-app:pinn_heat_step}, the transition from an extremely sharp initial condition to a rapidly smoothening solution in the domain proves difficult for the network to resolve. Even with increased network capacity, the PINN cannot resolve the transition, leading to high metrics.

\begin{figure}[!htbp]
    \centering
    \begin{subfigure}{.25\linewidth}
        \centering
        \includegraphics[width=\linewidth]{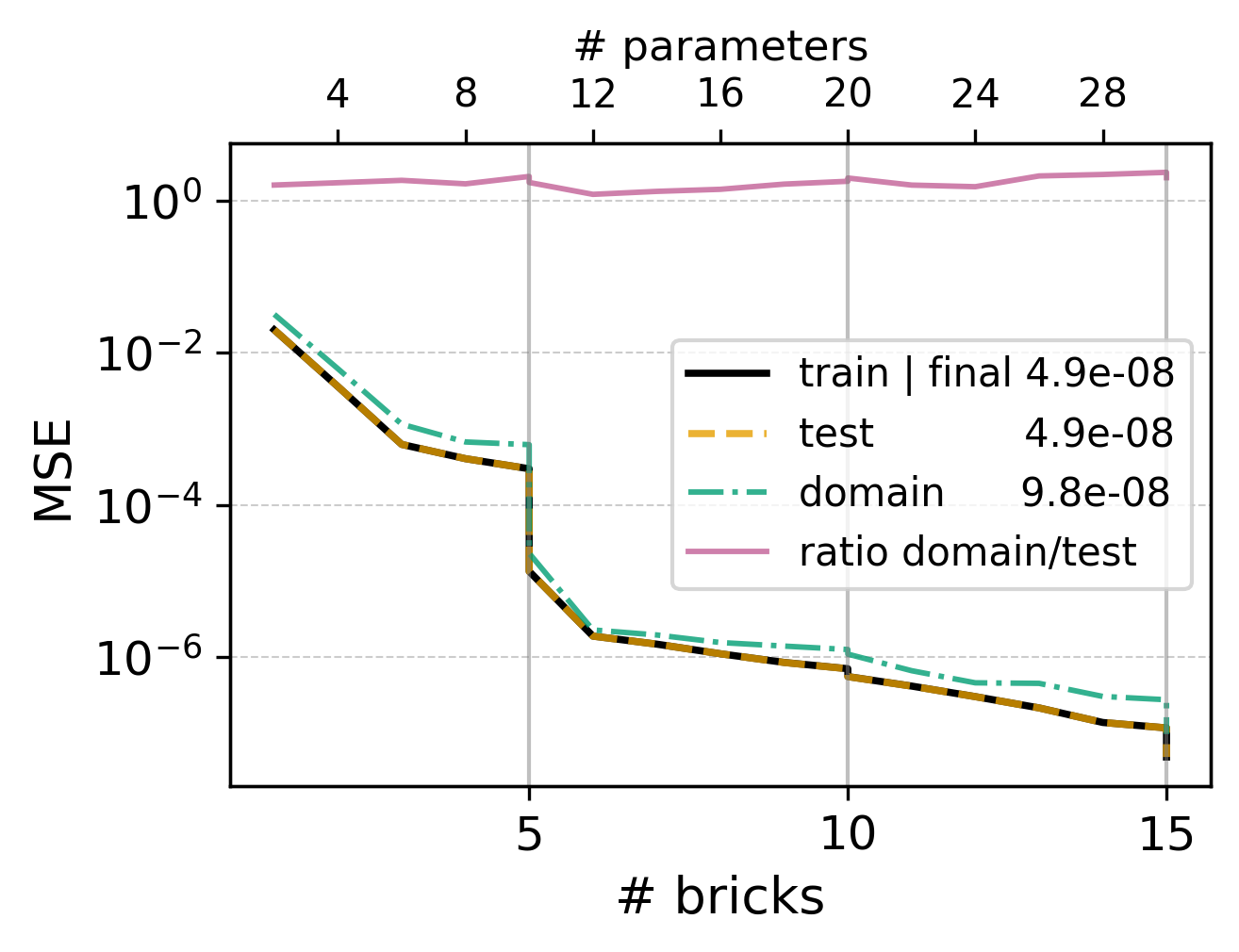}
        \caption{Fit progress.}
        \label{fig-app:liesolver_fit_wave_gauss}
    \end{subfigure} %
    \begin{subfigure}{.33\linewidth}
        \centering
        \includegraphics[width=\linewidth]{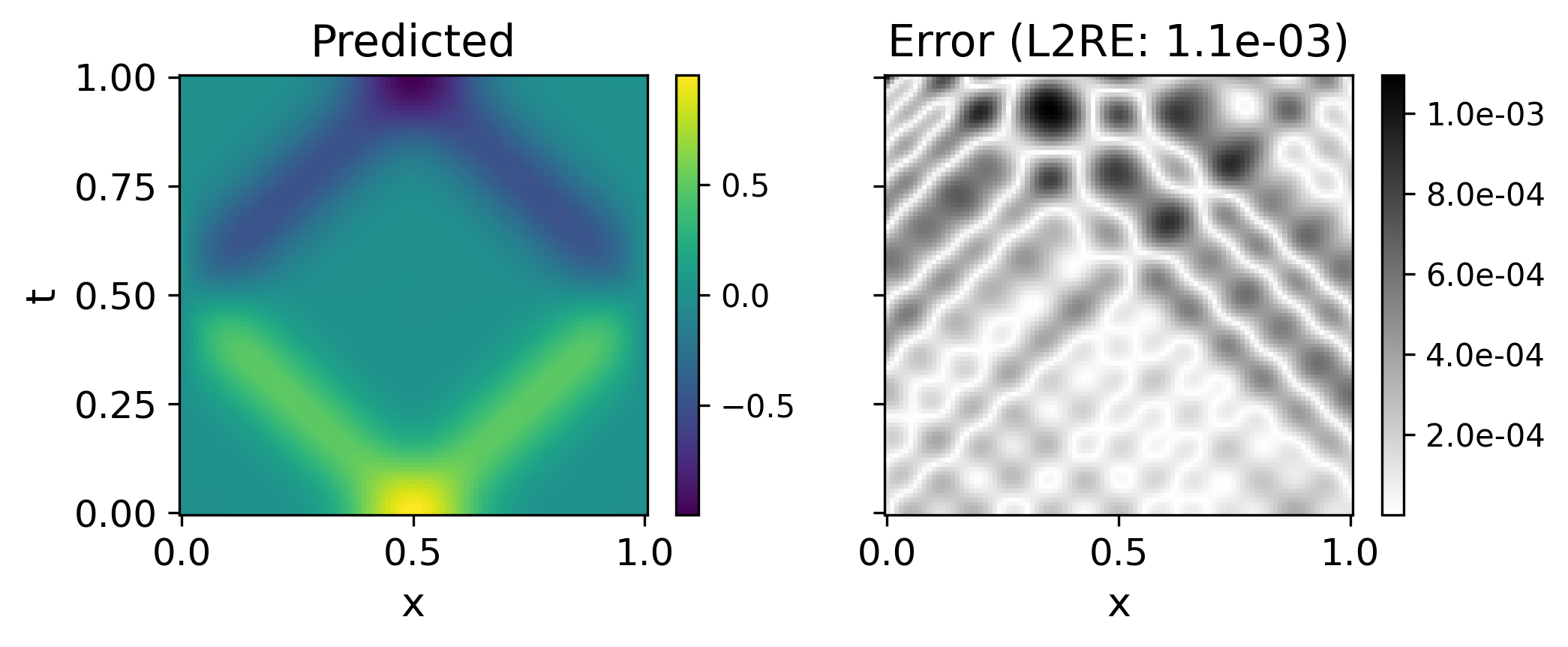}
        \caption{2D domain fit.}
        \label{fig-app:liesolver_domain_wave_gauss}
    \end{subfigure}
    \begin{subfigure}{.41\linewidth}
        \centering
        \includegraphics[width=\linewidth]{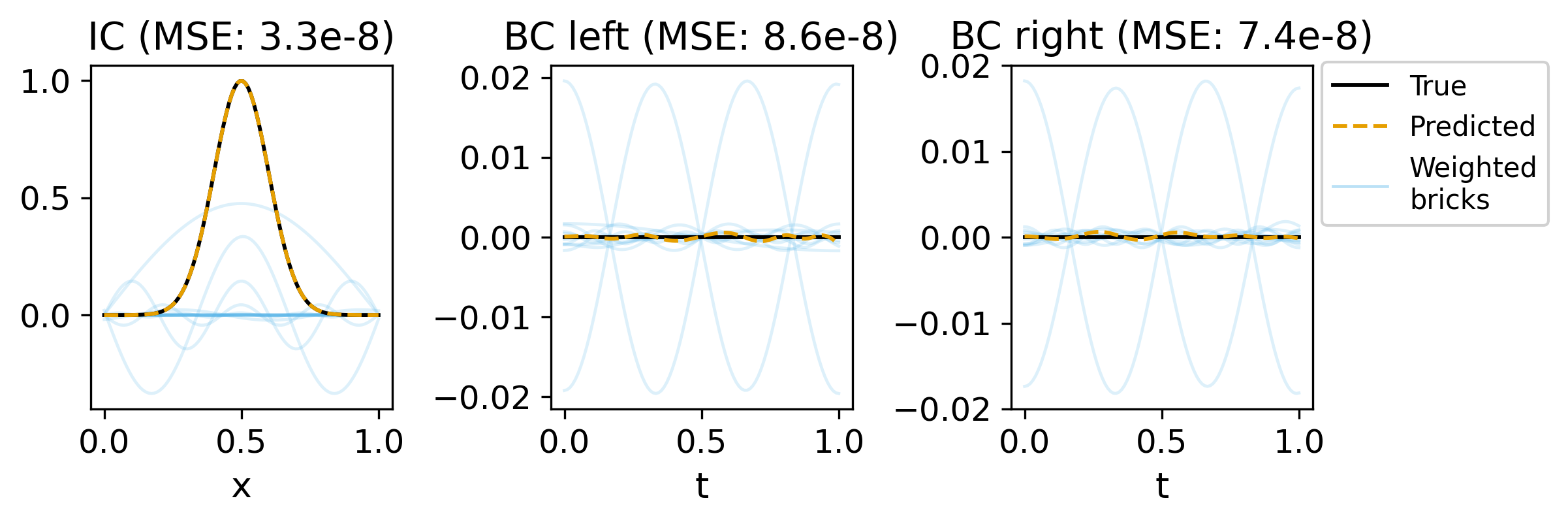}
        \caption{IBC fit.}
        \label{fig-app:liesolver_ibc_wave_gauss}
    \end{subfigure}
        \caption{\LieSolver results for wave equation with \IC{Gauss} IC.}
    \label{fig-app:liesolver_wave_gauss}
\end{figure}

\begin{figure}[!htbp]
    \centering
    \begin{subfigure}{.25\linewidth}
        \centering
        \includegraphics[width=\linewidth]{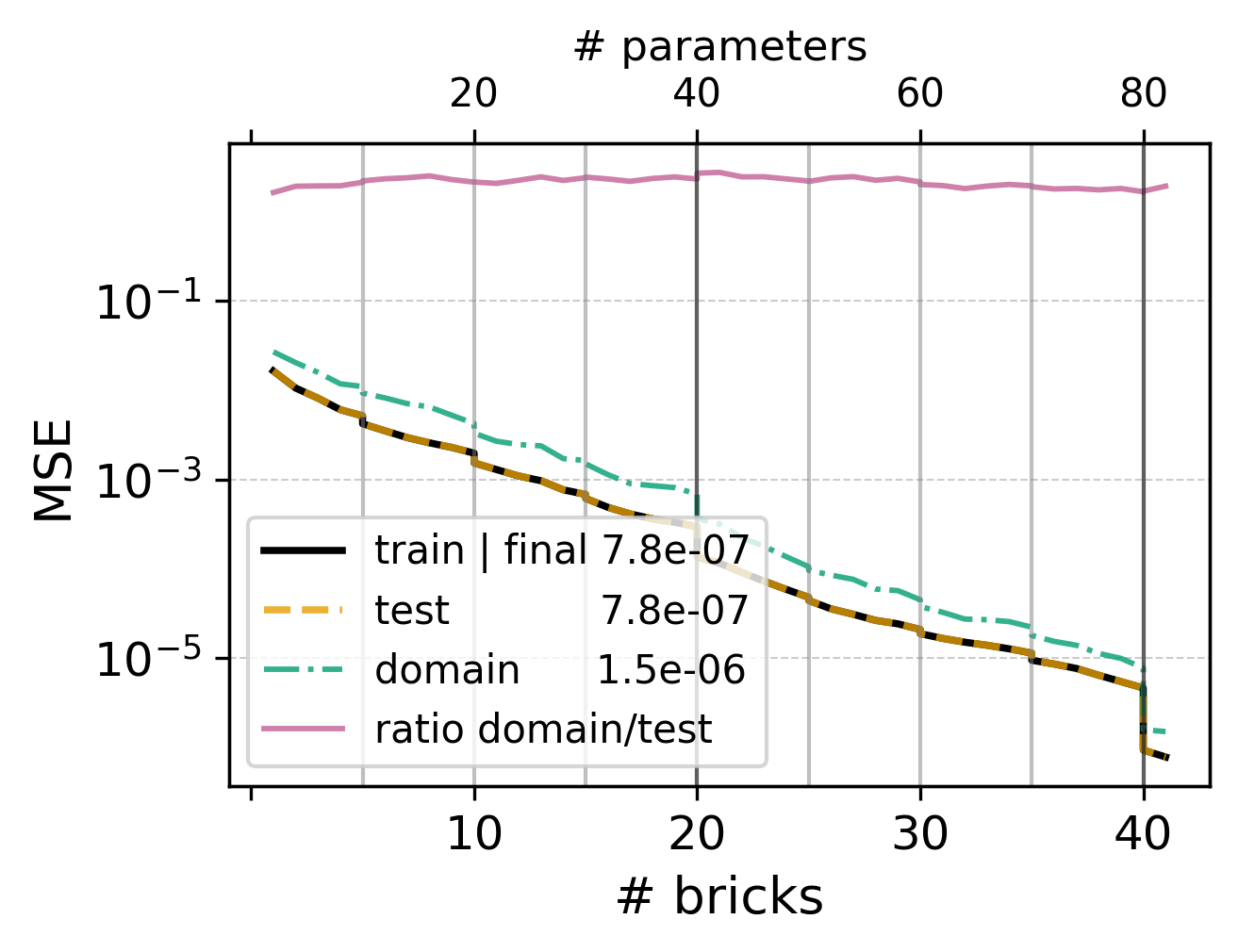}
        \caption{Fit progress.}
        \label{fig-app:liesolver_fit_wave_gaussmix}
    \end{subfigure} %
    \begin{subfigure}{.33\linewidth}
        \centering
        \includegraphics[width=\linewidth]{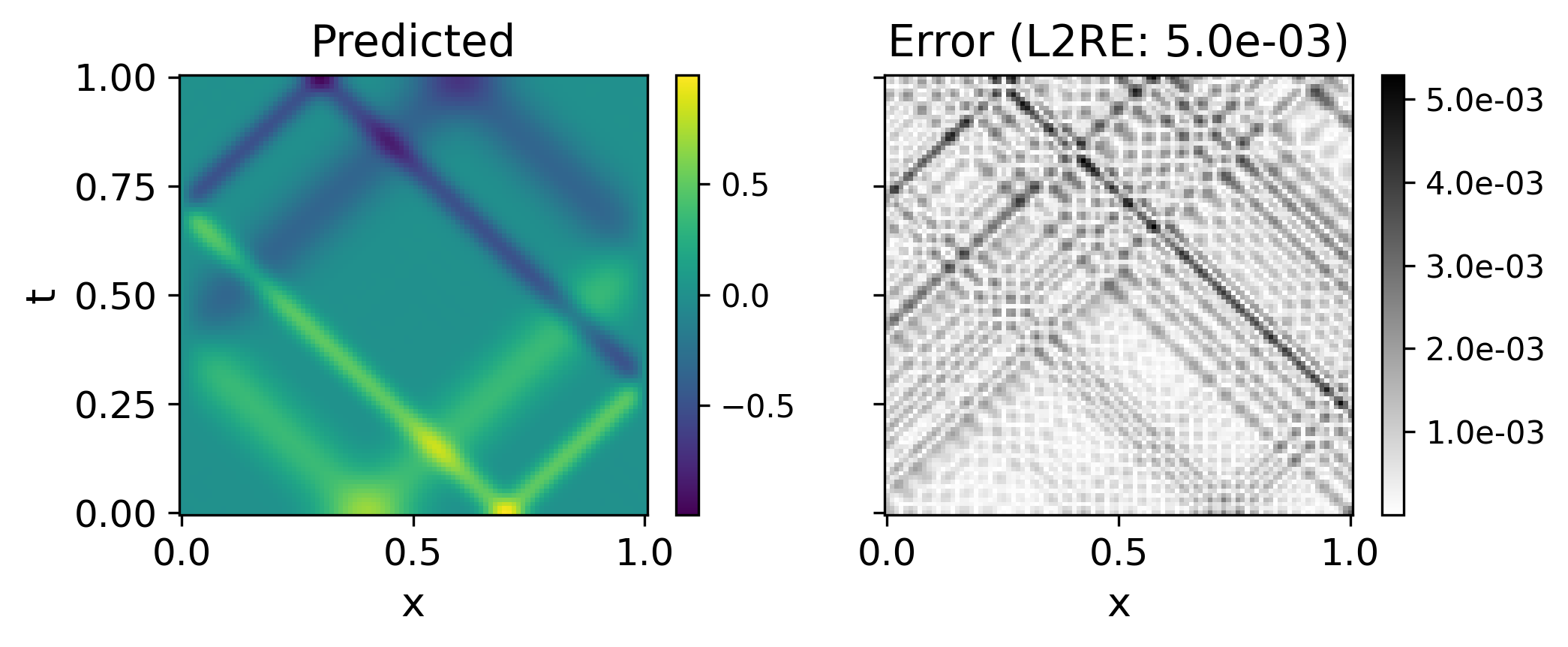}
        \caption{2D domain fit.}
        \label{fig-app:liesolver_domain_wave_gaussmix}
    \end{subfigure}
    \begin{subfigure}{.41\linewidth}
        \centering
        \includegraphics[width=\linewidth]{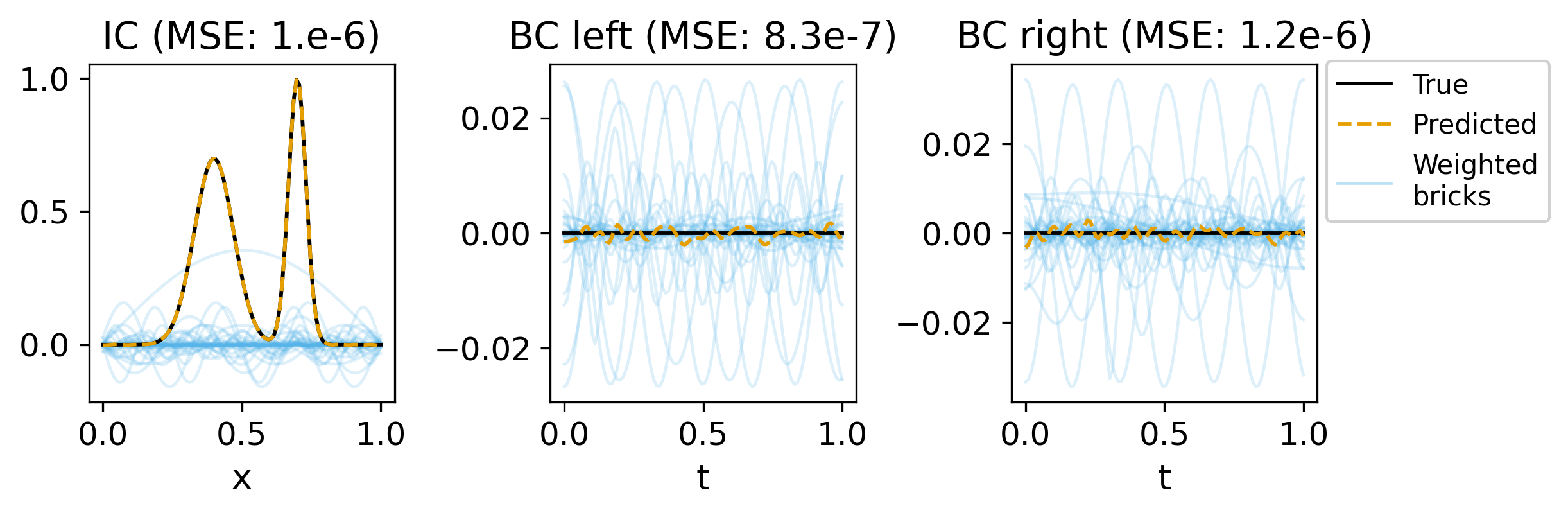}
        \caption{IBC fit.}
        \label{fig-app:liesolver_ibc_wave_gaussmix}
    \end{subfigure}
        \caption{\LieSolver results for wave equation with \IC{Gauss Mix} IC.}
    \label{fig-app:liesolver_wave_gaussmix}
\end{figure}

\begin{figure}[!htbp]
    \centering
    \begin{subfigure}{.25\linewidth}
        \centering
        \includegraphics[width=\linewidth]{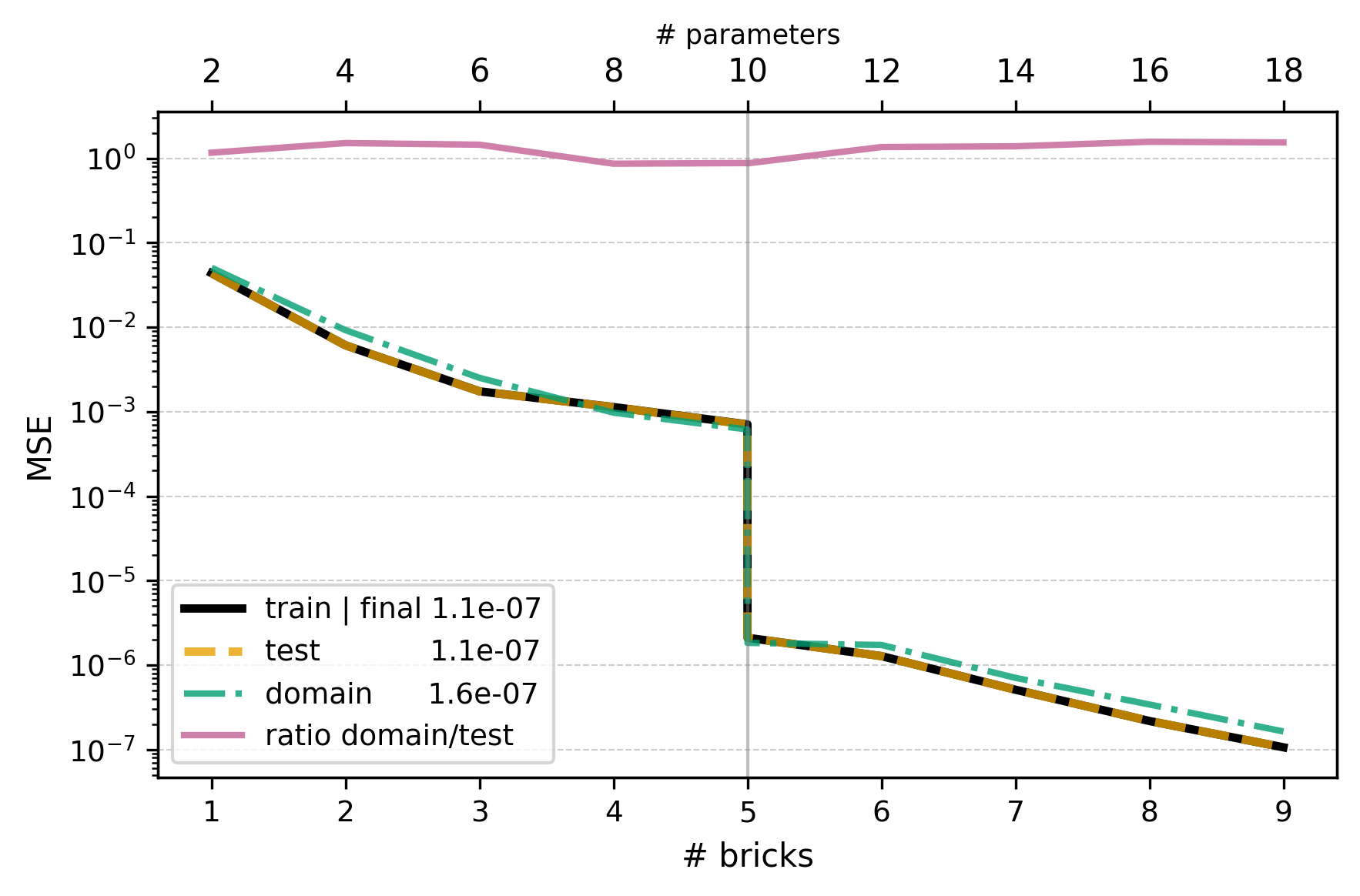}
        \caption{Fit progress.}
        \label{fig-app:liesolver_fit_wave_sine}
    \end{subfigure} %
    \begin{subfigure}{.33\linewidth}
        \centering
        \includegraphics[width=\linewidth]{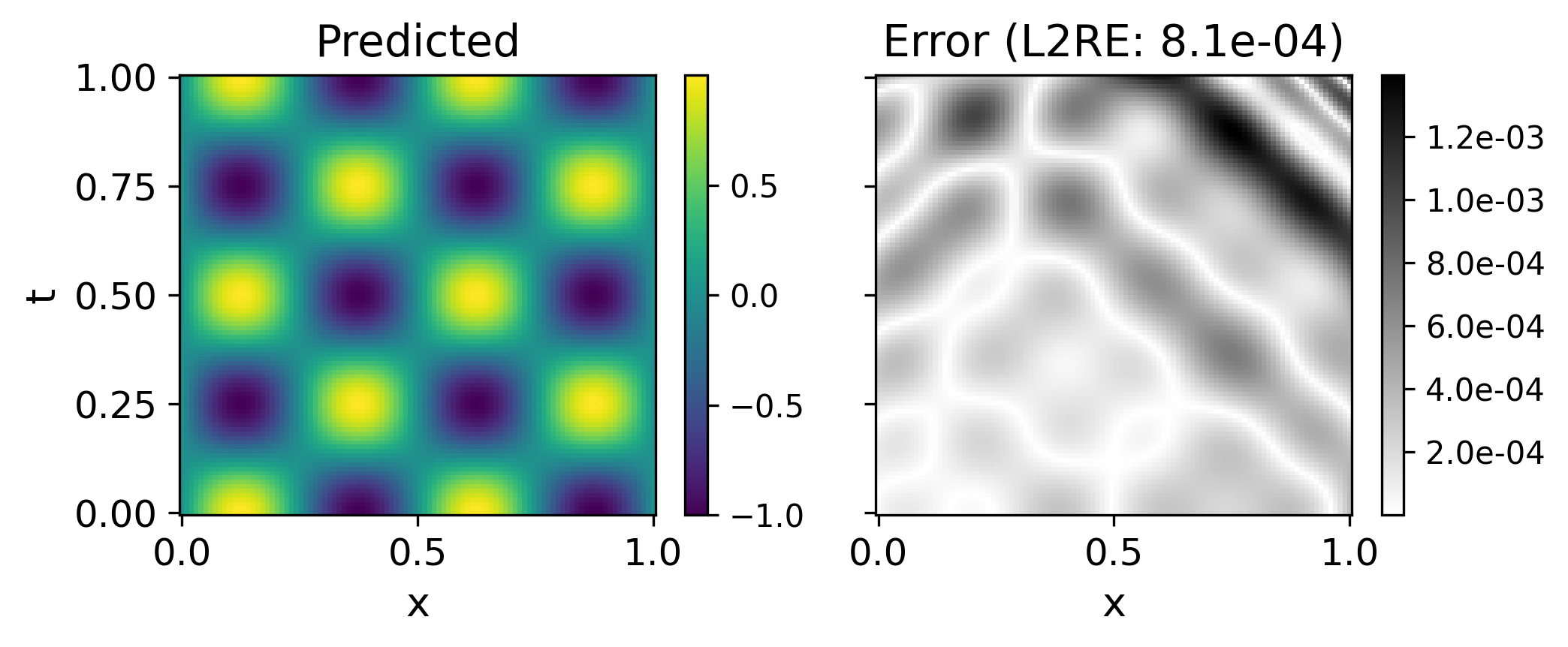}
        \caption{2D domain fit.}
        \label{fig-app:liesolver_domain_wave_sine}
    \end{subfigure}
    \begin{subfigure}{.41\linewidth}
        \centering
        \includegraphics[width=\linewidth]{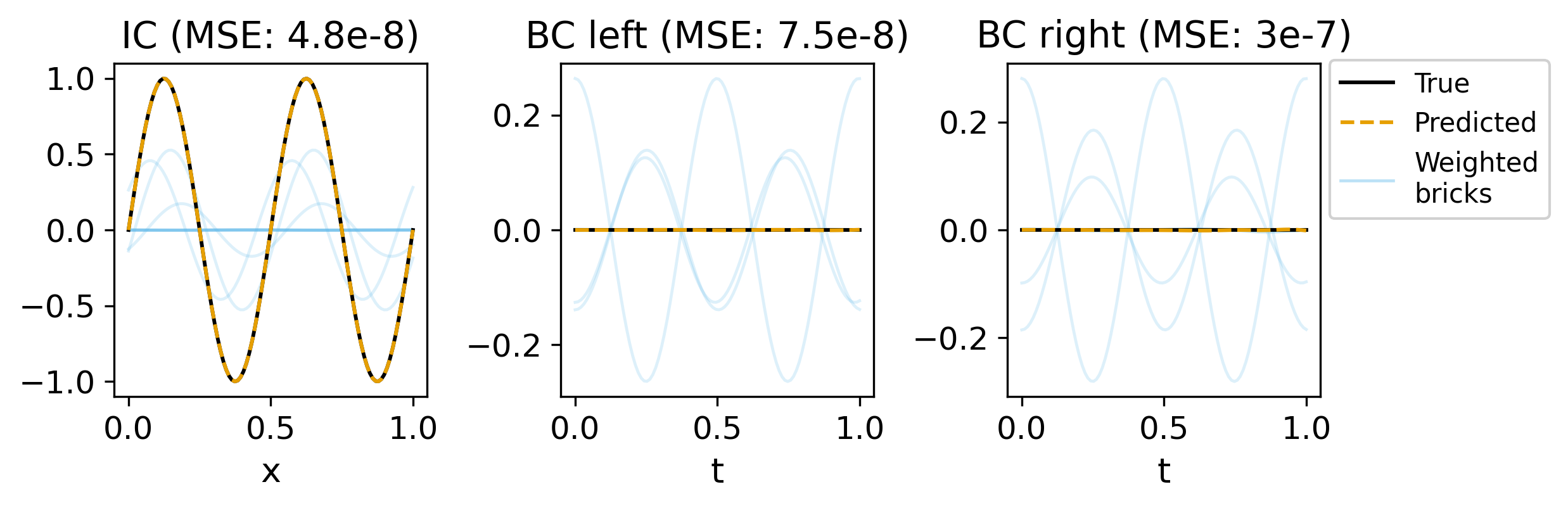}
        \caption{IBC fit.}
        \label{fig-app:liesolver_ibc_wave_sine}
    \end{subfigure}
        \caption{\LieSolver results for wave equation with \IC{Sine} IC.}
    \label{fig-app:liesolver_wave_sine}
\end{figure}

\begin{figure}[!htbp]
    \centering
    \begin{subfigure}{.25\linewidth}
        \centering
        \includegraphics[width=\linewidth]{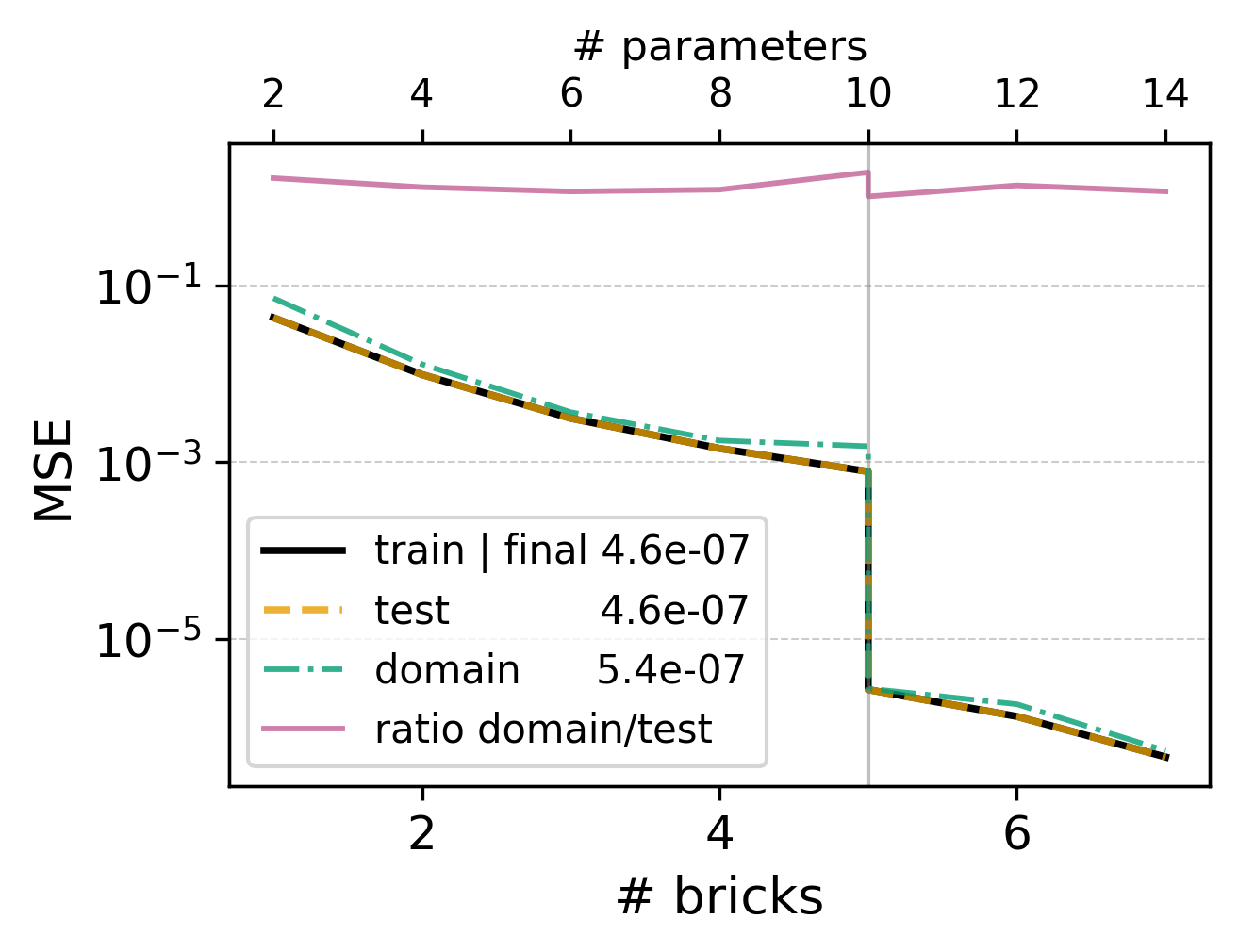}
        \caption{Fit progress.}
        \label{fig-app:liesolver_fit_wave_sinemix}
    \end{subfigure} %
    \begin{subfigure}{.33\linewidth}
        \centering
        \includegraphics[width=\linewidth]{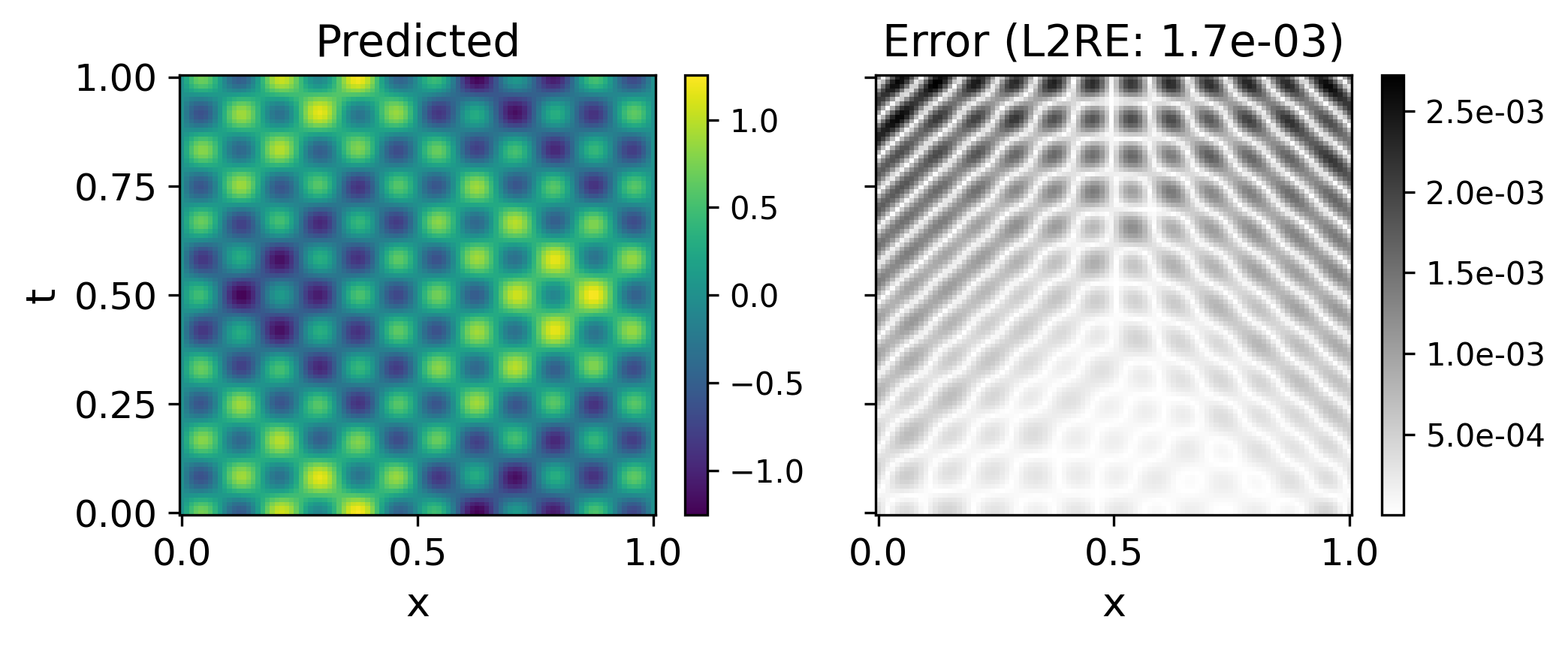}
        \caption{2D domain fit.}
        \label{fig-app:liesolver_domain_wave_sinemix}
    \end{subfigure}
    \begin{subfigure}{.41\linewidth}
        \centering
        \includegraphics[width=\linewidth]{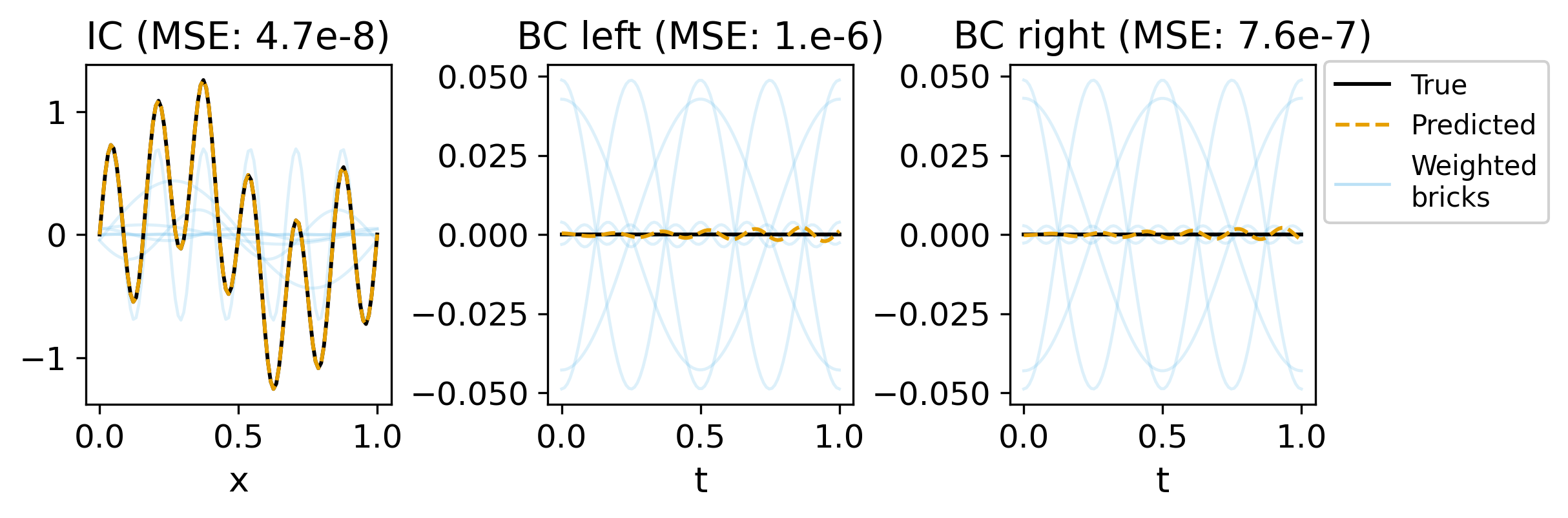}
        \caption{IBC fit.}
        \label{fig-app:liesolver_ibc_wave_sinemix}
    \end{subfigure}
        \caption{\LieSolver results for wave equation with \IC{Sine Mix} IC.}
    \label{fig-app:liesolver_wave_sinemix}
\end{figure}

\clearpage
\textbf{Wave Equation Cases:} For the wave equation, we observe a consistent structural learning pattern across the \IC{gauss}, \IC{gauss mix}, \IC{sine}, and \IC{sine mix} cases (\Cref{fig-app:liesolver_wave_gauss,fig-app:liesolver_wave_gaussmix,fig-app:liesolver_wave_sine,fig-app:liesolver_wave_sinemix}). The greedy initialization predominantly spawns sine-wave bricks $f^1$ first, as these patterns naturally fit the oscillatory nature of the wave equation's solutions, before refining the fit with localized Gaussian corrections where necessary.

\clearpage
\subsection{PINN Results}
\label{app-ssec:pinn_results}
\begin{figure}[!htbp]
    \centering
    \begin{subfigure}{.48\linewidth}
        \centering
        \includegraphics[width=\linewidth]{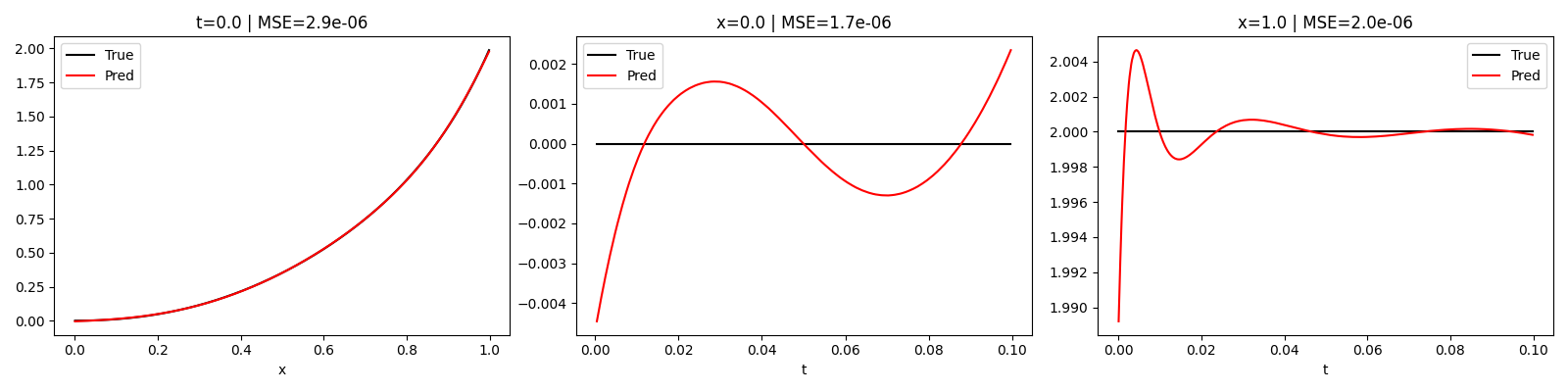}
        \caption{IBC fit.}
        \label{fig-app:pinn_heat_poly_ibc}
    \end{subfigure} %
    \begin{subfigure}{.48\linewidth}
        \centering
        \includegraphics[width=\linewidth]{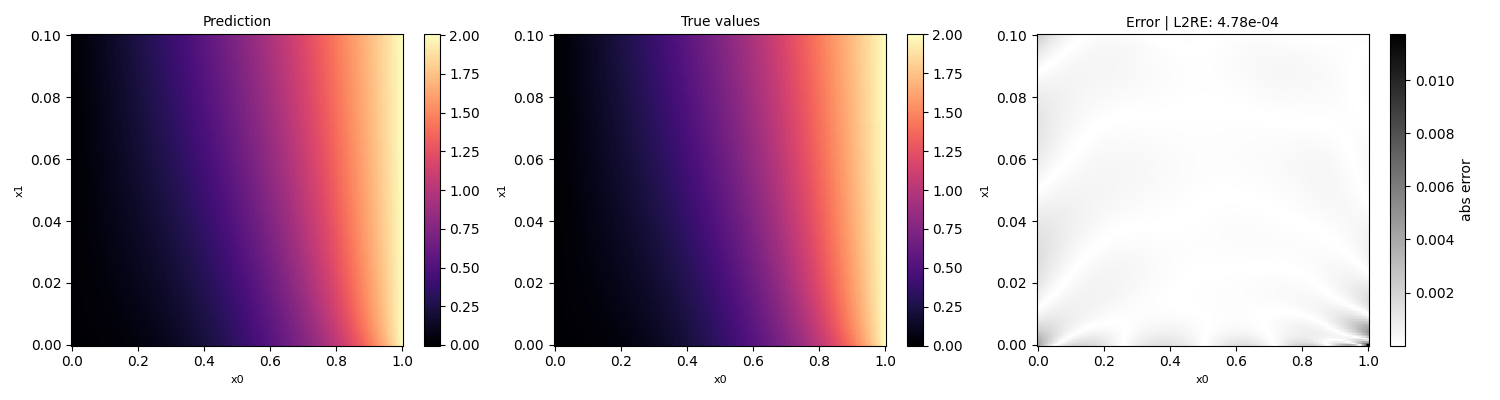}
        \caption{2D domain fit.}
        \label{fig-app:pinn_heat_poly_domain}
    \end{subfigure}
        \caption{PINN results for heat equation with \IC{Poly} IC.}
    \label{fig-app:pinn_heat_poly}
\end{figure}

\begin{figure}[!htbp]
    \centering
    \begin{subfigure}{.48\linewidth}
        \centering
        \includegraphics[width=\linewidth]{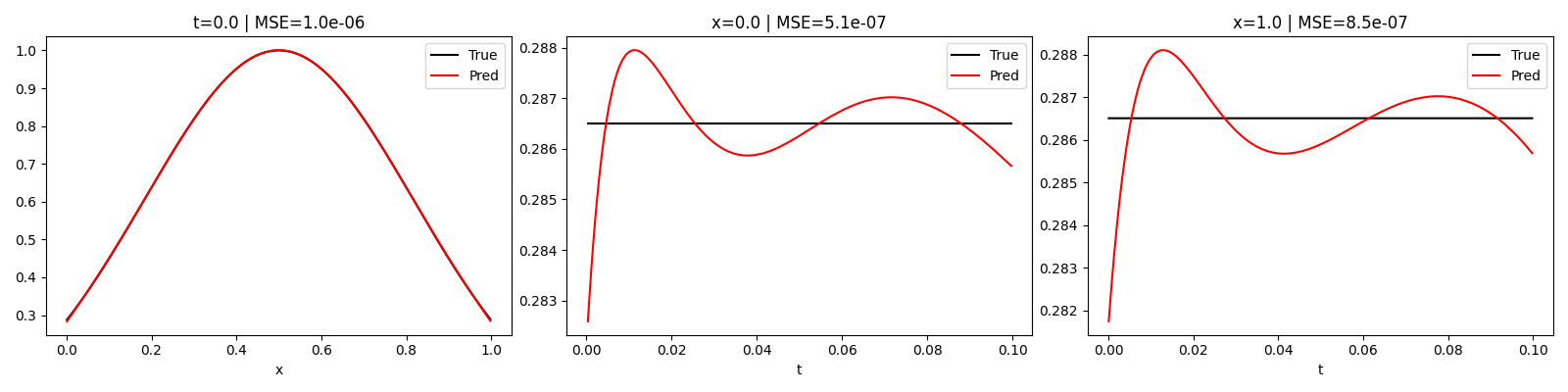}
        \caption{IBC fit.}
        \label{fig-app:pinn_heat_gauss_ibc}
    \end{subfigure} %
    \begin{subfigure}{.48\linewidth}
        \centering
        \includegraphics[width=\linewidth]{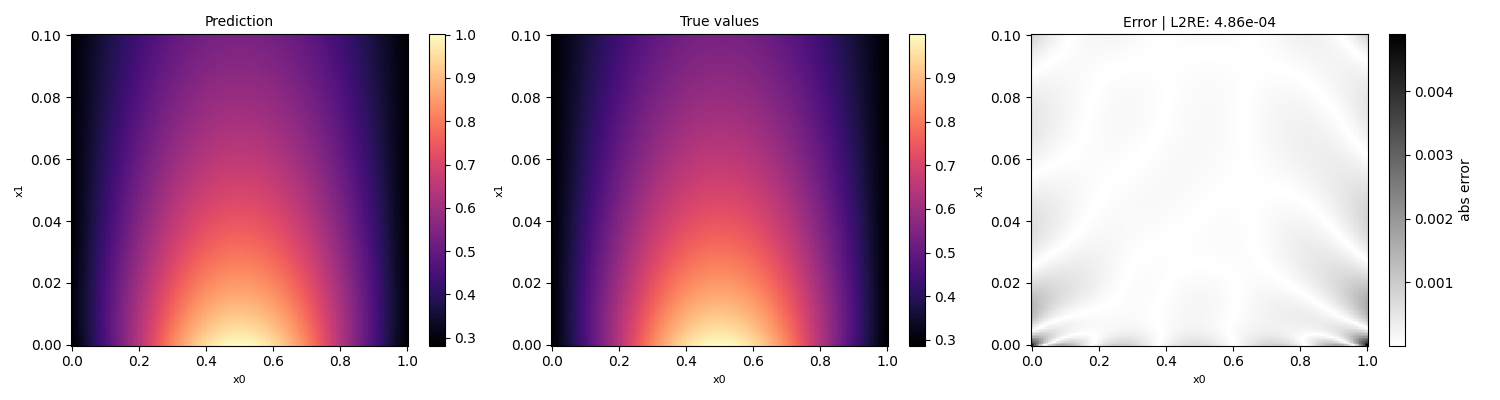}
        \caption{2D domain fit.}
        \label{fig-app:pinn_heat_gauss_domain}
    \end{subfigure}
    \caption{PINN results for heat equation with \IC{Gauss} IC.}
    \label{fig-app:pinn_heat_gauss}
\end{figure}

\begin{figure}[!htbp]
    \centering
    \begin{subfigure}{.48\linewidth}
        \centering
        \includegraphics[width=\linewidth]{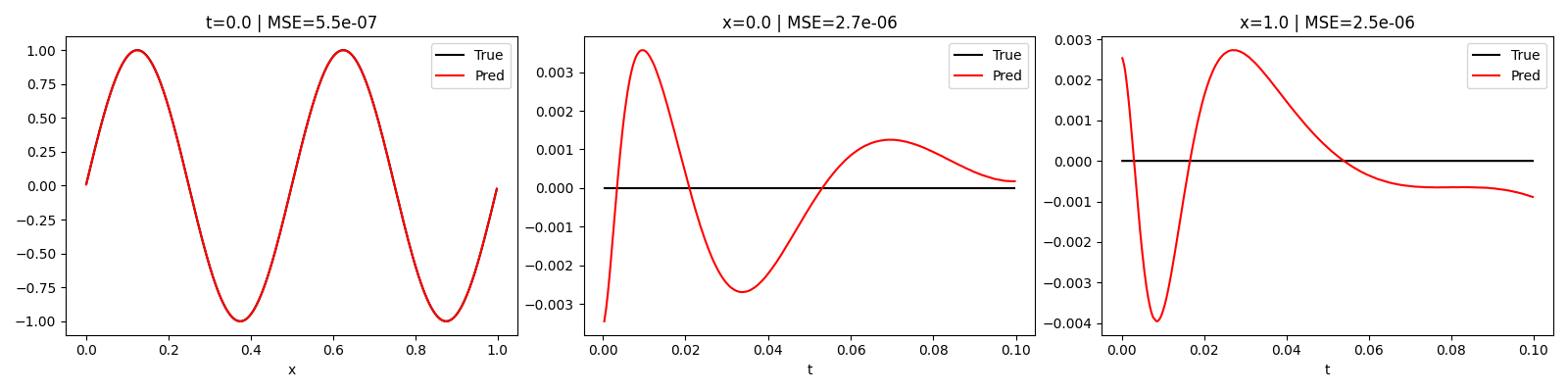}
        \caption{IBC fit.}
        \label{fig-app:pinn_heat_sine_ibc}
    \end{subfigure} %
    \begin{subfigure}{.48\linewidth}
        \centering
        \includegraphics[width=\linewidth]{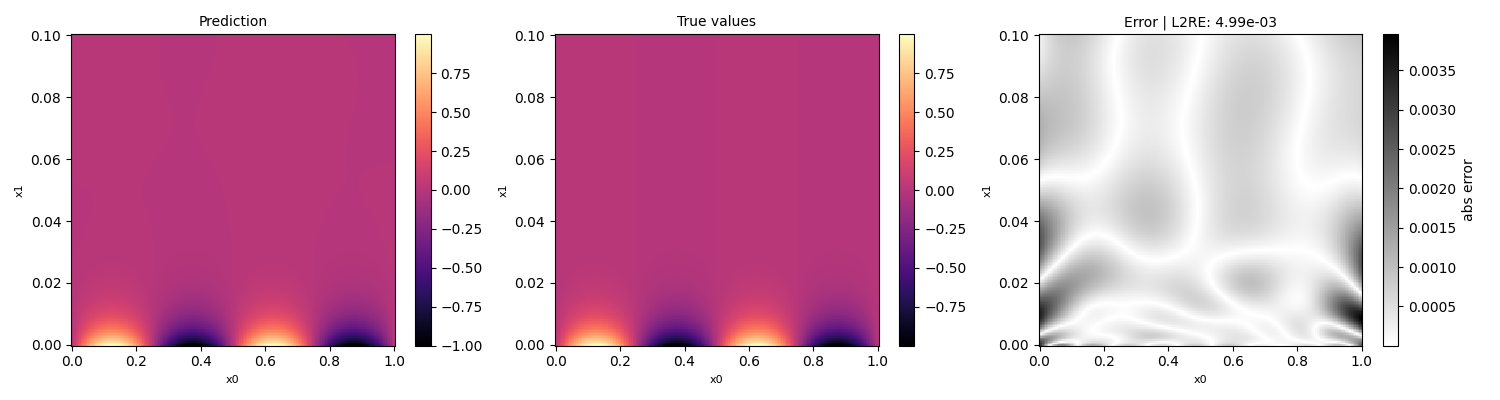}
        \caption{2D domain fit.}
        \label{fig-app:pinn_heat_sine_domain}
    \end{subfigure}
    \caption{PINN results for heat equation with \IC{Sine} IC.}
    \label{fig-app:pinn_heat_sine}
\end{figure}

\begin{figure}[!htbp]
    \centering
    \begin{subfigure}{.48\linewidth}
        \centering
        \includegraphics[width=\linewidth]{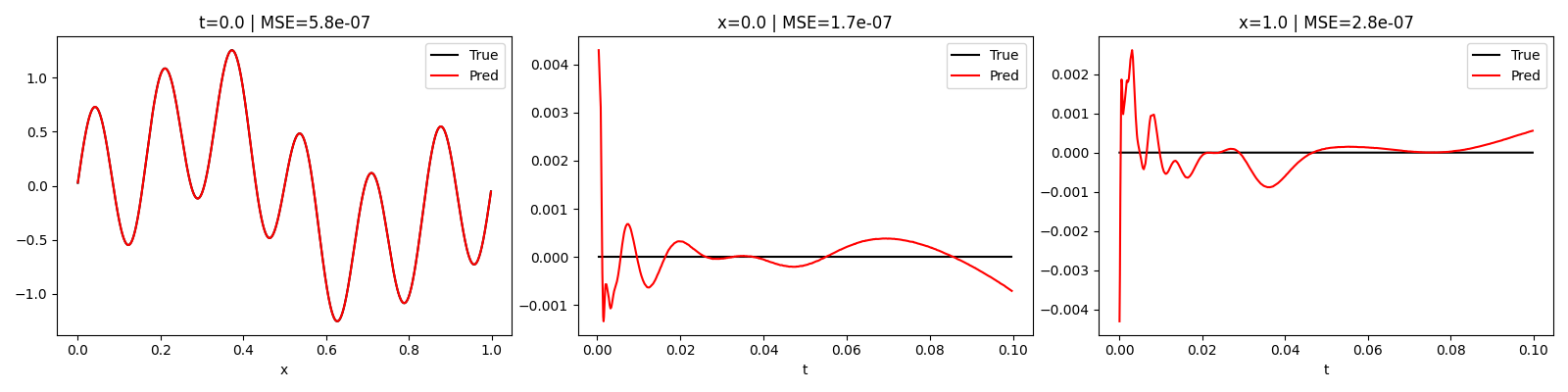}
        \caption{IBC fit.}
        \label{fig-app:pinn_heat_sinemix_ibc}
    \end{subfigure} %
    \begin{subfigure}{.48\linewidth}
        \centering
        \includegraphics[width=\linewidth]{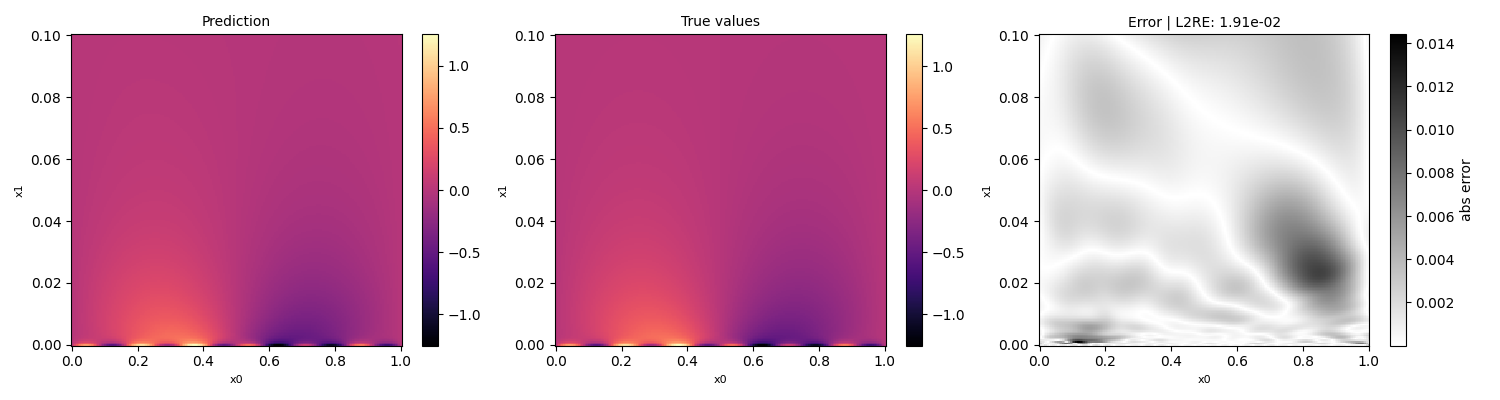}
        \caption{2D domain fit.}
        \label{fig-app:pinn_heat_sinemix_domain}
    \end{subfigure}
    \caption{PINN results for heat equation with \IC{Sine Mix} IC.}
    \label{fig-app:pinn_heat_sinemix}
\end{figure}

\begin{figure}[!htbp]
    \centering
    \begin{subfigure}{.48\linewidth}
        \centering
        \includegraphics[width=\linewidth]{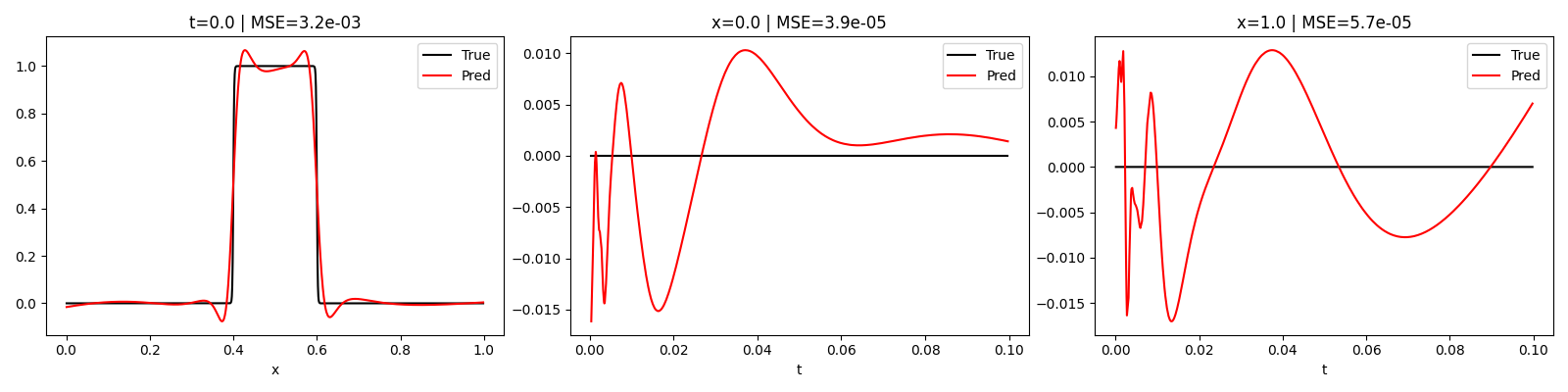}
        \caption{IBC fit.}
        \label{fig-app:pinn_heat_step_ibc}
    \end{subfigure} %
    \begin{subfigure}{.48\linewidth}
        \centering
        \includegraphics[width=\linewidth]{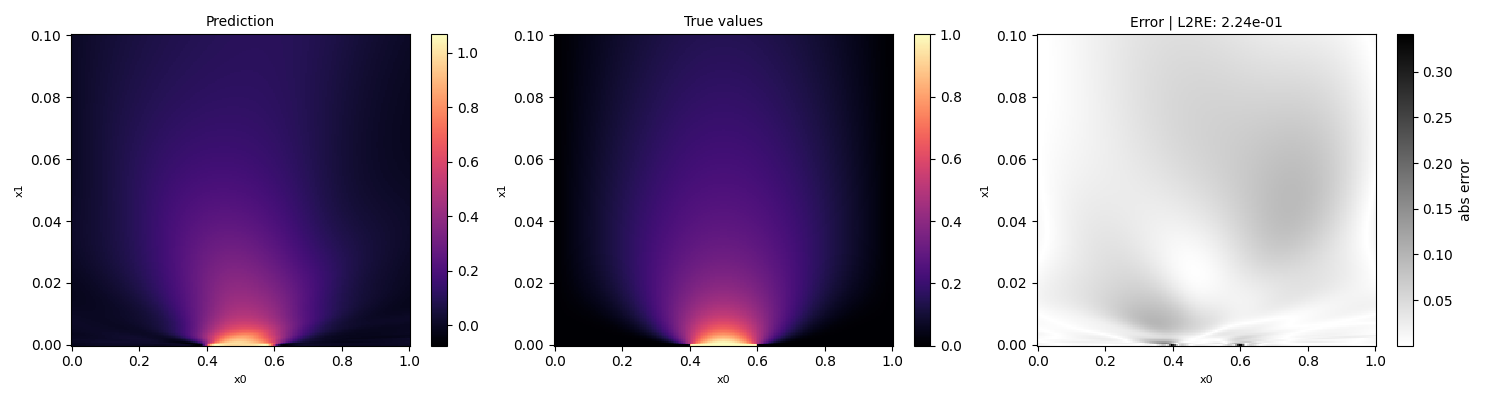}
        \caption{2D domain fit.}
        \label{fig-app:pinn_heat_step_domain}
    \end{subfigure}
    \caption{PINN results for heat equation with \IC{Step} IC.}
    \label{fig-app:pinn_heat_step}
\end{figure}

\begin{figure}[!htbp]
    \centering
    \begin{subfigure}{.48\linewidth}
        \centering
        \includegraphics[width=\linewidth]{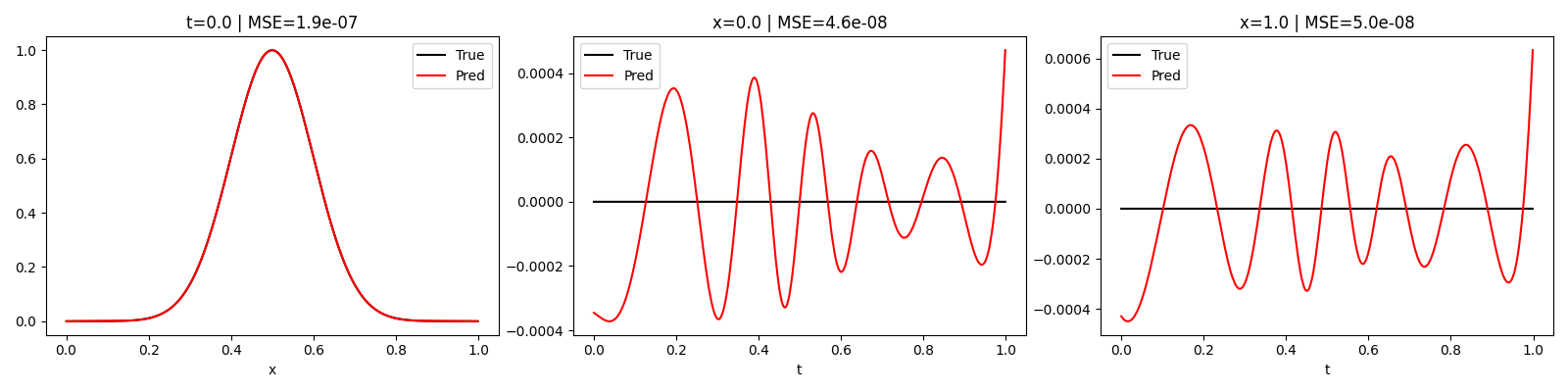}
        \caption{IBC fit.}
        \label{fig-app:pinn_wave_gauss_ibc}
    \end{subfigure} %
    \begin{subfigure}{.48\linewidth}
        \centering
        \includegraphics[width=\linewidth]{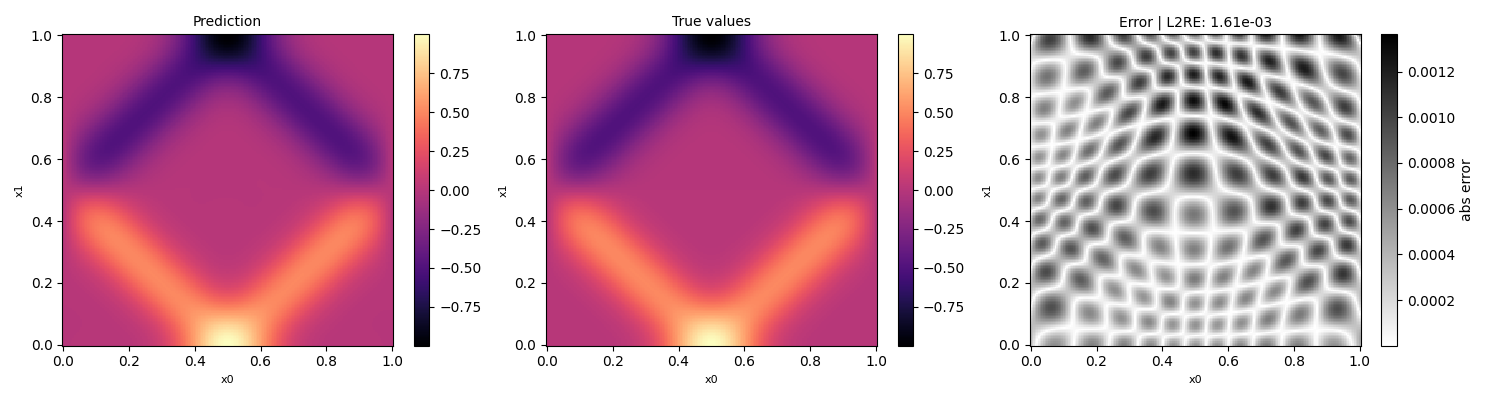}
        \caption{2D domain fit.}
        \label{fig-app:pinn_wave_gauss_domain}
    \end{subfigure}
    \caption{PINN results for wave equation with \IC{Gauss} IC.}
    \label{fig-app:pinn_wave_gauss}
\end{figure}

\begin{figure}[!htbp]
    \centering
    \begin{subfigure}{.48\linewidth}
        \centering
        \includegraphics[width=\linewidth]{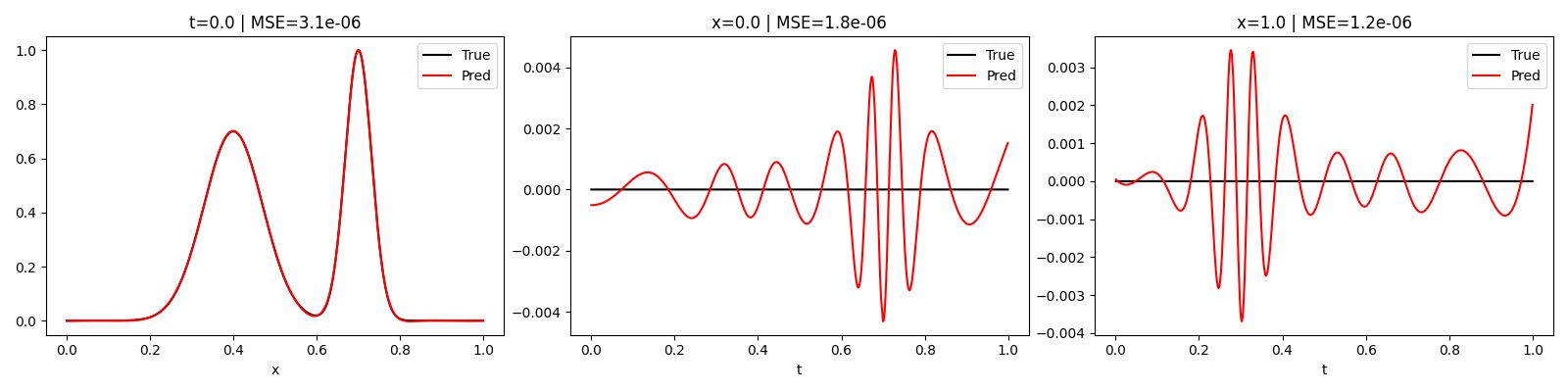}
        \caption{IBC fit.}
        \label{fig-app:pinn_wave_gaussmix_ibc}
    \end{subfigure} %
    \begin{subfigure}{.48\linewidth}
        \centering
        \includegraphics[width=\linewidth]{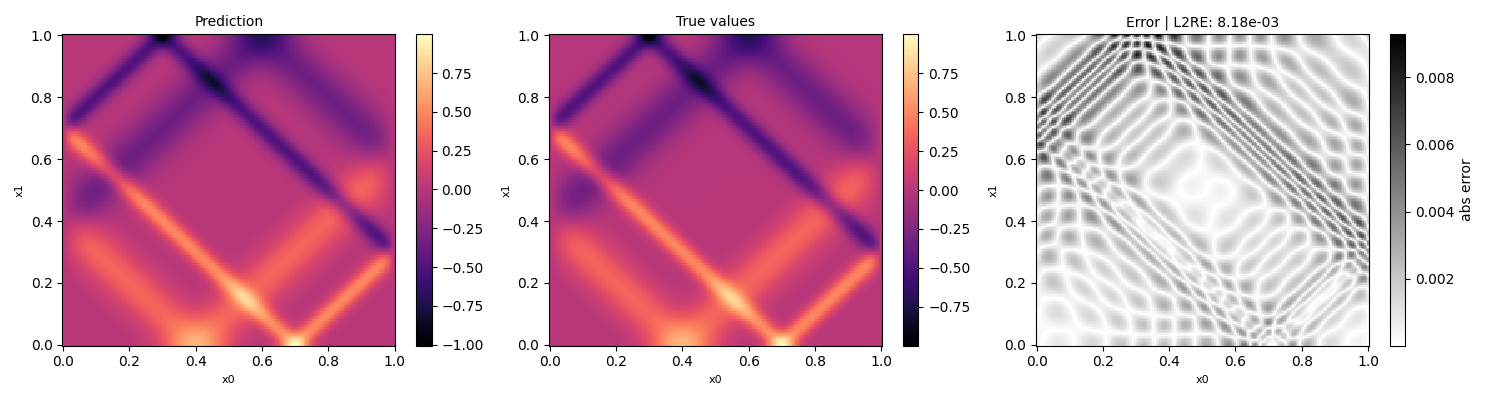}
        \caption{2D domain fit.}
        \label{fig-app:pinn_wave_gaussmix_domain}
    \end{subfigure}
    \caption{PINN results for wave equation with \IC{Gauss Mix} IC.}
    \label{fig-app:pinn_wave_gaussmix}
\end{figure}

\begin{figure}[!htbp]
    \centering
    \begin{subfigure}{.48\linewidth}
        \centering
        \includegraphics[width=\linewidth]{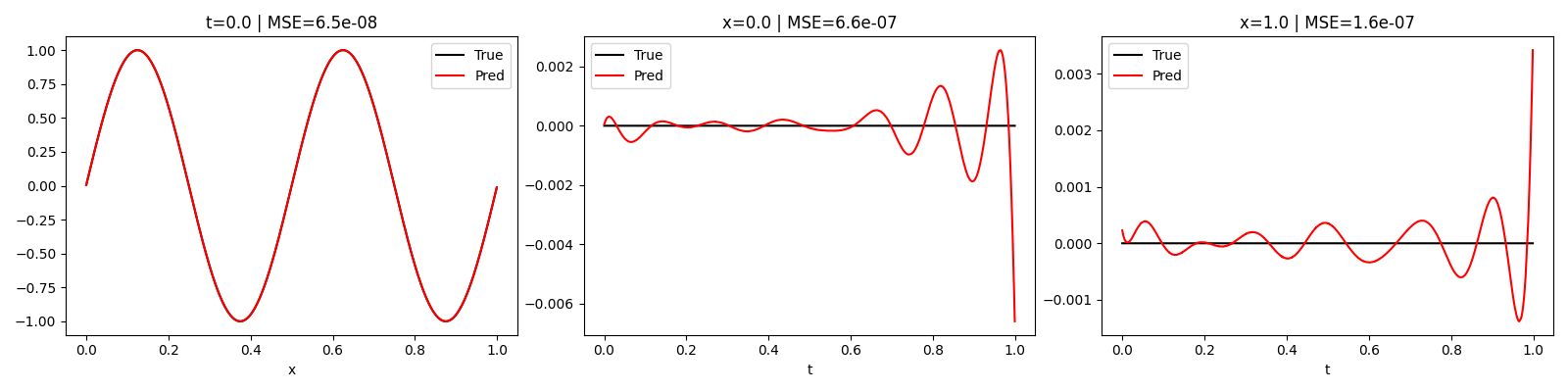}
        \caption{IBC fit.}
        \label{fig-app:pinn_wave_sine_ibc}
    \end{subfigure} %
    \begin{subfigure}{.48\linewidth}
        \centering
        \includegraphics[width=\linewidth]{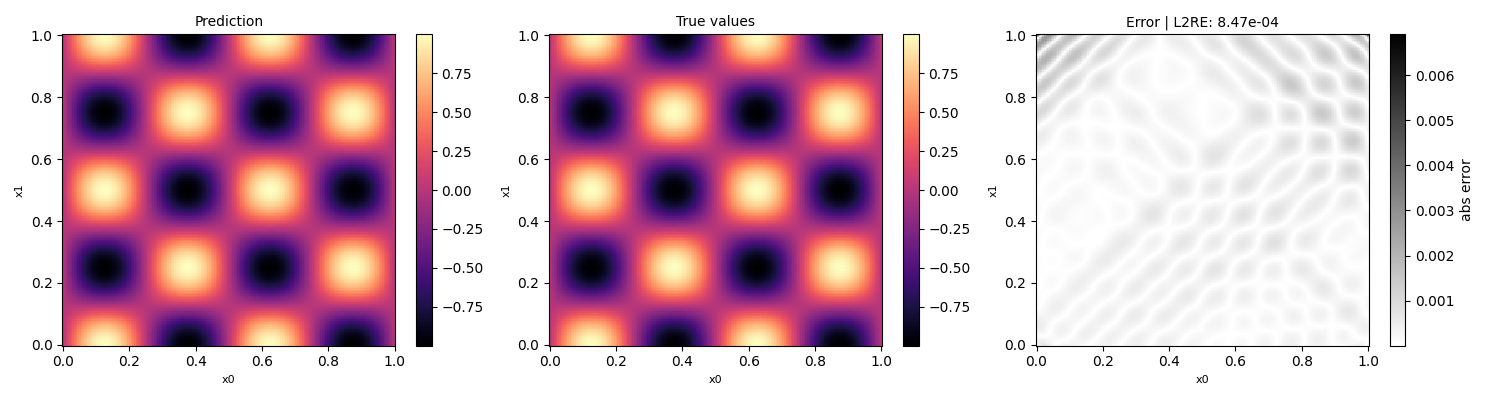}
        \caption{2D domain fit.}
        \label{fig-app:pinn_wave_sine_domain}
    \end{subfigure}
    \caption{PINN results for wave equation with \IC{Sine} IC.}
    \label{fig-app:pinn_wave_sine}
\end{figure}

\begin{figure}[!htbp]
    \centering
    \begin{subfigure}{.48\linewidth}
        \centering
        \includegraphics[width=\linewidth]{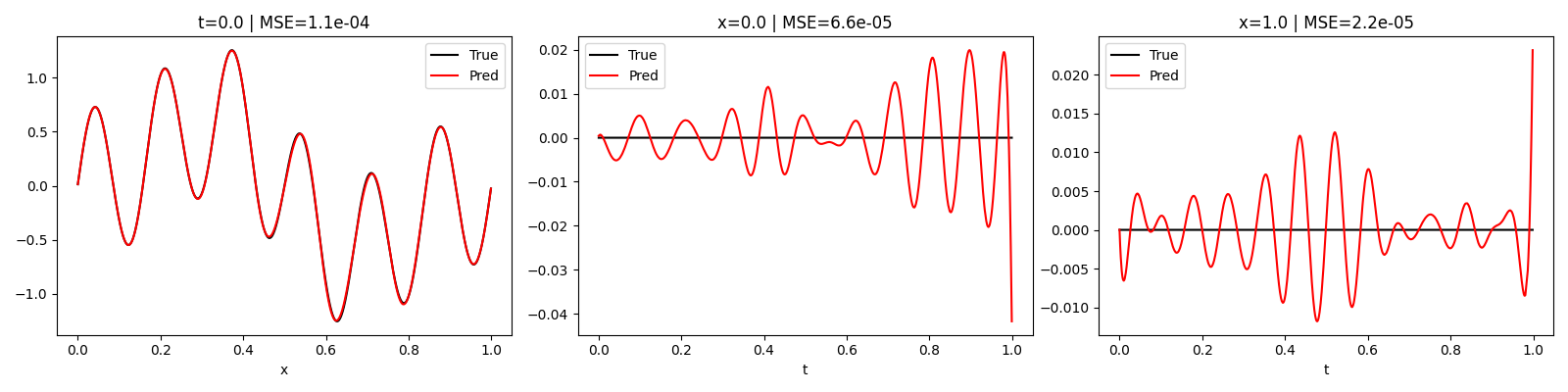}
        \caption{IBC fit.}
        \label{fig-app:pinn_wave_sinemix_ibc}
    \end{subfigure} %
    \begin{subfigure}{.48\linewidth}
        \centering
        \includegraphics[width=\linewidth]{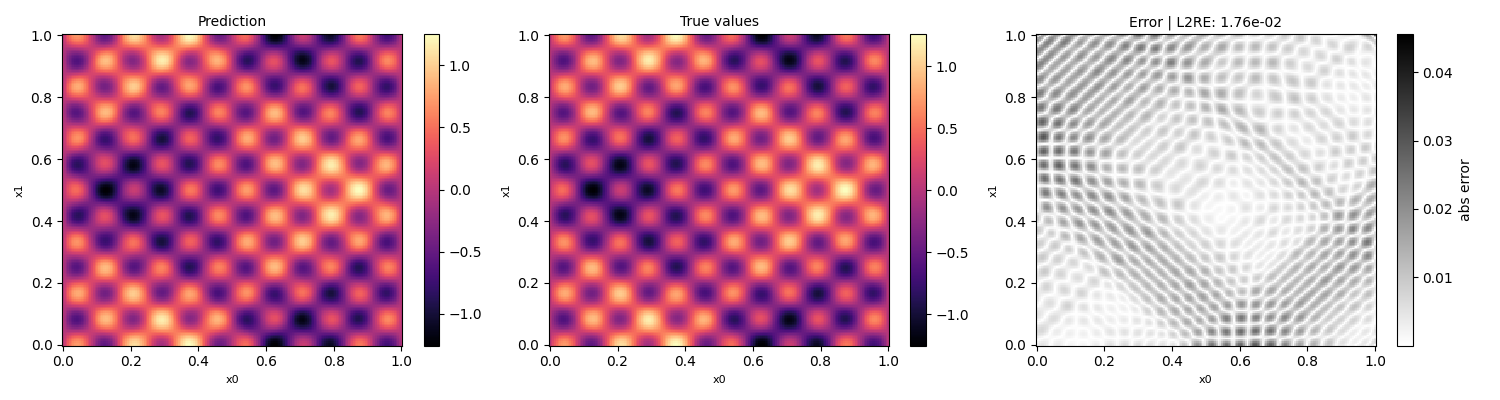}
        \caption{2D domain fit.}
        \label{fig-app:pinn_wave_sinemix_domain}
    \end{subfigure}
    \caption{PINN results for wave equation with \IC{Sine Mix} IC.}
    \label{fig-app:pinn_wave_sinemix}
\end{figure}

\begin{figure}[!htbp]
    \centering
    \begin{subfigure}{.48\linewidth}
        \centering
        \includegraphics[width=\linewidth]{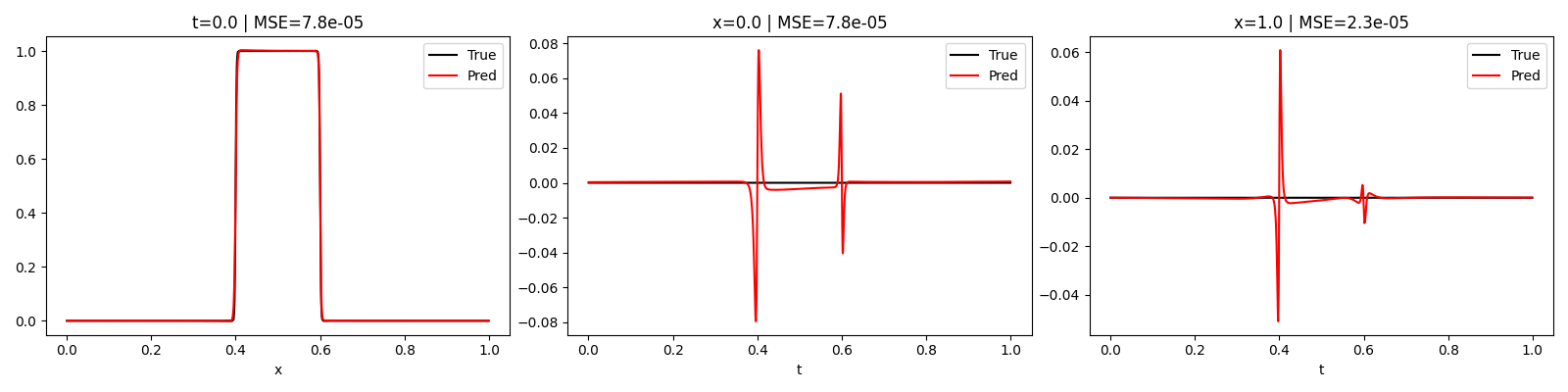}
        \caption{IBC fit.}
        \label{fig-app:pinn_wave_step_ibc}
    \end{subfigure} %
    \begin{subfigure}{.48\linewidth}
        \centering
        \includegraphics[width=\linewidth]{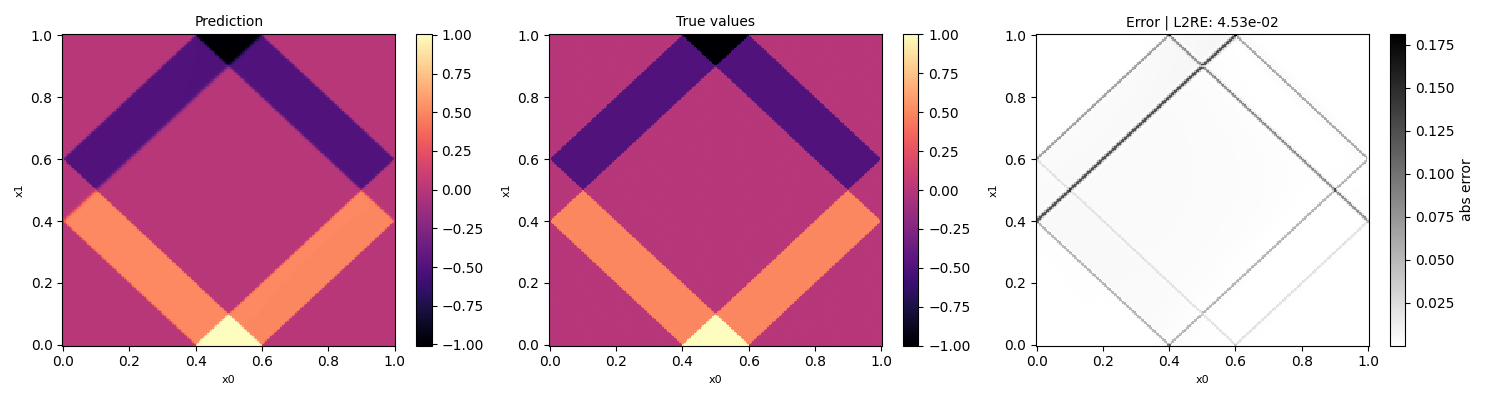}
        \caption{2D domain fit.}
        \label{fig-app:pinn_wave_step_domain}
    \end{subfigure}
    \caption{PINN results for wave equation with \IC{Step} IC.}
    \label{fig-app:pinn_wave_step}
\end{figure}

For completeness, we include the visual results of the PINN baselines in \Cref{fig-app:pinn_heat_poly} through \Cref{fig-app:pinn_wave_step}.

\clearpage

\end{document}